\PassOptionsToPackage{table}{xcolor}

\documentclass{article} 
\usepackage{iclr2027_conference,times}
\usepackage{amsmath,amsfonts,bm}

\def\eqref#1{equation~\ref{#1}}

\def\1{\bm{1}}

\DeclareMathAlphabet{\mathsfit}{\encodingdefault}{\sfdefault}{m}{sl}
\SetMathAlphabet{\mathsfit}{bold}{\encodingdefault}{\sfdefault}{bx}{n}

\usepackage{float,longtable,array,placeins,needspace}

\usepackage{hyperref}
\usepackage{url}

\usepackage{booktabs}
\usepackage{graphicx}
\usepackage{multirow}
\usepackage{makecell}
\usepackage{enumitem}

\usepackage[table]{xcolor}
\definecolor{gainrow}{RGB}{242,242,242}
\definecolor{brandblue}{RGB}{31,78,121}
\definecolor{abstractline}{RGB}{178,204,226}
\definecolor{brandorange}{RGB}{236,145,35}
\hypersetup{colorlinks=true,citecolor=brandblue,linkcolor=black,urlcolor=black}
\usepackage{booktabs}
\usepackage{colortbl}
\usepackage{wrapfig}
\usepackage{pgf}
\usepackage{array}
\usepackage{siunitx}
\usepackage{fontawesome5}
\usepackage{booktabs}
\usepackage{longtable}
\usepackage{array}
\usepackage{needspace}

\newcommand{\gfb}[1]{%
\pgfmathparse{abs(#1)}
\pgfmathsetmacro{\val}{\pgfmathresult}
\ifdim #1pt > 0pt
    \pgfmathsetmacro{\shadelevel}{min(70,\val/40.69*70)}
    \edef\temp{\noexpand\cellcolor{green!\shadelevel!white}}
    \temp
\else
    \ifdim #1pt < 0pt
        \pgfmathsetmacro{\shadelevel}{min(70,\val/56.05*70)}
        \edef\temp{\noexpand\cellcolor{red!\shadelevel!white}}
        \temp
    \fi
\fi
\textbf{#1}
}

\title{LIBERO-Pro2.0: A Benchmark for Evaluating
the Robustness of Embodied Foundation Models}

\author{%
\begin{minipage}[t]{\dimexpr\textwidth-2\tabcolsep\relax}
\normalfont\normalsize\centering
\textbf{Xueyang Zhou}\textsuperscript{1,*}\quad
\textbf{Yangming Xu}\textsuperscript{1,*}\quad
\textbf{Qianjiang Li}\textsuperscript{1}\quad
\textbf{Guiyao Tie}\textsuperscript{1}\\[3pt]
\textbf{Junhao Wu}\textsuperscript{2}\quad
\textbf{Ziheng Huang}\textsuperscript{2}\quad
\textbf{Guowen Zhang}\textsuperscript{1}\quad
\textbf{Zhijie Yuan}\textsuperscript{5}\\[3pt]
\textbf{Di Wu}\textsuperscript{1}\quad
\textbf{Duanfeng Chu}\textsuperscript{3}\quad
\textbf{Pan Zhou}\textsuperscript{1,\textdagger}\quad
\textbf{Lichao Sun}\textsuperscript{4}\quad
\textbf{Yongchao Chen}\textsuperscript{2}\\[8pt]
\small
\textsuperscript{1}Huazhong University of Science and Technology\quad
\textsuperscript{2}Tsinghua University\\[2pt]
\textsuperscript{3}Wuhan University of Technology\quad
\textsuperscript{4}Lehigh University\quad
\textsuperscript{5}University of Toronto\\[6pt]
\texttt{\{d202480819,U202312365,tgy,panzhou,D202581000\}@hust.edu.cn}\\[2pt]
\texttt{yongchaochen12@gmail.com}\quad
\texttt{lis221@lehigh.edu}\\[2pt]
\texttt{JunhaoWu7@outlook.com}\quad
\texttt{hzhstxdydsb@gmail.com}\\[2pt]
\texttt{racingemperor@outlook.com}\quad
\texttt{Georgeyuanzj@icloud.com}\\[6pt]
\textsuperscript{*}Equal contribution.\quad
\textsuperscript{\textdagger}Corresponding author: \texttt{panzhou@hust.edu.cn}.
\end{minipage}%
}

\newcommand{\projectwebpageurl}{https://racingemperor.github.io/LIBERO-Pro2.0}
\newcommand{\projectgithuburl}{https://github.com/LIBERO-Pro-Team/LIBERO-Pro2.0}
\newcommand{\projecthuggingfaceurl}{https://huggingface.co/datasets/zhouxueyang/LIBERO-Pro2.0-Eval}

\newcommand{\projectleaderboardurl}{https://racingemperor.github.io/LIBERO-Pro2.0/leaderboard.html}
\newcommand{\projectcommunityurl}{https://racingemperor.github.io/LIBERO-Pro2.0/community.html}

\definecolor{leaderboardgreen}{HTML}{228B57}
\definecolor{communitypurple}{HTML}{7C4DAB}

\makeatletter
\def\@maketitle{%
  \vspace*{-0.40in}%
  {\LARGE\bfseries\centering\@title\par}%
  \vspace{0.15in}%
  \@author\par
  \vspace{0.04in}%
  {\centering\sffamily\small
    \mbox{\href{\projectwebpageurl}{%
      \textcolor{brandblue}{%
        \faGlobe\hspace{0.35em}\textbf{Webpage}}}}%
    \hspace{0.75em}\textcolor{gray!45}{\textbar}\hspace{0.75em}%
    \mbox{\href{\projectgithuburl}{%
      \textcolor{black}{%
        \faGithub\hspace{0.35em}\textbf{GitHub}}}}%
    \hspace{0.75em}\textcolor{gray!45}{\textbar}\hspace{0.75em}%
    \mbox{\href{\projecthuggingfaceurl}{%
      \textcolor{brandorange}{%
        \faSmile\hspace{0.35em}\textbf{Hugging Face}}}}%
    \hspace{0.75em}\textcolor{gray!45}{\textbar}\hspace{0.75em}%
    \mbox{\href{\projectleaderboardurl}{%
      \textcolor{leaderboardgreen}{%
        \faTrophy\hspace{0.35em}\textbf{Leaderboard}}}}%
    \hspace{0.75em}\textcolor{gray!45}{\textbar}\hspace{0.75em}%
    \mbox{\href{\projectcommunityurl}{%
      \textcolor{communitypurple}{%
        \faUsers\hspace{0.35em}\textbf{Community}}}}%
    \par
  }%
  \vspace{0.08in}%
}
\makeatother

\fancypagestyle{titlepage}{%
  \fancyhf{}%
  \setlength{\headheight}{17pt}%
  \fancyhead[L]{\includegraphics[height=15pt]{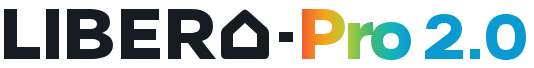}}%
  \renewcommand{\headrulewidth}{0.6pt}%
}

\renewenvironment{abstract}{%
  \par\vspace{0.1em}%
  \noindent
  {\sffamily\bfseries\footnotesize\color{brandblue}ABSTRACT}%
  \hspace{0.65em}%
  {\color{abstractline}\raisebox{0.35ex}{\rule{\dimexpr\linewidth-5.4em\relax}{0.6pt}}}\par
  \vspace{0.45em}%
  \small
  \setlength{\parskip}{0pt}%
}{%
  \par\vspace{0.5em}%
}

\iclrfinalcopy
\begin{document}

\maketitle
\pagestyle{fancy}
\fancyhead{}                            
\renewcommand{\headrulewidth}{0.4pt}     
\thispagestyle{titlepage}

\vspace{-10pt}
\begin{figure}[H]
    \centering
    \includegraphics[width=0.99\textwidth]{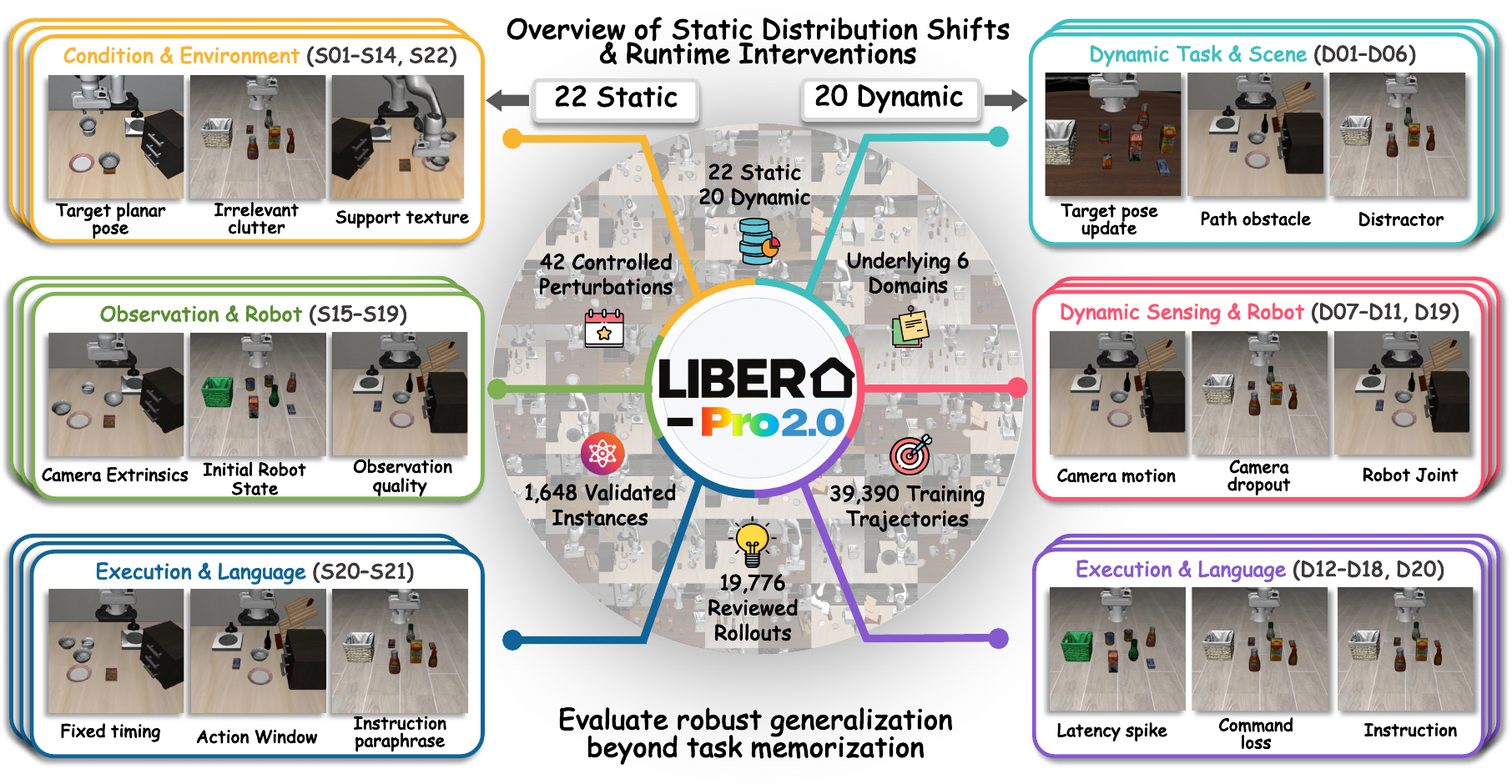}
    \caption{Overview of LIBERO-Pro2.0. Built upon LIBERO, our benchmark introduces 22 static distribution shifts and 20 runtime interventions across six dimensions: condition, environment, observation, robot, execution, and language. The left and right panels illustrate representative static and runtime perturbations, respectively.}
    \label{fig:promain}
    \vspace{-10pt}
\end{figure}

\begin{abstract}
LIBERO has emerged as a widely adopted benchmark for evaluating robotic manipulation policies. However, strong performance on LIBERO does not necessarily indicate robust embodied capabilities, as models may exploit benchmark-specific regularities rather than generalize to diverse operating conditions. In this work, we introduce \textbf{LIBERO-Pro2.0}, a comprehensive benchmark for systematically evaluating the robustness of embodied foundation models under distribution shifts. LIBERO-Pro2.0 encompasses \textbf{22 static perturbations} and \textbf{20 runtime interventions} across six dimensions: Condition, Environment, Observation, Execution, Robot State Configuration, and Language, yielding \textbf{1,648 validated shifted task instances}. To support robustness-oriented learning, we further collect \textbf{39,390 training trajectories} through model-generated and human demonstrations. All shifted task instances undergo six rounds of iterative human review and correction to ensure physical plausibility and task validity. We evaluate 17 representative robot policies spanning vision-language-action models, world-action models, and robustness-oriented policies. Our results reveal substantial robustness gaps across all evaluated paradigms: under runtime distribution shifts, the evaluated models exhibit an average performance drop of \textbf{26.5 percentage points}, with individual drops reaching \textbf{32.5 points}. These findings highlight the gap between nominal benchmark performance and robustness under distribution shifts, underscoring the need for systematic robustness evaluation of embodied foundation models.
\end{abstract}

\section{Introduction}
\label{sec:introduction}

Reliable evaluation is essential for measuring progress in embodied intelligence~\citep{yu2020metaworld,james2020rlbench,mees2022calvin,gu2023maniskill2,mu2025robotwin,zhang2025vlabench}. LIBERO has emerged as a de facto standard for tabletop manipulation~\citep{liu2023libero}, with increasingly many systems approaching saturated performance under its standard protocol~\citep{kim2024openvla,black2025pi0,kim2025openvlaoft,qu2025spatialvla,lee2025molmoact,fang2026molmoact2}. Yet a fundamental question remains: \textit{do these high success rates reflect generalizable embodied capabilities, or merely strong performance within a familiar training–evaluation distribution?}

To examine this question, we revisit high-performing systems on LIBERO and observe three striking failure patterns under simple task variations: (1) when the target object is replaced or even removed, models still execute the original grasping behavior; (2) when the task instruction is replaced with meaningless tokens, models continue to produce nearly identical action trajectories; and (3) when the target object is moved away from its familiar position, models still move toward the location commonly seen during training. These qualitative observations motivate a systematic evaluation of sensitivity to task-relevant visual, linguistic, and spatial changes, but do not by themselves establish memorization as the underlying failure mechanism.

These failures expose a broader limitation of current evaluation protocols. Standard benchmarks largely test whether a policy can reproduce successful behavior under conditions close to those seen during training~\citep{yu2020metaworld,james2020rlbench,mees2022calvin,gu2023maniskill2,liu2023libero}, while deployment requires the same system to remain effective when the scene changes before execution, when unexpected events occur during interaction~\citep{li2024simplerenv,fei2026liberoplus,zhang2026wamrobustness}, and when an initial action fails. Consequently, nominal task success alone does not characterize robustness when operating conditions change. A meaningful evaluation protocol should therefore separate training from evaluation distributions and explicitly test both generalization to static shifts and adaptation to runtime interventions.

To this end, we introduce LIBERO-Pro2.0, a large-scale robustness benchmark for embodied foundation models under static distribution shifts and runtime interventions. LIBERO-Pro2.0 defines 42 perturbation types (22 static and 20 runtime), yielding \textbf{1,648 perturbed task variants}. These perturbations span scene and object configurations, visual observations, camera settings, robot states, among others. We reserve 8 comprehensively perturbed tasks for evaluation and release the remaining perturbations for training, collecting \textbf{39,390 training trajectories}. Each perturbed task undergoes six rounds of human verification to ensure validity and reproducibility. We evaluate 17 representative policies spanning vision-language-action models, world-action models, and robustness-oriented policies. Results reveal substantial robustness degradation under both static and runtime distribution shifts, with runtime shifts causing an average performance drop of \textbf{26.5 percentage points}. Moreover, robustness varies substantially across perturbation types. High nominal success does not guarantee small robustness losses or preservation of policy rankings under distribution shifts. Our main contributions are threefold:
\begin{itemize}[leftmargin=1.2em]
    \item We introduce \textbf{LIBERO-Pro2.0}, a large-scale robustness benchmark for embodied foundation models, extending LIBERO beyond nominal task success to systematically evaluate robustness under both static distribution shifts and runtime interventions.

    \item We develop a comprehensive perturbation suite with \textbf{42 perturbation types}, yielding \textbf{1,648 perturbed task variants} and \textbf{39,390 training trajectories} across diverse environments, observations, instructions, robot states, and execution conditions, with rigorous human verification.

    \item We conduct a large-scale evaluation of \textbf{17 representative policies} across different model paradigms, revealing substantial robustness degradation under both static and runtime shifts and showing that nominal LIBERO performance does not consistently reflect robustness.
\end{itemize}

\begin{figure*}[t]
\vspace{-5pt}
    \centering
    \includegraphics[width=\textwidth]{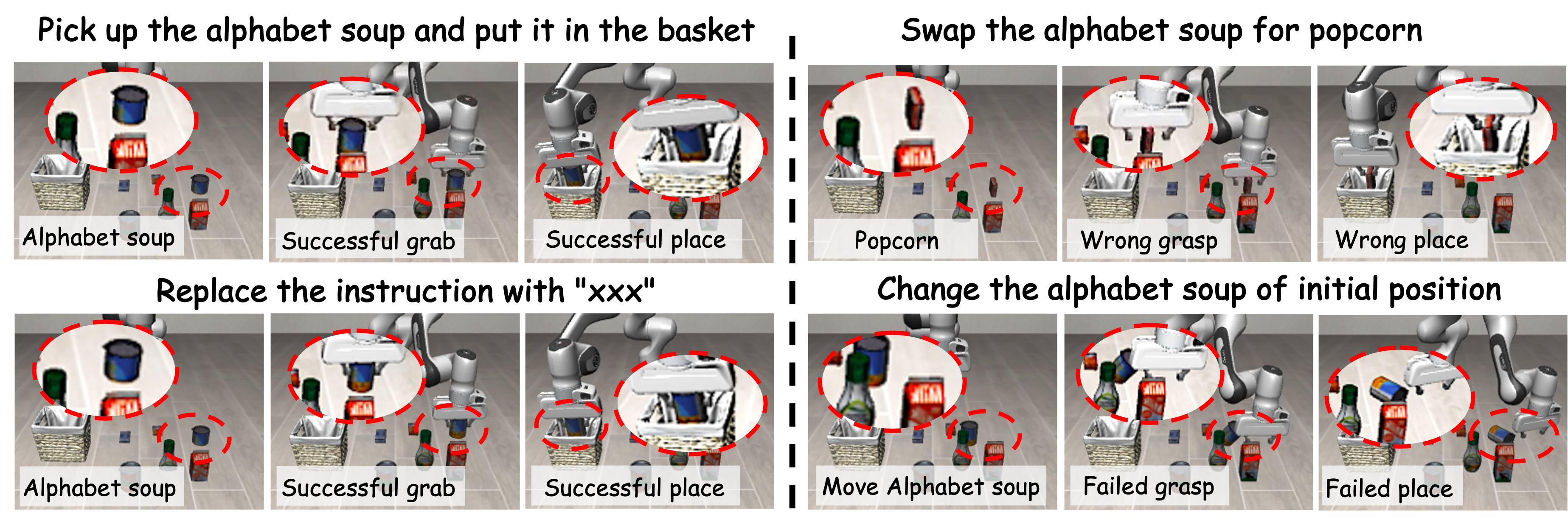}
    \caption{\textbf{Failure patterns under task variations.} Examples for the task ``\textit{Pick up the alphabet soup and put it in the basket.}'' The policy behaves similarly despite changes to the target object, instruction, or object position.}
    \label{fig:failure_patterns}
\vspace{-10pt}
\end{figure*}
\section{Preliminaries}
\label{preliminaries}
\subsection{Embodied Manipulation Tasks}

We consider a language-conditioned manipulation task
$\mathcal{T}=(\ell,e,\mathcal{G})$, where $\ell$ is the task instruction,
$e$ specifies the environment configuration, and $\mathcal{G}$ is the
task-success predicate. Starting from an initial state $s_0$, a policy
$\pi$ interacts with the environment for a finite episode horizon
$T_{\mathrm{ep}}$, receiving an observation $o_t$ and producing an action
$a_t\sim\pi(\cdot\mid o_t,\ell_t)$ at each step.
The instruction input satisfies $\ell_t=\ell$ by default; D20 permits
semantics-preserving restatements during execution without changing
the task objective or success predicate.
The resulting trajectory is
$\tau=(s_0,a_0,\ldots,s_{T_{\mathrm{ep}}})$, and an episode is successful
if $\mathcal{G}(\tau)=1$.

\subsection{Evaluation under Distribution Shifts}

The original LIBERO evaluation distribution closely resembles its training distribution. In real-world deployment, however, operating conditions can deviate substantially from those encountered during training, making the original evaluation insufficient for assessing policy generalization. We therefore introduce controlled distribution shifts into the LIBERO evaluation setting. Given the original evaluation distribution $\mathcal{D}_{\mathrm{eval}}$, an evaluation transformation $\phi$ parameterized by $\eta$ induces a shifted distribution $\mathcal{D}_{\mathrm{eval}}^{\phi} = \phi(\mathcal{D}_{\mathrm{eval}};\eta)$, while preserving the task objective and success criterion. Performance under $\mathcal{D}_{\mathrm{eval}}^{\phi}$ measures policy generalization when evaluation conditions deviate from those represented during training.

\subsection{Static Shifts and Runtime Interventions}

We distinguish two types of evaluation transformations according to their temporal behavior during an episode. A static shift $\phi_S$ is configured before policy execution, yielding
$\mathcal{T}^{S}=\phi_S(\mathcal{T};\eta)$.
Its intervention parameters or fault rules remain fixed throughout
the episod. In contrast, a \emph{runtime intervention} $\phi^{D}$ is applied during policy execution, with its occurrence and configuration determined by the interaction history $\tau_{0:t}$, i.e., $\phi^{D}(\mathcal{T};t,\tau_{0:t},\eta)$. Static shifts evaluate generalization under shifted but stationary conditions, whereas runtime interventions evaluate policy robustness to changes that arise during interaction.
\section{LIBERO-Pro2.0}
\label{sec:libero_pro}

LIBERO-Pro2.0 extends LIBERO with robustness evaluation under relevant variations in environments, observations, robot states, execution processes, and task instructions, while preserving task validity and feasibility. An overview of LIBERO-Pro2.0 and its evaluation dimensions is shown in Fig.~\ref{fig:promain}.

\begin{figure}[t]
    \centering
    \vspace{-15pt}
    \includegraphics[width=0.99\textwidth]{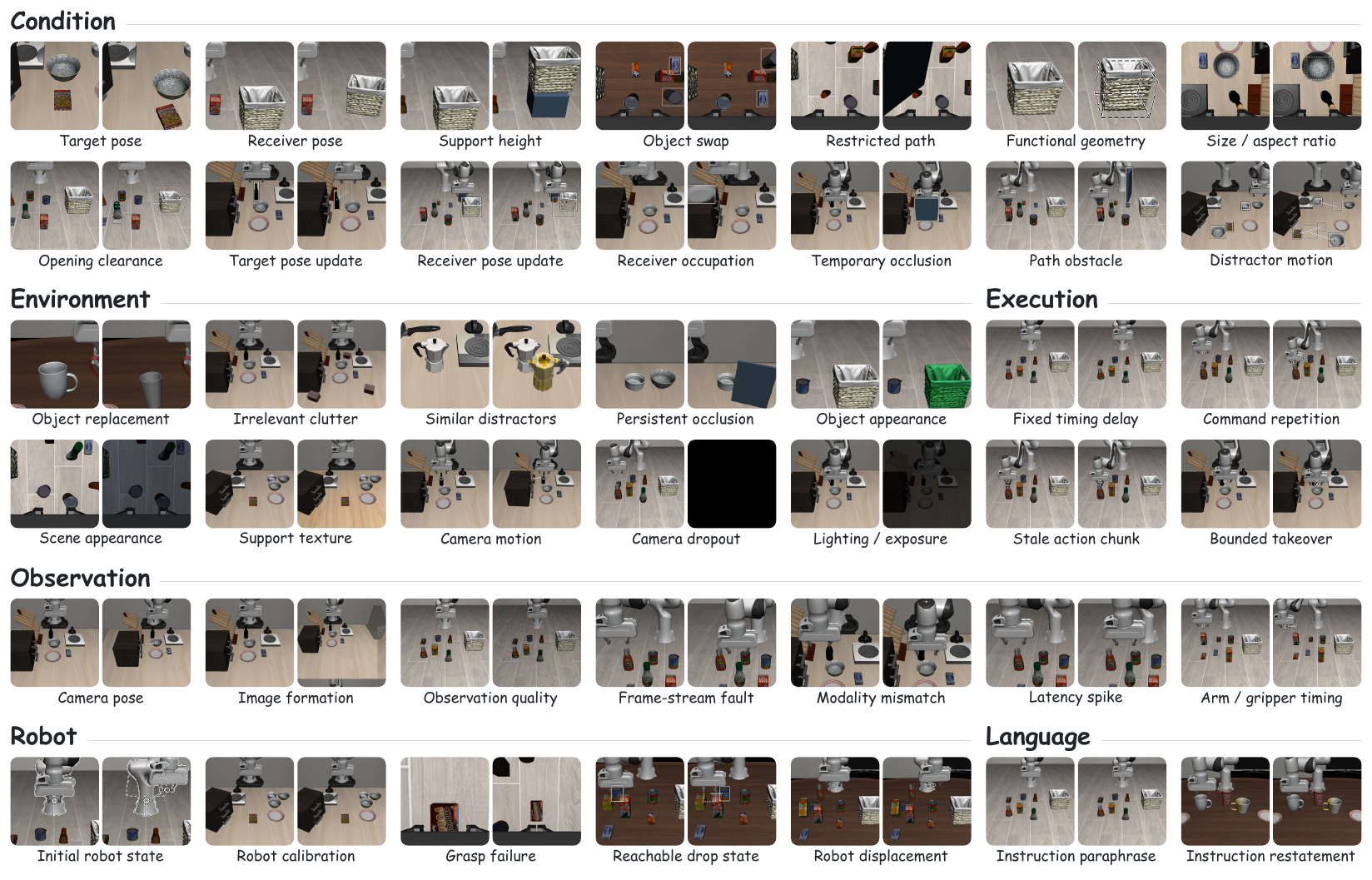}
    \caption{Cases of LIBERO-Pro2.0.}
    \label{fig:promain}
    \vspace{-20pt}
\end{figure}

\subsection{Systematic Distribution Shift Design}
\label{sec:distribution_shift}

We consider two forms of distribution shift according to when the operating condition changes: \textbf{Static Distribution Shifts} and \textbf{Runtime Distribution Shifts}.Static shifts are configured before execution, with intervention parameters or fault rules held fixed throughout an episode, while runtime shifts occur during task execution and alter the environment, observation, robot state, or execution process. The former evaluates out-of-distribution generalization, while the latter evaluates online adaptation and recovery.

\paragraph{Static distribution shifts.} Static distribution shifts characterize changes that are established before execution and remain stationary throughout an episode. Let $\phi^{S}\in\Phi^{S}$ denote a static shift operator and $\eta$ its configuration. Applying $\phi^{S}$ to a nominal task $\mathcal{T}$ induces a shifted task $\mathcal{T}^{S}=\phi^{S}(\mathcal{T};\eta)$, where $\eta$ is specified before the rollout and held fixed throughout execution. LIBERO-Pro2.0 instantiates \textbf{22 static distribution shifts} over the task environment, observation process, robot configuration, and language condition, covering object configuration and appearance, scene composition, camera and observation conditions, robot initialization and calibration, temporal configuration, and instruction expression. These shifts reflect deployment-relevant variations that arise in real-world robotic manipulation, making robustness to such out-of-distribution conditions essential for deployment.

\begin{wrapfigure}{r}{0.55\textwidth}
    \centering
    \vspace{-26pt}
    \includegraphics[width=0.55\textwidth]{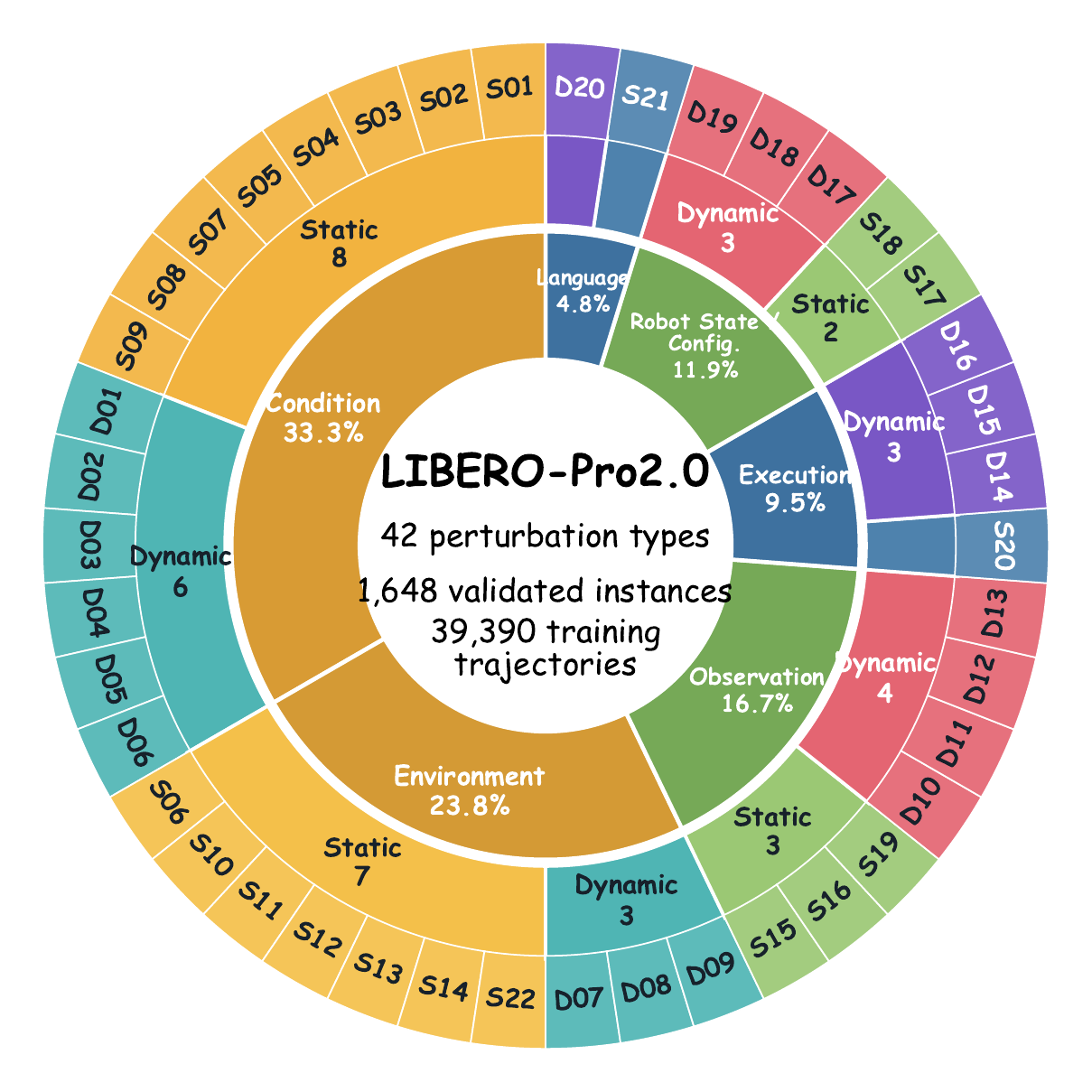}
    \caption{Perturbation distribution in LIBERO-Pro2.0.}
    \label{fig:your_figure}
    \vspace{-26pt}
\end{wrapfigure}

\paragraph{Runtime distribution shifts.} Runtime distribution shifts characterize changes that emerge after task execution has begun. Let $\phi^{D}\in\Phi^{D}$ denote a runtime shift operator and $\eta$ its configuration. Each shift is associated with a trigger function $g_{\phi}(\tau_{0:t})$ defined over the interaction trajectory $\tau_{0:t}$; the trigger determines when the shift is introduced, while $\eta$ specifies its realization and temporal extent. LIBERO-Pro2.0 instantiates \textbf{20 runtime distribution shifts} over the task environment, observation process, robot and action execution, and language condition, covering object or receiver relocation, dynamic occlusion and obstacles, camera and observation failures, frame-stream and synchronization anomalies, transient latency, action-execution disruptions, robot-state changes, semantics-preserving instruction restatements, and recoverable execution failures. These shifts capture execution-time changes that can occur during real-world deployment, where a robust policy must continuously adapt its behavior and recover from unexpected deviations to maintain stable task execution.

\subsection{Data Collection}
\label{sec:data_collection}

LIBERO-Pro2.0 is built from the 40 tasks in LIBERO-Goal, LIBERO-Spatial, LIBERO-Object, and LIBERO-10. Using random seed 142, we sample two tasks from each suite for held-out evaluation, yielding \textbf{8 evaluation tasks} and \textbf{32 training tasks}. After applying all applicable distribution shifts and filtering invalid task--shift combinations, the training and evaluation sets contain \textbf{1,313} and \textbf{335} valid instances, respectively, for a total of \textbf{1,648 validated shifted task instances}. For robustness-oriented learning, we collect \textbf{30 trajectories for each training instance}, yielding \textbf{39,390 trajectories} from $\pi_{0.5}$, GPT-5.6 Sol with Codex, and human demonstrations. All authors conduct \textbf{six rounds of iterative human review and correction} over all \textbf{1,648 distribution-shift instances}. In each round, every instance is executed twice using $\pi_{0.5}$, producing approximately \textbf{3,296 rollout videos} per round and \textbf{19,776 reviewed rollouts} across six rounds, totaling more than \textbf{82 hours of footage}. Reviewers verify that each shift is correctly realized and that tasks remain semantically valid and physically executable, including spatial consistency after relocation, valid occlusions or obstacles, feasible execution paths, and correct runtime triggering. Temporal and action-stream shifts are additionally checked against execution logs when necessary. Invalid or ambiguous instances are corrected and re-evaluated, while task--shift combinations that fail validation are removed.


\section{Experiments}

\begin{table*}[t]
\centering
\vspace{-10pt}
\caption{
Leaderboard of robot-policy robustness under \textbf{static distribution shifts}.
Category scores and Avg.\ report mean success rates (\%) over valid
suite--perturbation entries across four LIBERO suites.
Base denotes nominal success, and $\Delta=\mathrm{Avg.}-\mathrm{Base}$
is measured in percentage points. All calculations precede rounding.
OpenVLA-OFT variants follow LIBERO-Plus~\citep{fei2026liberoplus}:
$_{\mathrm{w}}$ excludes wrist-camera input,
$_{\mathrm{m}}$ uses joint fine-tuning across the four original suites,
and $+$ uses mixed fine-tuning on LIBERO-Plus data.
}
\label{tab:static_leaderboard}

\renewcommand{\arraystretch}{1.30}
\setlength{\tabcolsep}{7.5pt}

\resizebox{0.98\textwidth}{!}{%
\begin{tabular}{cccccccccc}
\specialrule{1.3pt}{0pt}{0pt}

\textbf{Model}
&
\begin{tabular}[c]{@{}c@{}}
\textbf{Cond.}\\[-1pt]
{\scriptsize\textit{S01--S05, S07--S09}}
\end{tabular}
&
\begin{tabular}[c]{@{}c@{}}
\textbf{Env.}\\[-1pt]
{\scriptsize\textit{S06, S10--S14, S22}}
\end{tabular}
&
\begin{tabular}[c]{@{}c@{}}
\textbf{Obs.}\\[-1pt]
{\scriptsize\textit{S15--S16, S19}}
\end{tabular}
&
\begin{tabular}[c]{@{}c@{}}
\textbf{Exec.}\\[-1pt]
{\scriptsize\textit{S20}}
\end{tabular}
&
\begin{tabular}[c]{@{}c@{}}
\textbf{Robot}\\[-1pt]
{\scriptsize\textit{S17--S18}}
\end{tabular}
&
\begin{tabular}[c]{@{}c@{}}
\textbf{Lang.}\\[-1pt]
{\scriptsize\textit{S21}}
\end{tabular}
& \textbf{Avg.}
& \textbf{Base}
& \textcolor{red!80!black}{\boldmath$\Delta$}
\\
\specialrule{0.9pt}{0pt}{0pt}
\rowcolor{gray!15}
\multicolumn{10}{c}{\textit{Vision-Language-Action Policies}} \\
\specialrule{0.7pt}{0pt}{0pt}

OpenVLA \citep{kim2024openvla}
& 25.2 & 37.6 & 34.4 & 0.0 & 54.0 & 67.5
& \cellcolor{red!6}33.9
& \cellcolor{green!14}64.6
& \textcolor{red!80!black}{$-30.7$}
\\

OpenVLA-OFT \citep{kim2025openvlaoft}
& 44.1 & 82.0 & 46.3 & 0.0 & 66.5 & 94.6
& \cellcolor{red!15}58.9
& \cellcolor{green!29}97.1
& \textcolor{red!80!black}{$-38.2$}
\\

OpenVLA-OFT$_{\mathrm{w}}$ \citep{kim2025openvlaoft}
& 40.1 & 64.9 & 67.6 & 0.4 & 55.4 & 90.0
& \cellcolor{red!12}53.8
& \cellcolor{green!25}89.2
& \textcolor{red!80!black}{$-35.4$}
\\

OpenVLA-OFT$_{\mathrm{m}}$ \citep{kim2025openvlaoft}
& 45.9 & 78.3 & 49.7 & 0.0 & 72.1 & 92.1
& \cellcolor{red!15}59.3
& \cellcolor{green!27}92.5
& \textcolor{red!80!black}{$-33.2$}
\\

OpenVLA-OFT+ \citep{fei2026liberoplus}
& 47.2 & 78.5 & 45.0 & 0.0 & 73.8 & 90.0
& \cellcolor{red!15}59.2
& \cellcolor{green!27}92.1
& \textcolor{red!80!black}{$-32.9$}
\\

UniVLA \citep{bu2025univla}
& 39.8 & 59.5 & 58.1 & 0.0 & 79.6 & 91.7
& \cellcolor{red!12}52.9
& \cellcolor{green!27}92.1
& \textcolor{red!80!black}{$-39.2$}
\\

$\pi_{0}$ \citep{black2025pi0}
& 25.0 & 53.8 & 60.0 & 5.0 & 19.2 & 46.3
& \cellcolor{red!8}38.6
& \cellcolor{green!14}63.8
& \textcolor{red!80!black}{$-25.1$}
\\

$\pi_{0}$-FAST \citep{pertsch2025fast}
& 43.8 & 71.5 & 39.9 & 0.0 & 57.7 & 81.2
& \cellcolor{red!12}53.2
& \cellcolor{green!23}80.4
& \textcolor{red!80!black}{$-27.3$}
\\

$\pi_{0.5}$ \citep{intelligence2025pi_}
& 60.7 & 83.9 & 43.3 & 0.8 & 85.0 & 93.3
& \cellcolor{red!20}66.7
& \cellcolor{green!27}93.3
& \textcolor{red!80!black}{$-26.6$}
\\

X-VLA \citep{zheng2025xvla}
& 42.1 & 81.7 & 49.3 & 71.7 & 61.7 & 90.8
& \cellcolor{red!16}61.2
& \cellcolor{green!28}95.0
& \textcolor{red!80!black}{$-33.8$}
\\

GR00T N1.7 \citep{bjorck2025gr00t}
& 46.6 & 79.2 & 48.5 & 0.0 & 77.9 & 88.8
& \cellcolor{red!16}60.0
& \cellcolor{green!25}90.4
& \textcolor{red!80!black}{$-30.4$}
\\

MolmoAct2 \citep{fang2026molmoact2}
& 63.8 & 85.0 & 65.1 & 0.0 & 85.0 & 94.6
& \cellcolor{red!24}71.2
& \cellcolor{green!29}97.2
& \textcolor{red!80!black}{$-26.0$}
\\

\specialrule{0.9pt}{1.2pt}{0pt}
\rowcolor{gray!15}
\multicolumn{10}{c}{\textit{World-Action Models}} \\
\specialrule{0.7pt}{0pt}{0pt}

Cosmos Policy \citep{kim2026cosmospolicy}
& 56.9 & 89.3 & 70.8 & 4.2 & 82.5 & 93.3
& \cellcolor{red!23}70.8
& \cellcolor{green!29}97.5
& \textcolor{red!80!black}{$-26.7$}
\\

FastWAM \citep{yuan2026fastwam}
& 43.8 & 84.7 & 5.2 & 2.8 & 87.5 & 90.3
& \cellcolor{red!13}55.9
& \cellcolor{green!30}97.8
& \textcolor{red!80!black}{$-41.8$}
\\

LingBot-VA \citep{li2026lingbotva}
& 52.2 & 86.1 & 35.7 & 86.7 & 98.3 & 97.1
& \cellcolor{red!22}68.7
& \cellcolor{green!30}98.5
& \textcolor{red!80!black}{$-29.7$}
\\

\specialrule{0.9pt}{1.2pt}{0pt}
\rowcolor{gray!15}
\multicolumn{10}{c}{\textit{Robustness-Oriented Policies}} \\
\specialrule{0.7pt}{0pt}{0pt}

Anchor-Align \citep{dalal2026generalizable}
& 43.2 & 74.3 & 38.6 & 0.8 & 76.5 & 90.8
& \cellcolor{red!13}55.9
& \cellcolor{green!26}91.3
& \textcolor{red!80!black}{$-35.4$}
\\

RIPT-VLA \citep{tan2025interactive}
& 40.6 & 76.0 & 42.8 & 0.0 & 65.0 & 86.7
& \cellcolor{red!12}54.8
& \cellcolor{green!23}87.9
& \textcolor{red!80!black}{$-33.1$}
\\

\specialrule{1.3pt}{0pt}{0pt}
\end{tabular}%
}

\vspace{-15pt}
\end{table*}

\begin{table*}[htb]
\centering
\caption{
Leaderboard of robot-policy robustness under
\textbf{runtime distribution shifts}.
}
\label{tab:runtime_leaderboard}

\renewcommand{\arraystretch}{1.30}
\setlength{\tabcolsep}{7.5pt}

\resizebox{0.98\textwidth}{!}{%
\begin{tabular}{cccccccccc}

\specialrule{1.3pt}{0pt}{0pt}

\textbf{Model}
&
\begin{tabular}[c]{@{}c@{}}
\textbf{Cond.}\\[-1pt]
{\scriptsize\textit{D01--D06}}
\end{tabular}
&
\begin{tabular}[c]{@{}c@{}}
\textbf{Env.}\\[-1pt]
{\scriptsize\textit{D07--D09}}
\end{tabular}
&
\begin{tabular}[c]{@{}c@{}}
\textbf{Obs.}\\[-1pt]
{\scriptsize\textit{D10--D13}}
\end{tabular}
&
\begin{tabular}[c]{@{}c@{}}
\textbf{Exec.}\\[-1pt]
{\scriptsize\textit{D14--D16}}
\end{tabular}
&
\begin{tabular}[c]{@{}c@{}}
\textbf{Robot}\\[-1pt]
{\scriptsize\textit{D17--D19}}
\end{tabular}
&
\begin{tabular}[c]{@{}c@{}}
\textbf{Lang.}\\[-1pt]
{\scriptsize\textit{D20}}
\end{tabular}
& \textbf{Avg.}
& \textbf{Base}
& \textcolor{red!80!black}{\boldmath$\Delta$}
\\

\specialrule{0.9pt}{0pt}{0pt}

\rowcolor{gray!15}
\multicolumn{10}{c}{\textit{Vision-Language-Action Policies}}
\\
\specialrule{0.7pt}{0pt}{0pt}

OpenVLA \citep{kim2024openvla}
& 20.6 & 41.9 & 41.3 & 53.6 & 29.9 & 63.8
& \cellcolor{red!8}36.4
& \cellcolor{green!14}64.6
& \textcolor{red!80!black}{$-28.2$}
\\

OpenVLA-OFT \citep{kim2025openvlaoft}
& 42.8 & 63.1 & 80.4 & 85.3 & 57.6 & 95.4
& \cellcolor{red!19}64.6
& \cellcolor{green!29}97.1
& \textcolor{red!80!black}{$-32.5$}
\\

OpenVLA-OFT$_{\mathrm{w}}$ \citep{kim2025openvlaoft}
& 34.9 & 77.8 & 74.6 & 81.0 & 54.0 & 91.7
& \cellcolor{red!17}61.9
& \cellcolor{green!25}89.2
& \textcolor{red!80!black}{$-27.3$}
\\

OpenVLA-OFT$_{\mathrm{m}}$ \citep{kim2025openvlaoft}
& 39.6 & 61.4 & 80.4 & 85.7 & 63.6 & 93.8
& \cellcolor{red!19}64.3
& \cellcolor{green!27}92.5
& \textcolor{red!80!black}{$-28.2$}
\\

OpenVLA-OFT+ \citep{fei2026liberoplus}
& 41.3 & 56.9 & 79.8 & 84.2 & 64.0 & 93.8
& \cellcolor{red!18}63.8
& \cellcolor{green!27}92.1
& \textcolor{red!80!black}{$-28.3$}
\\

UniVLA \citep{bu2025univla}
& 30.9 & 71.7 & 78.9 & 81.5 & 56.2 & 90.8
& \cellcolor{red!17}61.0
& \cellcolor{green!27}92.1
& \textcolor{red!80!black}{$-31.1$}
\\

$\pi_{0}$ \citep{black2025pi0}
& 24.9 & 56.9 & 52.7 & 30.8 & 7.2 & 61.3
& \cellcolor{red!8}35.3
& \cellcolor{green!14}63.8
& \textcolor{red!80!black}{$-28.4$}
\\

$\pi_{0}$-FAST \citep{pertsch2025fast}
& 45.6 & 45.4 & 60.9 & 70.0 & 58.3 & 79.2
& \cellcolor{red!14}55.9
& \cellcolor{green!23}80.4
& \textcolor{red!80!black}{$-24.5$}
\\

$\pi_{0.5}$ \citep{intelligence2025pi_}
& 60.5 & 58.1 & 86.6 & 88.9 & 77.5 & 94.2
& \cellcolor{red!27}73.8
& \cellcolor{green!27}93.3
& \textcolor{red!80!black}{$-19.5$}
\\

X-VLA \citep{zheng2025xvla}
& 41.7 & 59.7 & 93.6 & 70.1 & 70.1 & 95.0
& \cellcolor{red!20}66.0
& \cellcolor{green!28}95.0
& \textcolor{red!80!black}{$-29.0$}
\\

GR00T N1.7 \citep{bjorck2025gr00t}
& 41.7 & 66.0 & 77.9 & 82.4 & 60.3 & 89.6
& \cellcolor{red!18}63.9
& \cellcolor{green!25}90.4
& \textcolor{red!80!black}{$-26.6$}
\\

MolmoAct2 \citep{fang2026molmoact2}
& 58.6 & 74.2 & 90.3 & 90.4 & 79.0 & 94.2
& \cellcolor{red!30}76.9
& \cellcolor{green!29}97.2
& \textcolor{red!80!black}{$-20.3$}
\\

\specialrule{0.9pt}{1.2pt}{0pt}

\rowcolor{gray!15}
\multicolumn{10}{c}{\textit{World-Action Models}}
\\
\specialrule{0.7pt}{0pt}{0pt}

Cosmos Policy \citep{kim2026cosmospolicy}
& 45.6 & 84.7 & 87.9 & 85.6 & 70.6 & 95.8
& \cellcolor{red!25}72.2
& \cellcolor{green!29}97.5
& \textcolor{red!80!black}{$-25.3$}
\\

FastWAM \citep{yuan2026fastwam}
& 43.6 & 67.1 & 90.7 & 89.9 & 71.4 & 93.2
& \cellcolor{red!23}70.1
& \cellcolor{green!30}97.8
& \textcolor{red!80!black}{$-27.6$}
\\

LingBot-VA \citep{li2026lingbotva}
& 77.3 & 81.5 & 90.5 & 96.8 & 59.3 & 96.7
& \cellcolor{red!30}81.8
& \cellcolor{green!30}98.5
& \textcolor{red!80!black}{$-16.7$}
\\

\specialrule{0.9pt}{1.2pt}{0pt}

\rowcolor{gray!15}
\multicolumn{10}{c}{\textit{Robustness-Oriented Policies}}
\\
\specialrule{0.7pt}{0pt}{0pt}

Anchor-Align \citep{dalal2026generalizable}
& 42.4 & 47.4 & 77.4 & 87.5 & 58.8 & 88.3
& \cellcolor{red!17}61.6
& \cellcolor{green!26}91.3
& \textcolor{red!80!black}{$-29.7$}
\\

RIPT-VLA \citep{tan2025interactive}
& 39.4 & 58.6 & 74.5 & 79.6 & 54.3 & 89.6
& \cellcolor{red!15}60.1
& \cellcolor{green!23}87.9
& \textcolor{red!80!black}{$-27.8$}
\\

\specialrule{1.3pt}{0pt}{0pt}

\end{tabular}%
}

\vspace{-10pt}
\end{table*}

\subsection{Experimental Setup}
\label{sec}
We evaluate 17 representative robot manipulation policies spanning different policy paradigms, using their official checkpoints and inference settings.
Experiments are conducted on the 8 held-out tasks of LIBERO-Pro2.0 under nominal execution, 22 static shifts, and 20 runtime shifts.
Additional experimental details are provided in Appendix~\ref{app:perturbation_catalogue}. Our evaluation is organized around the following research questions:

\begin{itemize}[leftmargin=5mm]

    \item \textbf{Overall Performance Landscape:}
    How robust are current robot policies under diverse static and runtime distribution shifts?
    
    \item \textbf{RQ1:}
    What limits robustness under runtime distribution shifts?
    
    \item \textbf{RQ2:}
    Does robustness compose under compound distribution shifts?
    
    \item \textbf{RQ3:}
    Can robustness be learned from limited demonstrations and transferred to held-out tasks?

\end{itemize}

\subsection{Overall Performance Landscape}
\label{sec:overall_landscape}

\paragraph{Finding 1: High nominal performance does not imply strong robustness.}
Table~\ref{tab:static_leaderboard} and Table~\ref{tab:runtime_leaderboard} show that, despite nominal success rates ranging from 63.8\% to 98.5\%, all policies degrade substantially under distribution shifts.
Static shifts reduce average success rates to 33.9\%–71.2\% ($25.1$--$41.8$ points drop), while runtime shifts reduce them to 35.3\%--81.8\% ($16.7$--$32.5$ points drop).
Moreover, robustness rankings do not consistently follow nominal performance.
For example, FastWAM achieves higher nominal performance (97.8\%) than $\pi_{0.5}$ (93.3\%), but its runtime robustness decreases to 70.1\%, lower than $\pi_{0.5}$ (73.8\%). This ranking reversal indicates that nominal task success does not necessarily predict robustness under runtime distribution shifts. Therefore, robustness evaluation reveals additional capabilities beyond standard task solving metrics.

\begin{figure*}[t]
    \centering
    \includegraphics[width=\textwidth]{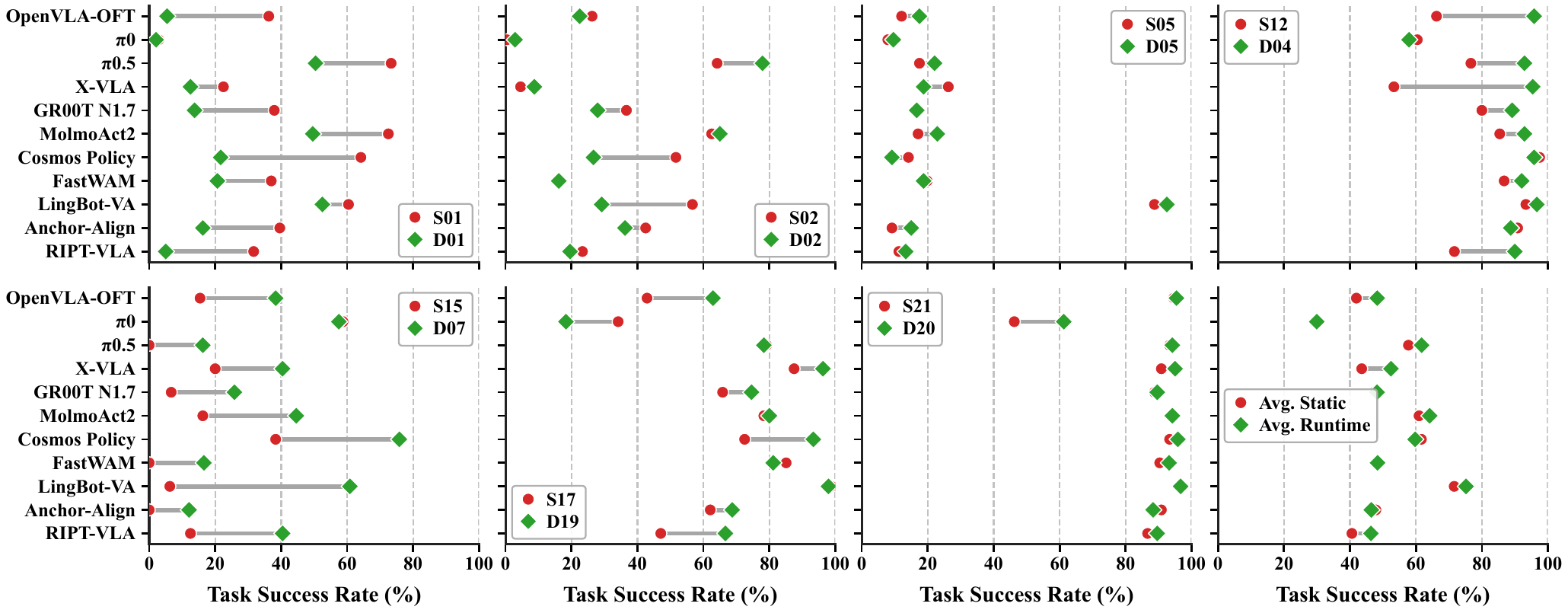}
    \caption{
Comparison between matched static and runtime perturbations.
Each pair contains a static shift occurring before execution and its runtime counterpart triggered during execution, including object pose changes (S01/D01, S02/D02), path constraints (S05/D05), occlusion (S12/D04), camera changes (S15/D07), robot-state variations (S17/D19), and language variations (S21/D20).
}
    \label{fig:static_runtime_comparison}
    \vspace{-5pt}
\end{figure*}

\textbf{Finding 2: Static evaluation does not fully characterize
runtime robustness.}
Figure~\ref{fig:static_runtime_comparison} compares semantically matched static--runtime perturbation pairs, including object relocation, goal relocation, path obstruction, occlusion, camera changes, and language variation.
Despite similar perturbation mechanisms, their effects differ across static and runtime settings.
For example, averaged across the four suites, LingBot-VA achieves
56.7\% success under static receiver-pose changes (S02), compared
with 29.2\% under runtime receiver-pose updates (D02).
Camera changes exhibit the opposite pattern: 6.3\% under S15
versus 60.8\% under D07.
These contrasting patterns show that static evaluation alone
does not fully characterize runtime robustness.

\paragraph{Finding 3: Different perturbation categories expose heterogeneous robustness bottlenecks.}
Robustness varies substantially across perturbation dimensions.
Language variations remain relatively easy, with all models achieving over 46\% success under S21, whereas execution-related shifts are considerably more challenging, reducing OpenVLA-OFT and $\pi_0$ to 0.0\% and 5.0\% under S20.
Runtime shifts further expose distinct weaknesses in observation, execution, and physical-state adaptation, showing that robustness is inherently multidimensional.

\subsection{RQ1: What Limits Robustness under Runtime Distribution Shifts?}
\label{sec:rq1}

\begin{table}[htb]
\centering
\vspace{-10pt}
\caption{
Feedback utilization gain $G_{\mathrm{feedback}}$ (pp) across dynamic perturbation categories.
}
\label{tab:rq1a_feedback_category}

\small
\renewcommand{\arraystretch}{1.25}
\setlength{\tabcolsep}{5.5pt}

\begin{tabular}{cccccccc}
\toprule

\textbf{Model}
&
\begin{tabular}[c]{@{}c@{}}
\textbf{Cond.}\\[-1pt]
{\scriptsize\textit{D01--D06}}
\end{tabular}
&
\begin{tabular}[c]{@{}c@{}}
\textbf{Env.}\\[-1pt]
{\scriptsize\textit{D07--D09}}
\end{tabular}
&
\begin{tabular}[c]{@{}c@{}}
\textbf{Obs.}\\[-1pt]
{\scriptsize\textit{D10--D13}}
\end{tabular}
&
\begin{tabular}[c]{@{}c@{}}
\textbf{Exec.}\\[-1pt]
{\scriptsize\textit{D14--D16}}
\end{tabular}
&
\begin{tabular}[c]{@{}c@{}}
\textbf{Robot}\\[-1pt]
{\scriptsize\textit{D17--D19}}
\end{tabular}
&
\begin{tabular}[c]{@{}c@{}}
\textbf{Lang.}\\[-1pt]
{\scriptsize\textit{D20}}
\end{tabular}
&
\textbf{Avg.}
\\

\midrule

OpenVLA-OFT
& \cellcolor{red!26}-26.6
& \cellcolor{red!23}-22.8
& \cellcolor{red!17}-15.0
& \cellcolor{red!12}-8.7
& \cellcolor{red!27}-27.3
& \cellcolor{red!6}-1.1
& \cellcolor{red!21}-19.8
\\

$\pi_{0.5}$
& \cellcolor{red!8}-3.5
& \cellcolor{red!20}-19.2
& \cellcolor{red!8}-3.2
& \cellcolor{red!9}-5.3
& \cellcolor{green!15}+7.5
& \cellcolor{green!9}+2.1
& \cellcolor{red!9}-4.1
\\

Anchor-Align
& \cellcolor{red!17}-15.1
& \cellcolor{red!20}-19.5
& \cellcolor{red!15}-12.9
& \cellcolor{red!13}-10.8
& \cellcolor{red!12}-8.9
& \cellcolor{red!11}-7.9
& \cellcolor{red!16}-13.4
\\

FastWAM
& \cellcolor{red!18}-15.6
& \cellcolor{red!12}-9.2
& \cellcolor{red!6}-1.4
& \cellcolor{red!11}-7.5
& \cellcolor{red!10}-6.8
& \cellcolor{red!8}-3.3
& \cellcolor{red!12}-8.7
\\

Cosmos Policy
& \cellcolor{red!14}-11.8
& \cellcolor{red!13}-10.3
& \cellcolor{red!10}-6.4
& \cellcolor{red!15}-13.2
& \cellcolor{green!7}+1.1
& \cellcolor{red!9}-5.4
& \cellcolor{red!12}-8.4
\\

\bottomrule
\end{tabular}

\vspace{-10pt}
\end{table}

We investigate why policies fail under runtime distribution shifts through three controlled interventions targeting feedback utilization, closed-loop update frequency, and post-perturbation recovery.
\paragraph{RQ1-A: Can Policies Use Updated Observations?}
We examine whether policies can exploit observations updated after runtime perturbations.
Under \emph{live} feedback, the policy continuously receives post-perturbation visual observations, whereas under \emph{freeze} feedback, it repeatedly receives the last observation before the perturbation and therefore does not receive updated visual observations.
Here, $S_1$ denotes the success rate under a dynamic perturbation, while $S_0$ denotes the success rate under the corresponding no-perturbation condition.
Therefore, $S_{1,\mathrm{live}}$ and $S_{1,\mathrm{freeze}}$ measure performance with and without access to updated observations after a perturbation, whereas $S_{0,\mathrm{live}}$ and $S_{0,\mathrm{freeze}}$ capture the normal feedback effect in the absence of perturbations.
We define
\[
G_{\mathrm{feedback}}
=
(S_{1,\mathrm{live}}-S_{1,\mathrm{freeze}})
-
(S_{0,\mathrm{live}}-S_{0,\mathrm{freeze}}),
\]
where positive values indicate a larger live-versus-freeze benefit
under perturbation than in the corresponding no-perturbation condition.
Table~\ref{tab:rq1a_feedback_category} summarizes the results, with complete results in Appendix~\ref{app:rq1a_full}. \textbf{(1) Reduced feedback benefit under perturbations.}
Negative values indicate that the benefit of updated visual
observations is smaller than under nominal execution, not that
updated observations provide no benefit.
For example, under D01, OpenVLA-OFT achieves 5.4\% success with live
observations and 0.0\% with frozen observations, despite
$G_{\mathrm{feedback}}=-46.1$ percentage points. \textbf{(2) Feedback benefit depends on perturbation type.}
Although most category-level contrasts are negative, $\pi_{0.5}$
and Cosmos Policy obtain positive average contrasts of 7.5 and
1.1 percentage points, respectively, on robot-state changes
(D17--D19).

\begin{figure*}[t]
    \centering
    \vspace{-5pt}
    \includegraphics[width=0.99\textwidth]{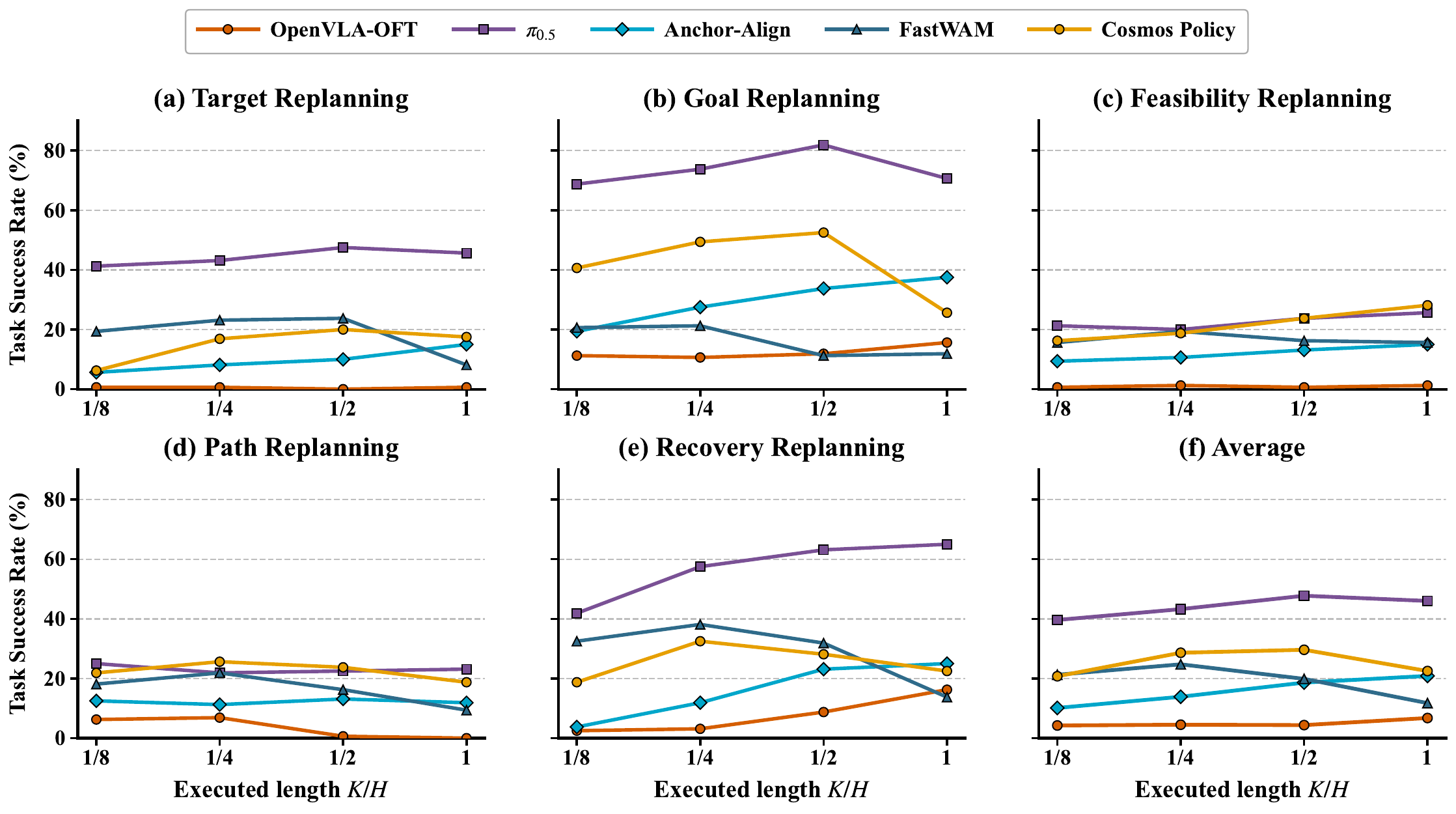}
    \vspace{-10pt}
    \caption{
Effect of policy execution horizon on robustness to dynamic perturbations. Panels (a)--(e) report task success rates under five representative perturbation types; panel (f) shows the average across these perturbations.
}
    \label{fig:RQ1-B}
\vspace{-10pt}
\end{figure*}

\paragraph{RQ1-B: Does Closed-Loop Update Frequency Limit Robustness?}
\label{sec:rq1b}

We study whether action chunking delays the use of updated observations by fixing the prediction horizon \(H\) and varying the number of executed actions \(K\) before re-observation. Detailed horizon settings are shown in Appendix~\ref{app:evaluation_protocol}. We evaluate target replanning (D01), goal replanning (D02), feasibility replanning (D03), path replanning (D05), and recovery replanning (D18) under execution ratios \(K/H\). Figure~\ref{fig:RQ1-B} presents the results. \textbf{(1) Replanning frequency exhibits a non-monotonic effect.} Average success increases from \(21.6\%\) at \(K/H=1\) to \(24.1\%\) at \(1/2\), but drops to \(19.2\%\) at \(1/8\), showing that more frequent re-observation does not uniformly
improve robustness. \textbf{(2) Frequency sensitivity is policy-dependent.} FastWAM and Cosmos Policy benefit more from shorter execution intervals, whereas Anchor-Align performs better with longer horizons. OpenVLA-OFT remains weak across settings, indicating that update frequency alone cannot compensate for limited perturbation handling.

\begin{wraptable}{r}{0.6\columnwidth}
\vspace{-20pt}
\centering
\caption{
Average performance and intervention gains in RQ1-C.
$\Delta$ denotes the change in success rate relative to normal continuation.
}
\label{tab:rq1c_summary}

\small
\renewcommand{\arraystretch}{1.20}
\setlength{\tabcolsep}{4.5pt}

\resizebox{0.59\columnwidth}{!}{
\begin{tabular}{ccccc}
\toprule

\textbf{Model}
& \textbf{Continue}
& $\boldsymbol{\Delta_{\mathrm{Hold}}}$
& $\boldsymbol{\Delta_{\mathrm{Retry}}}$
& $\boldsymbol{\Delta_{\mathrm{Reset}}}$
\\

\midrule

OpenVLA-OFT
& 64.6
& \cellcolor{red!14}-15.7
& \cellcolor{red!30}-40.8
& \cellcolor{red!24}-31.6
\\

$\pi_{0.5}$
& 73.8
& \cellcolor{green!18}+18.9
& \cellcolor{red!20}-24.1
& \cellcolor{green!8}+0.7
\\

Anchor-Align
& 61.6
& \cellcolor{green!26}+28.1
& \cellcolor{red!15}-16.8
& \cellcolor{green!8}+0.9
\\

FastWAM
& 70.1
& \cellcolor{green!22}+24.3
& \cellcolor{red!18}-20.9
& \cellcolor{red!6}-0.4
\\

Cosmos Policy
& 72.2
& \cellcolor{green!24}+26.0
& \cellcolor{red!19}-23.0
& \cellcolor{green!10}+2.5
\\

\bottomrule
\end{tabular}
}

\vspace{-8pt}
\end{wraptable}

\paragraph{RQ1-C: How Do Interventions Affect Recovery after a Runtime Shift?}
\label{sec:rq1c}
We compare three recovery interventions following runtime shifts:
\emph{Hold} pauses execution,
\emph{Retry} replays actions from an unperturbed reference trajectory,
and \emph{Reset} restores the robot to its pre-perturbation pose.
The baseline, \emph{Continue}, proceeds under the perturbed condition
without an additional recovery intervention.
For each intervention $I$, we report the success-rate change
$\Delta_I=S_I-S_{\mathrm{continue}}$ in percentage points.
Table~\ref{tab:rq1c_summary} summarizes the average effects,
with per-perturbation results in Appendix~\ref{app:rq1c_full}. \textbf{(1) Hold has policy-dependent effects.}
Hold increases average success by 18.9--28.1 percentage points
for four of the five policies, but decreases it by 15.7 points
for OpenVLA-OFT.
\textbf{(2) Retry reduces success on average.}
Retry lowers average success by 16.8--40.8 percentage points
across all five policies, although it improves performance
under some individual perturbations.
\textbf{(3) Reset has small average effects for most policies.}
The average change ranges from $-0.4$ to $+2.5$ percentage points
for $\pi_{0.5}$, Anchor-Align, FastWAM, and Cosmos Policy,
whereas OpenVLA-OFT declines by 31.6 points.

\subsection{RQ2: Does Robustness Compose?}
\label{sec:rq2}

\begin{figure*}[t]
    \centering
    \includegraphics[width=0.99\textwidth]{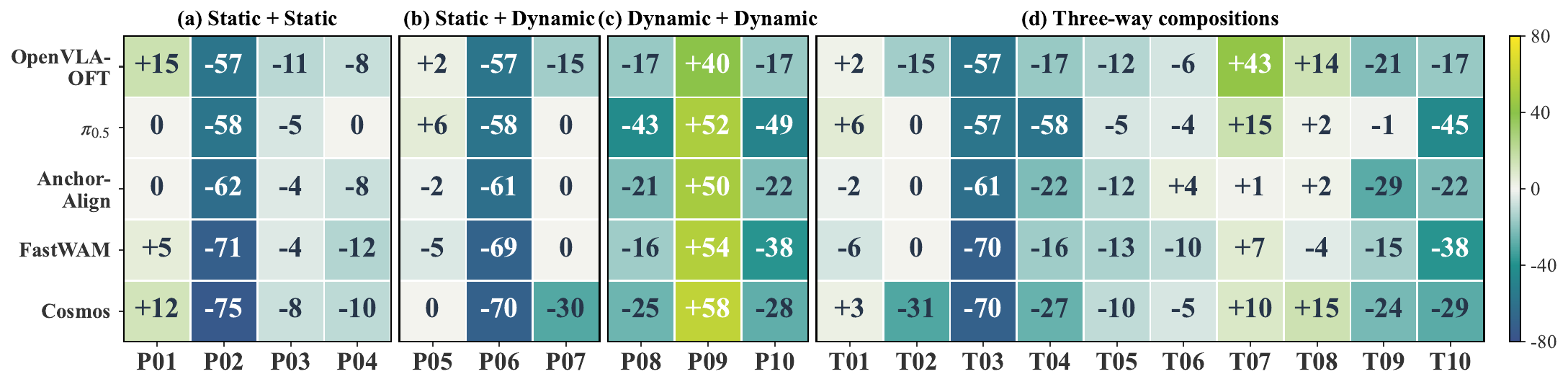}
    \vspace{-10pt}
    \caption{
        Weakest-single gap $\Delta_m(\mathcal{C})$ under composed perturbations.
        The heatmap compares static--static, static--dynamic, dynamic--dynamic, and three-way compositions across policies.
    }
    \label{fig:rq2_gap}
\vspace{-5pt}
\end{figure*}

Real-world perturbations often occur jointly, raising the question of whether robustness to individual shifts transfers to compound settings. For a model $m$ and perturbation composition $\mathcal{C}=\{c_1,\dots,c_k\}$, we define the weakest-single gap as $\Delta_m(\mathcal{C})=p_m(\mathcal{C})-\min_{c_i\in\mathcal{C}}p_m(c_i)$. A negative value indicates lower success than the weakest constituent
perturbation; it does not by itself establish an interaction effect
beyond independent failure accumulation. Figure~\ref{fig:rq2_gap} summarizes these gaps, while Appendix~\ref{app:rq2_full} reports detailed success rates.

\paragraph{Robustness to individual perturbations does not reliably compose.}
Table~\ref{tab:rq2_key_evidence} shows that P02 (\texttt{S09+S13}), P06 (\texttt{S09+D13}), and T03 (\texttt{S09+S22+D13}) reduce success to only $9.2\%$--$12.9\%$, although their constituent perturbations remain individually manageable ($69.6\%$--$99.2\%$). Thus, robustness to isolated shifts can collapse when multiple assumptions are violated jointly.

\paragraph{Recovery capability is also non-compositional.}
As shown in Appendix~\ref{app:rq2_full}, $\pi_{0.5}$ performs well on D02, D18, and D20 individually, yet drops to $22.9\%$ on P08 (\texttt{D02+D18}) and $17.5\%$ on P10 (\texttt{D18+D20}). This indicates that recovery from one runtime disturbance does not
guarantee recovery under concurrent shifts.

\begin{table}[htb]
\centering
\small
\vspace{-15pt}
\caption{
Key evidence for non-compositional robustness. Although all models perform reasonably well on the constituent perturbations S09, S13, D13, and S22, they all collapse on the composed settings P02, P06, and T03.
}
\label{tab:rq2_key_evidence}

\renewcommand{\arraystretch}{1.20}
\setlength{\tabcolsep}{4.5pt}

\resizebox{\columnwidth}{!}{
\begin{tabular}{cccccccc}
\toprule

\textbf{Model}
& \textbf{S09}
& \textbf{S13}
& \textbf{D13}
& \textbf{S22}
& \textbf{P02: S09+S13}
& \textbf{P06: S09+D13}
& \textbf{T03: S09+S22+D13}
\\

\midrule

OpenVLA-OFT
& \cellcolor{green!18}69.6
& \cellcolor{green!28}94.2
& \cellcolor{green!18}69.6
& \cellcolor{green!27}93.3
& \cellcolor{red!18}\textbf{12.5}
& \cellcolor{red!18}\textbf{12.5}
& \cellcolor{red!18}\textbf{12.5}
\\

$\pi_{0.5}$
& \cellcolor{green!18}70.0
& \cellcolor{green!27}92.5
& \cellcolor{green!23}83.8
& \cellcolor{green!28}94.2
& \cellcolor{red!18}\textbf{12.5}
& \cellcolor{red!17}\textbf{12.1}
& \cellcolor{red!19}\textbf{12.9}
\\

Anchor-Align
& \cellcolor{green!19}72.5
& \cellcolor{green!24}86.7
& \cellcolor{green!22}79.6
& \cellcolor{green!25}87.9
& \cellcolor{red!14}\textbf{10.0}
& \cellcolor{red!16}\textbf{11.3}
& \cellcolor{red!17}\textbf{11.7}
\\

FastWAM
& \cellcolor{green!22}80.3
& \cellcolor{green!26}91.2
& \cellcolor{green!23}83.7
& \cellcolor{green!25}88.3
& \cellcolor{red!13}\textbf{9.2}
& \cellcolor{red!17}\textbf{11.7}
& \cellcolor{red!14}\textbf{10.0}
\\

Cosmos Policy
& \cellcolor{green!25}87.5
& \cellcolor{green!30}99.2
& \cellcolor{green!22}81.7
& \cellcolor{green!29}96.7
& \cellcolor{red!19}\textbf{12.9}
& \cellcolor{red!17}\textbf{11.7}
& \cellcolor{red!17}\textbf{12.1}
\\

\bottomrule
\end{tabular}
}
\vspace{-15pt}
\end{table}

\subsection{RQ3: Can Robustness Be Learned and Generalized?}
\label{sec:rq3}
\begin{table}[t]
\centering
\vspace{-10pt}
\caption{
Transfer of robustness from training set to tasks held out from additional fine-tuning.
Each column reports the average success rate over a static--dynamic perturbation pair.
\textit{Gain} denotes the improvement of Ours over $\pi_{0.5}$.
}
\label{tab:rq3_transfer}

\small
\renewcommand{\arraystretch}{1.20}
\setlength{\tabcolsep}{5.0pt}

\resizebox{\linewidth}{!}{
\begin{tabular}{ccccccc}

\toprule

\textbf{Model}
& \textbf{Object Pose}
& \textbf{Receiver Pose}
& \textbf{Path Obstacle}
& \textbf{Camera Calibration}
& \textbf{Robot State}
& \textbf{Total}
\\

\midrule

OpenVLA-OFT
& 20.8
& 24.4
& 14.8
& 26.9
& 52.9
& 28.0
\\

$\pi_{0.5}$
& 61.9
& 71.0
& 19.8
& 8.1
& 78.7
& 47.9
\\

Anchor-Align
& 27.9
& 39.4
& 12.1
& 6.0
& 65.4
& 30.2
\\

FastWAM
& 28.8
& 16.0
& 19.3
& 8.3
& 83.1
& 31.1
\\

Cosmos Policy
& 42.9
& 39.2
& 11.7
& \textbf{57.1}
& 82.9
& 46.7
\\

\midrule

\textbf{Ours}
& \textbf{68.8}
& \textbf{73.2}
& \textbf{96.3}
& 20.8
& \textbf{91.7}
& \textbf{70.1}
\\

\textit{Gain}
& \cellcolor{green!14}+6.9
& \cellcolor{green!10}+2.1
& \cellcolor{green!30}+76.5
& \cellcolor{green!18}+12.7
& \cellcolor{green!18}+12.9
& \cellcolor{green!22}+22.2
\\

\bottomrule

\end{tabular}
}
\vspace{-15pt}
\end{table}

We ask whether robustness can be learned from limited demonstrations and transferred to tasks held out from additional fine-tuning. We train $\pi_{0.5}$ on eight training tasks with ten trajectories per task, while keeping all evaluation tasks disjoint. We consider five semantically matched static--dynamic perturbation families: S01/D01 (object pose), S02/D02 (receiver pose), S05/D05 (path obstacle), S15/D07 (camera calibration), and S17/D19 (robot state). Training details are provided in Appendix ~\ref{app:train_setup}.

\paragraph{Robustness can be learned from limited demonstrations.}
After robustness-oriented fine-tuning, our model achieves $70.1\%$ average robustness on unseen tasks, outperforming existing policies ($28.0\%$--$47.9\%$) in Table~\ref{tab:rq3_transfer}. The largest gain appears on path changes (S05/D05), where our model reaches $96.3\%$, compared with below $20\%$ for all baselines.

\paragraph{Generalization depends on perturbation semantics.}
Learned robustness transfers well in receiver pose ($73.2\%$) and robot-state shifts ($91.7\%$), but remains limited under camera calibration shifts ($20.8\%$). This suggests robustness generalizes better when perturbation semantics preserve stable task constraints than when they require broader perception--action recalibration.
\section{Related Work}
\label{sec:related_work}
\noindent\textbf{Embodied Foundation Models.}
Recent advances in large-scale robot datasets and foundation models have significantly improved the generalization capabilities of robotic manipulation policies. Datasets such as Open X-Embodiment and DROID provide diverse demonstrations across tasks, environments, and robotic platforms~\citep{openx2023,khazatsky2024droid}. Building on these resources, vision-language-action (VLA) models, including OpenVLA, $\pi_0$, and OpenVLA-OFT, learn generalizable visuomotor policies from large-scale multimodal data~\citep{kim2024openvla,black2025pi0,kim2025openvlaoft}. More recently, world-action models such as Cosmos Policy, LingBot-VA, and FastWAM have explored predictive world representations to support action generation and visuomotor control~\citep{kim2026cosmospolicy,li2026lingbotva,yuan2026fastwam}.

\noindent\textbf{Robot Manipulation Benchmarks and Robustness Evaluation.}
Existing benchmarks, including Meta-World, RLBench, CALVIN, ManiSkill2, and LIBERO, provide standardized protocols for evaluating multi-task and language-conditioned robotic manipulation~\citep{yu2020metaworld,james2020rlbench,mees2022calvin,gu2023maniskill2,liu2023libero}. Recent benchmarks such as RoboCasa, RoboTwin, and VLABench further expand task complexity and scene diversity~\citep{nasiriany2024robocasa,mu2025robotwin,zhang2025vlabench}, while RoboDojo introduces a unified simulation-and-real-world evaluation framework covering diverse manipulation capabilities~\citep{chen2026robodojo}. Beyond nominal task success, recent studies investigate policy robustness through environmental variations, domain randomization, and controlled perturbations~\citep{li2024simplerenv,chen2025robotwin2,zhang2026wamrobustness}. In particular, LIBERO-PRO exposes the reliance of VLA models on memorized behaviors under changes in objects, initial states, instructions, and environments~\citep{zhou2025liberopro}, while LIBERO-Plus evaluates robustness under progressively challenging static distribution shifts~\citep{fei2026liberoplus}. Despite these advances, existing benchmarks primarily emphasize task diversity or robustness to pre-execution variations, leaving adaptation to disturbances arising during execution insufficiently evaluated. LIBERO-Pro2.0 addresses this critical gap by systematically assessing both static generalization and runtime robustness within a unified benchmark.
\section{Conclusion}
\label{sec:conclusion}

We present LIBERO-Pro2.0, a systematic benchmark for evaluating the robustness of embodied foundation models under static distribution shifts and runtime interventions. Our evaluation reveals substantial robustness gaps that remain hidden by strong nominal performance, while further analyses highlight challenges in online adaptation, recovery, compound robustness, and robustness transfer. We hope LIBERO-Pro2.0 provides a practical foundation for developing embodied policies that remain reliable under diverse and evolving deployment conditions.

\clearpage

\bibliography{iclr2027_conference}
\bibliographystyle{iclr2027_conference}

\appendix
\clearpage
\begingroup
\raggedbottom
\setlength{\floatsep}{10pt plus 2pt minus 2pt}
\setlength{\textfloatsep}{12pt plus 2pt minus 2pt}
\setlength{\intextsep}{10pt plus 2pt minus 2pt}

\section{Additional Experimental Results}
\label{sec:additional_results}

This appendix provides the complete LIBERO-Pro2.0 results supplementing the
feedback-utilization, recovery-intervention, and robustness-composition analyses
in the main paper.

\subsection{Main Results}

\begin{table*}[htb]
\centering
\caption{Static and runtime perturbation results on LIBERO-Goal. All values are reported in percentages (\%).}
\label{tab:libero_goal_full}
\renewcommand{\arraystretch}{1.15}
\setlength{\tabcolsep}{2.8pt}

\textbf{(a) Static Perturbations}
\vspace{1.5mm}

\resizebox{\textwidth}{!}{%
\begin{tabular}{lcccccccccccccccccccccc}
\toprule
\textbf{Model} & \textbf{S1} & \textbf{S2} & \textbf{S3} & \textbf{S4} & \textbf{S5} & \textbf{S6} & \textbf{S7} & \textbf{S8} & \textbf{S9} & \textbf{S10} & \textbf{S11} & \textbf{S12} & \textbf{S13} & \textbf{S14} & \textbf{S15} & \textbf{S16} & \textbf{S17} & \textbf{S18} & \textbf{S19} & \textbf{S20} & \textbf{S21} & \textbf{S22} \\
\midrule
OpenVLA & 1.7 & 8.3 & 0.0 & 0.0 & 0.0 & 48.3 & 35.0 & 33.3 & 43.3 & 46.7 & 41.7 & 0.0 & 31.7 & 16.7 & 1.7 & 0.0 & 15.0 & 51.7 & 46.7 & 0.0 & 63.3 & 43.3 \\
OpenVLA-OFT & 33.3 & 0.0 & 0.0 & 0.0 & 0.0 & 86.7 & 58.3 & 83.3 & 81.7 & 91.7 & 85.0 & 55.0 & 90.0 & 93.3 & 0.0 & 8.3 & 0.0 & 81.7 & 90.0 & 0.0 & 88.3 & 88.3 \\
OpenVLA-OFT$_{\mathrm{w}}$ & 20.0 & 1.7 & 0.0 & 0.0 & 0.0 & 76.7 & 55.0 & 86.7 & 81.7 & 86.7 & 90.0 & 0.0 & 78.3 & 75.0 & 43.3 & 0.0 & 40.0 & 76.7 & 86.7 & 0.0 & 85.0 & 83.3 \\
OpenVLA-OFT$_{\mathrm{m}}$ & 40.0 & 1.7 & 0.0 & 0.0 & 0.0 & 76.7 & 55.0 & 73.3 & 71.7 & 80.0 & 56.7 & 53.3 & 85.0 & 86.7 & 0.0 & 1.7 & 0.0 & 75.0 & 86.7 & 0.0 & 80.0 & 85.0 \\
OpenVLA-OFT+ & 28.3 & 1.7 & 0.0 & 5.0 & 0.0 & 60.0 & 45.0 & 60.0 & 56.7 & 55.0 & 63.3 & 65.0 & 76.7 & 88.3 & 0.0 & 0.0 & 0.0 & 83.3 & 78.3 & 0.0 & 80.0 & 81.7 \\
UniVLA & 3.3 & 5.0 & 5.0 & 8.3 & 0.0 & 61.7 & 81.7 & 76.7 & 85.0 & 76.7 & 76.7 & 0.0 & 73.3 & 78.3 & 26.7 & 0.0 & 40.0 & 85.0 & 83.3 & 0.0 & 83.3 & 83.3 \\
$\pi_0$ & 3.3 & 0.0 & 0.0 & 0.0 & 0.0 & 48.3 & 51.7 & 58.3 & 56.7 & 55.0 & 45.0 & 63.3 & 65.0 & 43.3 & 50.0 & 51.7 & 30.0 & 1.7 & 51.7 & 13.3 & 40.0 & 51.7 \\
$\pi_0$-FAST & 55.0 & 36.7 & 8.3 & 8.3 & 0.0 & 71.7 & 88.3 & 71.7 & 76.7 & 86.7 & 63.3 & 78.3 & 93.3 & 88.3 & 0.0 & 25.0 & 43.3 & 50.0 & 90.0 & 0.0 & 93.3 & 91.7 \\
$\pi_{0.5}$ & 73.3 & 51.7 & 15.0 & 61.7 & 6.7 & 95.0 & 75.0 & 93.3 & 96.7 & 93.3 & 95.0 & 66.7 & 96.7 & 95.0 & 0.0 & 0.0 & 50.0 & 93.3 & 95.0 & 0.0 & 96.7 & 98.3 \\
X-VLA & 23.3 & 0.0 & 0.0 & 0.0 & 0.0 & 93.3 & 93.3 & 96.7 & 100.0 & 95.0 & 85.0 & 46.7 & 93.3 & 96.7 & 0.0 & 0.0 & 65.0 & 28.3 & 98.3 & 48.3 & 95.0 & 95.0 \\
GR00T N1.7 & 21.7 & 1.7 & 30.0 & 0.0 & 0.0 & 73.3 & 75.0 & 85.0 & 81.7 & 86.7 & 73.3 & 71.7 & 83.3 & 80.0 & 0.0 & 0.0 & 30.0 & 85.0 & 90.0 & 0.0 & 81.7 & 83.3 \\
MolmoAct2 & 73.3 & 46.7 & 6.7 & 28.3 & 3.3 & 86.7 & 66.7 & 100.0 & 96.7 & 98.3 & 61.7 & 86.7 & 96.7 & 96.7 & 1.7 & 55.0 & 58.3 & 96.7 & 98.3 & 0.0 & 93.3 & 95.0 \\
Anchor-Align & 35.0 & 5.0 & 0.0 & 5.0 & 0.0 & 61.7 & 50.0 & 90.0 & 91.7 & 80.0 & 56.7 & 85.0 & 83.3 & 83.3 & 0.0 & 0.0 & 21.7 & 91.7 & 91.7 & 0.0 & 91.7 & 90.0 \\
RIPT-VLA & 20.0 & 0.0 & 0.0 & 0.0 & 0.0 & 73.3 & 48.3 & 55.0 & 58.3 & 65.0 & 66.7 & 55.0 & 60.0 & 66.7 & 0.0 & 0.0 & 1.7 & 56.7 & 71.7 & 0.0 & 63.3 & 61.7 \\
Cosmos Policy & 56.7 & 43.3 & 23.3 & 13.3 & 23.3 & 73.3 & 90.0 & 80.0 & 90.0 & 96.7 & 83.3 & 96.7 & 100.0 & 96.7 & 23.3 & 53.3 & 56.7 & 96.7 & 100.0 & 16.7 & 86.7 & 96.7 \\
FastWAM & 43.3 & 0.0 & 0.0 & 1.7 & 0.0 & 85.0 & 55.0 & 81.7 & 91.7 & 90.0 & 86.7 & 76.7 & 76.7 & 76.7 & 0.0 & 0.0 & 63.3 & 90.0 & 0.0 & 0.0 & 76.7 & 83.3 \\
LingBot-VA & 38.3 & 5.0 & 45.0 & 3.3 & 78.3 & 58.3 & 46.7 & 88.3 & 48.3 & 86.7 & 90.0 & 90.0 & 88.3 & 93.3 & 0.0 & 0.0 & 96.7 & 95.0 & 66.7 & 73.3 & 90.0 & 91.7 \\
\bottomrule
\end{tabular}%
}

\vspace{2mm}

\textbf{(b) Runtime Perturbations}
\vspace{1.5mm}

\resizebox{\textwidth}{!}{%
\begin{tabular}{lcccccccccccccccccccc}
\toprule
\textbf{Model} & \textbf{D1} & \textbf{D2} & \textbf{D3} & \textbf{D4} & \textbf{D5} & \textbf{D6} & \textbf{D7} & \textbf{D8} & \textbf{D9} & \textbf{D10} & \textbf{D11} & \textbf{D12} & \textbf{D13} & \textbf{D14} & \textbf{D15} & \textbf{D16} & \textbf{D17} & \textbf{D18} & \textbf{D19} & \textbf{D20} \\
\midrule
OpenVLA & 1.7 & 23.3 & 26.7 & 11.7 & 16.7 & 48.3 & 0.0 & 30.0 & 40.0 & 26.7 & 20.0 & 13.3 & 6.7 & 16.7 & 35.0 & 25.0 & 40.0 & 0.0 & 11.7 & 48.3 \\
OpenVLA-OFT & 8.3 & 0.0 & 50.0 & 88.3 & 30.0 & 85.0 & 8.3 & 25.0 & 85.0 & 71.7 & 68.3 & 60.0 & 50.0 & 38.3 & 86.7 & 86.7 & 86.7 & 1.7 & 18.3 & 91.7 \\
OpenVLA-OFT$_{\mathrm{w}}$ & 5.0 & 8.3 & 48.3 & 48.3 & 25.0 & 88.3 & 70.0 & 58.3 & 91.7 & 70.0 & 66.7 & 41.7 & 45.0 & 46.7 & 83.3 & 78.3 & 81.7 & 0.0 & 46.7 & 83.3 \\
OpenVLA-OFT$_{\mathrm{m}}$ & 28.3 & 1.7 & 31.7 & 80.0 & 26.7 & 8.3 & 5.0 & 1.7 & 86.7 & 66.7 & 61.7 & 61.7 & 55.0 & 48.3 & 76.7 & 75.0 & 85.0 & 28.3 & 33.3 & 81.7 \\
OpenVLA-OFT+ & 21.7 & 6.7 & 36.7 & 80.0 & 23.3 & 25.0 & 1.7 & 18.3 & 83.3 & 68.3 & 61.7 & 66.7 & 61.7 & 55.0 & 81.7 & 86.7 & 90.0 & 26.7 & 36.7 & 86.7 \\
UniVLA & 1.7 & 11.7 & 46.7 & 31.7 & 16.7 & 81.7 & 53.3 & 40.0 & 85.0 & 50.0 & 63.3 & 40.0 & 33.3 & 21.7 & 80.0 & 78.3 & 60.0 & 13.3 & 55.0 & 83.3 \\
$\pi_0$ & 5.0 & 0.0 & 40.0 & 51.7 & 13.3 & 50.0 & 58.3 & 43.3 & 55.0 & 65.0 & 56.7 & 56.7 & 30.0 & 15.0 & 33.3 & 20.0 & 3.3 & 0.0 & 26.7 & 55.0 \\
$\pi_0$-FAST & 30.0 & 53.3 & 50.0 & 90.0 & 45.0 & 81.7 & 3.3 & 50.0 & 95.0 & 48.3 & 45.0 & 66.7 & 61.7 & 56.7 & 86.7 & 91.7 & 80.0 & 46.7 & 53.3 & 90.0 \\
$\pi_{0.5}$ & 38.3 & 90.0 & 50.0 & 100.0 & 40.0 & 96.7 & 0.0 & 50.0 & 98.3 & 86.7 & 86.7 & 86.7 & 85.0 & 88.3 & 93.3 & 93.3 & 93.3 & 78.3 & 56.7 & 93.3 \\
X-VLA & 33.3 & 0.0 & 50.0 & 93.3 & 28.3 & 88.3 & 0.0 & 58.3 & 96.7 & 95.0 & 93.3 & 96.7 & 71.7 & 91.7 & 85.0 & 3.3 & 85.0 & 5.0 & 96.7 & 93.3 \\
GR00T N1.7 & 23.3 & 20.0 & 50.0 & 75.0 & 36.7 & 65.0 & 0.0 & 63.3 & 88.3 & 65.0 & 46.7 & 55.0 & 55.0 & 56.7 & 76.7 & 80.0 & 71.7 & 13.3 & 48.3 & 81.7 \\
MolmoAct2 & 35.0 & 76.7 & 50.0 & 96.7 & 41.7 & 96.7 & 35.0 & 83.3 & 96.7 & 98.3 & 93.3 & 93.3 & 90.0 & 93.3 & 93.3 & 96.7 & 93.3 & 68.3 & 51.7 & 93.3 \\
Anchor-Align & 31.7 & 11.7 & 50.0 & 91.7 & 36.7 & 41.7 & 0.0 & 5.0 & 93.3 & 70.0 & 63.3 & 73.3 & 78.3 & 66.7 & 96.7 & 86.7 & 76.7 & 25.0 & 38.3 & 90.0 \\
RIPT-VLA & 3.3 & 0.0 & 50.0 & 70.0 & 15.0 & 61.7 & 8.3 & 5.0 & 65.0 & 50.0 & 51.7 & 45.0 & 48.3 & 38.3 & 66.7 & 55.0 & 58.3 & 0.0 & 21.7 & 66.7 \\
Cosmos Policy & 10.0 & 23.3 & 36.7 & 96.7 & 13.3 & 93.3 & 56.7 & 73.3 & 96.7 & 100.0 & 90.0 & 90.0 & 80.0 & 73.3 & 96.7 & 93.3 & 90.0 & 10.0 & 93.3 & 90.0 \\
FastWAM & 31.7 & 1.7 & 48.3 & 83.3 & 28.3 & 85.0 & 0.0 & 81.7 & 88.3 & 91.7 & 88.3 & 80.0 & 78.3 & 81.7 & 91.7 & 80.0 & 91.7 & 38.3 & 48.3 & 86.7 \\
LingBot-VA & 40.0 & 0.0 & 93.3 & 91.7 & 76.7 & 91.7 & 0.0 & 63.3 & 90.0 & 81.7 & 76.7 & 60.0 & 78.3 & 95.0 & 93.3 & 88.3 & 85.0 & 1.7 & 96.7 & 90.0 \\
\bottomrule
\end{tabular}%
}

\end{table*}

\begin{table*}[htb]
\centering
\caption{Static and runtime perturbation results on LIBERO-Object. All values are reported in percentages (\%).}
\label{tab:libero_object_full}
\renewcommand{\arraystretch}{1.15}
\setlength{\tabcolsep}{2.8pt}

\textbf{(a) Static Perturbations}
\vspace{1.5mm}

\resizebox{\textwidth}{!}{%
\begin{tabular}{lcccccccccccccccccccccc}
\toprule
\textbf{Model} & \textbf{S1} & \textbf{S2} & \textbf{S3} & \textbf{S4} & \textbf{S5} & \textbf{S6} & \textbf{S7} & \textbf{S8} & \textbf{S9} & \textbf{S10} & \textbf{S11} & \textbf{S12} & \textbf{S13} & \textbf{S14} & \textbf{S15} & \textbf{S16} & \textbf{S17} & \textbf{S18} & \textbf{S19} & \textbf{S20} & \textbf{S21} & \textbf{S22} \\
\midrule

OpenVLA
& 0.0 & 3.3 & 6.7 & 3.3 & 0.0 & 0.0 & 45.0 & 80.0 & 81.7 & 78.3 & 11.7 & 0.0 & 76.7 & 5.0 & 10.0 & 0.0 & 36.7 & 88.3 & 88.3 & 0.0 & 81.7 & 75.0 \\

OpenVLA-OFT
& 8.3 & 1.7 & 0.0 & 16.7 & 1.7 & 0.0 & 95.0 & 91.7 & 60.0 & 98.3 & 80.0 & 70.0 & 98.3 & 100.0 & 10.0 & 20.0 & 51.7 & 100.0 & 96.7 & 0.0 & 96.7 & 100.0 \\

OpenVLA-OFT$_{\mathrm{w}}$
& 0.0 & 0.0 & 1.7 & 0.0 & 0.0 & 0.0 & 95.0 & 98.3 & 100.0 & 100.0 & 61.7 & 0.0 & 98.3 & 86.7 & 50.0 & 45.0 & 96.7 & 15.0 & 100.0 & 0.0 & 100.0 & 98.3 \\

OpenVLA-OFT$_{\mathrm{m}}$
& 43.3 & 6.7 & 1.7 & 3.3 & 0.0 & 0.0 & 76.7 & 93.3 & 40.0 & 98.3 & 66.7 & 98.3 & 86.7 & 93.3 & 3.3 & 5.0 & 86.7 & 100.0 & 98.3 & 0.0 & 98.3 & 100.0 \\

OpenVLA-OFT+
& 48.3 & 63.3 & 0.0 & 0.0 & 0.0 & 0.0 & 81.7 & 93.3 & 83.3 & 100.0 & 68.3 & 100.0 & 95.0 & 100.0 & 0.0 & 3.3 & 91.7 & 100.0 & 100.0 & 0.0 & 100.0 & 98.3 \\

UniVLA
& 5.0 & 3.3 & 3.3 & 35.0 & 0.0 & 0.0 & 78.3 & 91.7 & 100.0 & 86.7 & 26.7 & 0.0 & 100.0 & 51.7 & 0.0 & 0.0 & 66.7 & 98.3 & 100.0 & 0.0 & 96.7 & 93.3 \\

$\pi_0$
& 1.7 & 0.0 & 0.0 & 0.0 & 0.0 & 0.0 & 61.7 & 78.3 & 81.7 & 81.7 & 41.7 & 71.7 & 88.3 & 80.0 & 85.0 & 76.7 & 20.0 & 0.0 & 73.3 & 1.7 & 41.7 & 85.0 \\

$\pi_0$-FAST
& 56.7 & 65.0 & 0.0 & 1.7 & 5.0 & 0.0 & 83.3 & 96.7 & 100.0 & 95.0 & 83.3 & 100.0 & 100.0 & 95.0 & 0.0 & 20.0 & 93.3 & 95.0 & 91.7 & 0.0 & 98.3 & 93.3 \\

$\pi_{0.5}$
& 88.3 & 85.0 & 21.7 & 56.7 & 3.3 & 0.0 & 71.7 & 61.7 & 51.7 & 96.7 & 75.0 & 95.0 & 98.3 & 98.3 & 0.0 & 3.3 & 98.3 & 100.0 & 100.0 & 0.0 & 100.0 & 100.0 \\

X-VLA
& 10.0 & 0.0 & 0.0 & 5.0 & 18.3 & 0.0 & 78.3 & 88.3 & 100.0 & 98.3 & 65.0 & 80.0 & 100.0 & 98.3 & 0.0 & 18.3 & 96.7 & 25.0 & 98.3 & 91.7 & 100.0 & 100.0 \\

GR00T N1.7
& 30.0 & 45.0 & 5.0 & 0.0 & 3.3 & 0.0 & 66.7 & 93.3 & 95.0 & 96.7 & 68.3 & 98.3 & 100.0 & 95.0 & 0.0 & 0.0 & 86.7 & 100.0 & 98.3 & 0.0 & 98.3 & 96.7 \\

MolmoAct2
& 100.0 & 80.0 & 5.0 & 96.7 & 5.0 & 0.0 & 76.7 & 100.0 & 100.0 & 100.0 & 85.0 & 100.0 & 100.0 & 100.0 & 0.0 & 100.0 & 98.3 & 96.7 & 98.3 & 0.0 & 100.0 & 100.0 \\

Anchor-Align
& 1.7 & 50.0 & 0.0 & 0.0 & 0.0 & 0.0 & 58.3 & 100.0 & 96.7 & 33.3 & 68.3 & 96.7 & 91.7 & 100.0 & 0.0 & 0.0 & 95.0 & 100.0 & 96.7 & 0.0 & 96.7 & 100.0 \\

RIPT-VLA & 5.0 & 0.0 & 0.0 & 16.7 & 0.0 & 0.0 & 95.0 & 90.0 & 68.3 & 96.7 & 76.7 & 88.3 & 96.7 & 96.7 & 1.7 & 1.7 & 60.0 & 96.7 & 98.3 & 0.0 & 95.0 & 96.7 \\

Cosmos Policy
& 50.0 & 36.7 & 20.0 & 3.3 & 0.0 & 20.0 & 100.0 & 100.0 & 96.7 & 100.0 & 86.7 & 100.0 & 100.0 & 100.0 & 20.0 & 66.7 & 66.7 & 86.7 & 96.7 & 0.0 & 100.0 & 100.0 \\

FastWAM
& 39.7 & 7.7 & 0.0 & 1.7 & 0.0 & 0.0 & 81.7 & 99.0 & 100.0 & 100.0 & 81.7 & 100.0 & 100.0 & 100.0 & 0.0 & 0.0 & 98.3 & 100.0 & 38.0 & 0.0 & 100.0 & 100.0 \\

LingBot-VA
& 73.3 & 95.0 & 100.0 & 15.0 & 100.0 & 75.0 & 40.0 & 100.0 & 0.0 & 100.0 & 98.3 & 100.0 & 96.7 & 100.0 & 5.0 & 31.7 & 100.0 & 100.0 & 100.0 & 88.3 & 100.0 & 98.3 \\

\bottomrule
\end{tabular}%
}

\vspace{2mm}

\textbf{(b) Runtime Perturbations}
\vspace{1.5mm}

\resizebox{\textwidth}{!}{%
\begin{tabular}{lcccccccccccccccccccc}
\toprule
\textbf{Model} & \textbf{D1} & \textbf{D2} & \textbf{D3} & \textbf{D4} & \textbf{D5} & \textbf{D6} & \textbf{D7} & \textbf{D8} & \textbf{D9} & \textbf{D10} & \textbf{D11} & \textbf{D12} & \textbf{D13} & \textbf{D14} & \textbf{D15} & \textbf{D16} & \textbf{D17} & \textbf{D18} & \textbf{D19} & \textbf{D20} \\
\midrule

OpenVLA
& 0.0 & 1.7 & 0.0 & 18.3 & 3.3 & 66.7 & 15.0 & 6.7 & 81.7 & 46.7 & 51.7 & 40.0 & 56.7 & 50.0 & 85.0 & 68.3 & 63.3 & 0.0 & 40.0 & 80.0 \\

OpenVLA-OFT
& 0.0 & 1.7 & 0.0 & 98.3 & 0.0 & 100.0 & 0.0 & 76.7 & 98.3 & 100.0 & 98.3 & 95.0 & 91.7 & 80.0 & 95.0 & 100.0 & 91.7 & 50.0 & 90.0 & 95.0 \\

OpenVLA-OFT$_{\mathrm{w}}$
& 0.0 & 0.0 & 0.0 & 88.3 & 1.7 & 100.0 & 86.7 & 28.3 & 100.0 & 100.0 & 96.7 & 91.7 & 90.0 & 93.3 & 100.0 & 100.0 & 96.7 & 8.3 & 96.7 & 100.0 \\

OpenVLA-OFT$_{\mathrm{m}}$
& 25.0 & 16.7 & 0.0 & 100.0 & 1.7 & 100.0 & 3.3 & 70.0 & 100.0 & 98.3 & 96.7 & 86.7 & 90.0 & 85.0 & 98.3 & 100.0 & 95.0 & 60.0 & 93.3 & 100.0 \\

OpenVLA-OFT+
& 28.3 & 35.0 & 0.0 & 100.0 & 3.3 & 98.3 & 1.7 & 18.3 & 100.0 & 100.0 & 98.3 & 88.3 & 88.3 & 83.3 & 100.0 & 100.0 & 96.7 & 66.7 & 95.0 & 100.0 \\

UniVLA
& 1.7 & 1.7 & 0.0 & 18.3 & 0.0 & 95.0 & 58.3 & 90.0 & 96.7 & 98.3 & 98.3 & 95.0 & 88.3 & 91.7 & 96.7 & 91.7 & 88.3 & 30.0 & 86.7 & 96.7 \\

$\pi_0$
& 0.0 & 0.0 & 0.0 & 78.3 & 6.7 & 75.0 & 80.0 & 80.0 & 73.3 & 76.7 & 88.3 & 81.7 & 33.3 & 20.0 & 73.3 & 58.3 & 1.7 & 0.0 & 3.3 & 75.0 \\

$\pi_0$-FAST
& 38.3 & 50.0 & 0.0 & 100.0 & 13.3 & 100.0 & 0.0 & 26.7 & 98.3 & 83.3 & 83.3 & 81.7 & 80.0 & 76.7 & 100.0 & 96.7 & 95.0 & 63.3 & 65.0 & 98.3 \\

$\pi_{0.5}$
& 81.7 & 86.7 & 0.0 & 98.3 & 0.0 & 96.7 & 0.0 & 98.3 & 96.7 & 91.7 & 93.3 & 91.7 & 85.0 & 88.3 & 96.7 & 100.0 & 90.0 & 63.3 & 88.3 & 100.0 \\

X-VLA
& 1.7 & 0.0 & 0.0 & 100.0 & 3.3 & 95.0 & 13.3 & 1.7 & 100.0 & 98.3 & 96.7 & 100.0 & 93.3 & 98.3 & 95.0 & 58.3 & 88.3 & 33.3 & 100.0 & 100.0 \\

GR00T N1.7
& 15.0 & 15.0 & 0.0 & 100.0 & 3.3 & 100.0 & 0.0 & 90.0 & 100.0 & 96.7 & 91.7 & 85.0 & 85.0 & 71.7 & 95.0 & 91.7 & 100.0 & 45.0 & 100.0 & 98.3 \\

MolmoAct2
& 91.7 & 78.3 & 0.0 & 100.0 & 3.3 & 100.0 & 20.0 & 93.3 & 98.3 & 100.0 & 100.0 & 95.0 & 93.3 & 96.7 & 100.0 & 100.0 & 100.0 & 88.3 & 95.0 & 100.0 \\

Anchor-Align
& 3.3 & 38.3 & 0.0 & 96.7 & 0.0 & 100.0 & 0.0 & 41.7 & 95.0 & 73.3 & 71.7 & 78.3 & 71.7 & 88.3 & 98.3 & 100.0 & 100.0 & 21.7 & 83.3 & 96.7 \\

RIPT-VLA & 0.0 & 0.0 & 0.0 & 100.0 & 0.0 & 100.0 & 8.3 & 73.3 & 98.3 & 96.7 & 96.7 & 90.0 & 78.3 & 81.7 & 95.0 & 96.7 & 93.3 & 46.7 & 95.0 & 100.0 \\

Cosmos Policy
& 26.7 & 10.0 & 0.0 & 100.0 & 3.3 & 100.0 & 80.0 & 100.0 & 100.0 & 83.3 & 96.7 & 93.3 & 90.0 & 90.0 & 93.3 & 100.0 & 93.3 & 73.3 & 100.0 & 100.0 \\

FastWAM
& 37.3 & 4.0 & 0.0 & 98.3 & 1.7 & 100.0 & 0.0 & 99.0 & 100.0 & 100.0 & 100.0 & 100.0 & 99.0 & 99.0 & 100.0 & 100.0 & 100.0 & 59.0 & 100.0 & 98.3 \\

LingBot-VA
& 78.3 & 0.0 & 100.0 & 100.0 & 100.0 & 100.0 & 83.3 & 98.3 & 100.0 & 100.0 & 100.0 & 100.0 & 98.3 & 100.0 & 100.0 & 98.3 & 98.3 & 0.0 & 100.0 & 100.0 \\

\bottomrule
\end{tabular}%
}

\end{table*}

\begin{table*}[htb]
\centering
\caption{Static and runtime perturbation results on LIBERO-Spatial. All values are reported in percentages (\%).}
\label{tab:libero_spatial_full}
\renewcommand{\arraystretch}{1.15}
\setlength{\tabcolsep}{2.8pt}

\textbf{(a) Static Perturbations}
\vspace{1.5mm}

\resizebox{\textwidth}{!}{%
\begin{tabular}{lcccccccccccccccccccccc}
\toprule
\textbf{Model} & \textbf{S1} & \textbf{S2} & \textbf{S3} & \textbf{S4} & \textbf{S5} & \textbf{S6} & \textbf{S7} & \textbf{S8} & \textbf{S9} & \textbf{S10} & \textbf{S11} & \textbf{S12} & \textbf{S13} & \textbf{S14} & \textbf{S15} & \textbf{S16} & \textbf{S17} & \textbf{S18} & \textbf{S19} & \textbf{S20} & \textbf{S21} & \textbf{S22} \\
\midrule

OpenVLA
& 15.0 & 5.0 & 3.3 & N/A & 0.0 & 70.0 & 88.3 & 70.0 & 68.3 & 36.7 & 71.7 & 0.0 & 73.3 & 38.3 & 40.0 & 28.3 & 83.3 & 83.3 & 88.3 & 0.0 & 75.0 & 80.0 \\

OpenVLA-OFT
& 73.3 & 76.7 & 30.0 & N/A & 0.0 & 80.0 & 100.0 & 93.3 & 98.3 & 88.3 & 100.0 & 98.3 & 93.3 & 98.3 & 51.7 & 81.7 & 96.7 & 86.7 & 100.0 & 0.0 & 98.3 & 100.0 \\

OpenVLA-OFT$_{\mathrm{w}}$
& 40.0 & 15.0 & 5.0 & N/A & 1.7 & 75.0 & 98.3 & 91.7 & 98.3 & 48.3 & 96.7 & 6.7 & 98.3 & 98.3 & 93.3 & 90.0 & 100.0 & 10.0 & 98.3 & 0.0 & 98.3 & 100.0 \\

OpenVLA-OFT$_{\mathrm{m}}$
& 73.3 & 96.7 & 28.3 & N/A & 0.0 & 75.0 & 100.0 & 95.0 & 95.0 & 70.0 & 95.0 & 53.3 & 90.0 & 98.3 & 41.7 & 65.0 & 100.0 & 93.3 & 100.0 & 0.0 & 98.3 & 100.0 \\

OpenVLA-OFT+
& 66.7 & 98.3 & 35.0 & N/A & 0.0 & 81.7 & 96.7 & 95.0 & 95.0 & 86.7 & 88.3 & 66.7 & 98.3 & 98.3 & 38.3 & 56.7 & 98.3 & 100.0 & 98.3 & 0.0 & 98.3 & 98.3 \\

UniVLA
& 18.3 & 55.0 & 10.0 & N/A & 0.0 & 75.0 & 98.3 & 98.3 & 98.3 & 68.3 & 98.3 & 1.7 & 51.7 & 95.0 & 73.3 & 80.0 & 98.3 & 98.3 & 98.3 & 0.0 & 98.3 & 95.0 \\

$\pi_0$
& 6.7 & 0.0 & 8.3 & N/A & 0.0 & 40.0 & 61.7 & 71.7 & 65.0 & 55.0 & 66.7 & 61.7 & 56.7 & 60.0 & 58.3 & 71.7 & 55.0 & 10.0 & 68.3 & 3.3 & 60.0 & 60.0 \\

$\pi_0$-FAST
& 83.3 & 90.0 & 25.0 & N/A & 0.0 & 83.3 & 98.3 & 93.3 & 100.0 & 95.0 & 98.3 & 96.7 & 91.7 & 100.0 & 15.0 & 96.7 & 93.3 & 78.3 & 98.3 & 0.0 & 98.3 & 100.0 \\

$\pi_{0.5}$
& 95.0 & 100.0 & 40.0 & N/A & 6.7 & 85.0 & 100.0 & 98.3 & 93.3 & 96.7 & 100.0 & 81.7 & 100.0 & 100.0 & 0.0 & 100.0 & 98.3 & 100.0 & 100.0 & 0.0 & 98.3 & 98.3 \\

X-VLA
& 40.0 & 1.7 & 3.3 & N/A & 0.0 & 63.3 & 98.3 & 95.0 & 93.3 & 81.7 & 90.0 & 31.7 & 88.3 & 98.3 & 31.7 & 48.3 & 93.3 & 65.0 & 95.0 & 98.3 & 93.3 & 91.7 \\

GR00T N1.7
& 85.0 & 76.7 & 23.3 & N/A & 0.0 & 81.7 & 100.0 & 95.0 & 95.0 & 95.0 & 93.3 & 98.3 & 100.0 & 100.0 & 23.3 & 98.3 & 96.7 & 100.0 & 98.3 & 0.0 & 98.3 & 100.0 \\

MolmoAct2
& 75.0 & 95.0 & 31.7 & N/A & 0.0 & 85.0 & 100.0 & 88.3 & 100.0 & 95.0 & 100.0 & 98.3 & 98.3 & 100.0 & 40.0 & 98.3 & 96.7 & 100.0 & 100.0 & 0.0 & 96.7 & 98.3 \\

Anchor-Align
& 95.0 & 95.0 & 36.7 & N/A & 0.0 & 21.7 & 100.0 & 68.3 & 81.7 & 91.7 & 98.3 & 98.3 & 93.3 & 96.7 & 0.0 & 95.0 & 100.0 & 100.0 & 98.3 & 0.0 & 98.3 & 100.0 \\

RIPT-VLA & 78.3 & 76.7 & 26.7 & N/A & 0.0 & 75.0 & 98.3 & 93.3 & 95.0 & 88.3 & 100.0 & 98.3 & 91.7 & 96.7 & 48.3 & 81.7 & 100.0 & 93.3 & 100.0 & 0.0 & 100.0 & 100.0 \\

Cosmos Policy
& 73.3 & 56.7 & 53.3 & N/A & 13.3 & 90.0 & 100.0 & 93.3 & 96.7 & 90.0 & 90.0 & 93.3 & 96.7 & 96.7 & 63.3 & 93.3 & 83.3 & 93.3 & 90.0 & 0.0 & 90.0 & 100.0 \\

FastWAM
& 40.0 & 49.0 & 27.0 & N/A & 1.0 & 81.0 & 98.0 & 95.0 & 91.0 & 88.0 & 91.0 & 77.0 & 88.0 & 94.0 & 0.0 & 23.0 & 92.0 & 98.0 & 1.0 & 11.0 & 88.0 & 85.0 \\

LingBot-VA
& 83.3 & 75.0 & 30.0 & N/A & 93.3 & 0.0 & 0.0 & 91.7 & 0.0 & 98.3 & 100.0 & 93.3 & 100.0 & 98.3 & 13.3 & 21.7 & 100.0 & 100.0 & 96.7 & 93.3 & 100.0 & 100.0 \\

\bottomrule
\end{tabular}%
}

\vspace{2mm}

\textbf{(b) Runtime Perturbations}
\vspace{1.5mm}

\resizebox{\textwidth}{!}{%
\begin{tabular}{lcccccccccccccccccccc}
\toprule
\textbf{Model} & \textbf{D1} & \textbf{D2} & \textbf{D3} & \textbf{D4} & \textbf{D5} & \textbf{D6} & \textbf{D7} & \textbf{D8} & \textbf{D9} & \textbf{D10} & \textbf{D11} & \textbf{D12} & \textbf{D13} & \textbf{D14} & \textbf{D15} & \textbf{D16} & \textbf{D17} & \textbf{D18} & \textbf{D19} & \textbf{D20} \\
\midrule

OpenVLA
& 3.3 & 16.7 & 15.0 & 31.7 & 0.0 & 73.3 & 80.0 & 40.0 & 75.0 & 68.3 & 65.0 & 50.0 & 63.3 & 63.3 & 80.0 & 66.7 & 73.3 & 0.0 & 68.3 & 75.0 \\

OpenVLA-OFT
& 13.3 & 70.0 & 1.7 & 100.0 & 0.0 & 96.7 & 100.0 & 60.0 & 100.0 & 93.3 & 98.3 & 71.7 & 38.3 & 76.7 & 93.3 & 95.0 & 98.3 & 13.3 & 96.7 & 100.0 \\

OpenVLA-OFT$_{\mathrm{w}}$
& 5.0 & 16.7 & 3.3 & 55.0 & 0.0 & 88.3 & 98.3 & 88.3 & 100.0 & 100.0 & 90.0 & 85.0 & 45.0 & 78.3 & 93.3 & 78.3 & 96.7 & 0.0 & 100.0 & 100.0 \\

OpenVLA-OFT$_{\mathrm{m}}$
& 23.3 & 80.0 & 1.7 & 90.0 & 0.0 & 90.0 & 96.7 & 78.3 & 100.0 & 96.7 & 96.7 & 73.3 & 46.7 & 83.3 & 91.7 & 90.0 & 91.7 & 26.7 & 93.3 & 98.3 \\

OpenVLA-OFT+
& 25.0 & 80.0 & 3.3 & 91.7 & 0.0 & 86.7 & 98.3 & 88.3 & 100.0 & 98.3 & 100.0 & 83.3 & 61.7 & 78.3 & 91.7 & 95.0 & 93.3 & 33.3 & 93.3 & 98.3 \\

UniVLA
& 1.7 & 66.7 & 0.0 & 23.3 & 0.0 & 91.7 & 98.3 & 40.0 & 98.3 & 100.0 & 100.0 & 98.3 & 63.3 & 95.0 & 98.3 & 86.7 & 90.0 & 0.0 & 98.3 & 98.3 \\

$\pi_0$
& 3.3 & 1.7 & 15.0 & 60.0 & 0.0 & 48.3 & 50.0 & 60.0 & 56.7 & 58.3 & 55.0 & 48.3 & 31.7 & 16.7 & 25.0 & 18.3 & 3.3 & 0.0 & 28.3 & 66.7 \\

$\pi_0$-FAST
& 55.0 & 61.7 & 23.3 & 93.3 & 0.0 & 98.3 & 78.3 & 53.3 & 96.7 & 93.3 & 95.0 & 70.0 & 61.7 & 71.7 & 90.0 & 96.7 & 91.7 & 60.0 & 98.3 & 95.0 \\

$\pi_{0.5}$
& 76.7 & 98.3 & 26.7 & 98.3 & 5.0 & 96.7 & 21.7 & 95.0 & 100.0 & 95.0 & 98.3 & 95.0 & 93.3 & 88.3 & 95.0 & 98.3 & 100.0 & 95.0 & 98.3 & 100.0 \\

X-VLA
& 15.0 & 18.3 & 0.0 & 90.0 & 0.0 & 93.3 & 86.7 & 78.3 & 95.0 & 93.3 & 93.3 & 96.7 & 86.7 & 93.3 & 81.7 & 28.3 & 76.7 & 25.0 & 93.3 & 91.7 \\

GR00T N1.7
& 16.7 & 51.7 & 1.7 & 100.0 & 0.0 & 88.3 & 78.3 & 100.0 & 100.0 & 98.3 & 98.3 & 88.3 & 76.7 & 90.0 & 88.3 & 96.7 & 83.3 & 38.3 & 100.0 & 100.0 \\

MolmoAct2
& 70.0 & 83.3 & 26.7 & 90.0 & 1.7 & 93.3 & 56.7 & 100.0 & 100.0 & 100.0 & 98.3 & 98.3 & 86.7 & 90.0 & 86.7 & 100.0 & 83.3 & 86.7 & 100.0 & 100.0 \\

Anchor-Align
& 30.0 & 86.7 & 3.3 & 100.0 & 1.7 & 91.7 & 10.0 & 71.7 & 100.0 & 98.3 & 100.0 & 95.0 & 86.7 & 98.3 & 96.7 & 96.7 & 98.3 & 41.7 & 100.0 & 98.3 \\

RIPT-VLA & 16.7 & 65.0 & 0.0 & 100.0 & 0.0 & 96.7 & 100.0 & 60.0 & 100.0 & 100.0 & 98.3 & 80.0 & 40.0 & 85.0 & 91.7 & 91.7 & 98.3 & 13.3 & 100.0 & 100.0 \\

Cosmos Policy
& 10.0 & 30.0 & 30.0 & 90.0 & 10.0 & 96.7 & 70.0 & 80.0 & 93.3 & 90.0 & 90.0 & 76.7 & 66.7 & 53.3 & 96.7 & 90.0 & 83.3 & 6.7 & 96.7 & 96.7 \\

FastWAM
& 12.0 & 54.0 & 13.0 & 90.0 & 0.0 & 92.0 & 48.0 & 88.0 & 93.0 & 94.0 & 95.0 & 79.0 & 59.0 & 82.0 & 87.0 & 82.0 & 92.0 & 24.0 & 88.0 & 91.0 \\

LingBot-VA
& 70.0 & 80.0 & 100.0 & 100.0 & 98.3 & 98.3 & 88.3 & 98.3 & 100.0 & 91.7 & 98.3 & 100.0 & 98.3 & 100.0 & 100.0 & 98.3 & 85.0 & 1.7 & 95.0 & 100.0 \\

\bottomrule
\end{tabular}%
}

\end{table*}

\begin{table*}[htb]
\centering
\caption{Static and runtime perturbation results on LIBERO-10. All values are reported in percentages (\%).}
\label{tab:libero_10_full}
\renewcommand{\arraystretch}{1.15}
\setlength{\tabcolsep}{2.8pt}

\textbf{(a) Static Perturbations}
\vspace{1.5mm}

\resizebox{\textwidth}{!}{%
\begin{tabular}{lcccccccccccccccccccccc}
\toprule
\textbf{Model} & \textbf{S1} & \textbf{S2} & \textbf{S3} & \textbf{S4} & \textbf{S5} & \textbf{S6} & \textbf{S7} & \textbf{S8} & \textbf{S9} & \textbf{S10} & \textbf{S11} & \textbf{S12} & \textbf{S13} & \textbf{S14} & \textbf{S15} & \textbf{S16} & \textbf{S17} & \textbf{S18} & \textbf{S19} & \textbf{S20} & \textbf{S21} & \textbf{S22} \\
\midrule

OpenVLA
& 18.3 & 5.0 & 26.7 & 0.0 & 13.3 & 1.7 & 55.0 & 53.3 & 18.3 & 36.7 & 43.3 & 0.0 & 46.7 & 38.3 & 28.3 & 31.7 & 23.3 & 50.0 & 50.0 & 0.0 & 50.0 & 41.7 \\

OpenVLA-OFT
& 30.0 & 26.7 & 36.7 & 0.0 & 46.7 & 33.3 & 91.7 & 95.0 & 38.3 & 80.0 & 96.7 & 41.7 & 95.0 & 68.3 & 0.0 & 6.7 & 23.3 & 91.7 & 90.0 & 0.0 & 95.0 & 85.0 \\

OpenVLA-OFT$_{\mathrm{w}}$
& 35.0 & 8.3 & 33.3 & 3.3 & 61.7 & 11.7 & 78.3 & 85.0 & 46.7 & 70.0 & 66.7 & 0.0 & 75.0 & 70.0 & 73.3 & 60.0 & 41.7 & 63.3 & 71.7 & 1.7 & 76.7 & 66.7 \\

OpenVLA-OFT$_{\mathrm{m}}$
& 31.7 & 30.0 & 56.7 & 0.0 & 66.7 & 15.0 & 95.0 & 100.0 & 46.7 & 90.0 & 95.0 & 90.0 & 91.7 & 73.3 & 10.0 & 88.3 & 43.3 & 78.3 & 96.7 & 0.0 & 91.7 & 88.3 \\

OpenVLA-OFT+
& 36.7 & 35.0 & 48.3 & 0.0 & 68.3 & 10.0 & 90.0 & 86.7 & 43.3 & 83.3 & 85.0 & 91.7 & 85.0 & 88.3 & 0.0 & 75.0 & 41.7 & 75.0 & 90.0 & 0.0 & 81.7 & 85.0 \\

UniVLA
& 36.7 & 28.3 & 36.7 & 6.7 & 30.0 & 35.0 & 63.3 & 68.3 & 5.0 & 66.7 & 75.0 & 0.0 & 86.7 & 26.7 & 73.3 & 78.3 & 61.7 & 88.3 & 83.3 & 0.0 & 88.3 & 81.7 \\

$\pi_0$
& 0.0 & 1.7 & 18.3 & 0.0 & 31.7 & 10.0 & 55.0 & 43.3 & 18.3 & 51.7 & 50.0 & 45.0 & 40.0 & 41.7 & 41.7 & 41.7 & 31.7 & 5.0 & 50.0 & 1.7 & 43.3 & 48.3 \\

$\pi_0$-FAST
& 3.3 & 0.0 & 21.7 & 0.0 & 31.7 & 3.3 & 23.3 & 16.7 & 18.3 & 36.7 & 31.7 & 31.7 & 33.3 & 33.3 & 0.0 & 20.0 & 6.7 & 1.7 & 21.7 & 0.0 & 35.0 & 26.7 \\

$\pi_{0.5}$
& 36.7 & 20.0 & 68.3 & 76.7 & 53.3 & 25.0 & 71.7 & 70.0 & 38.3 & 83.3 & 81.7 & 63.3 & 75.0 & 75.0 & 0.0 & 50.0 & 70.0 & 70.0 & 71.7 & 3.3 & 78.3 & 80.0 \\

X-VLA
& 16.7 & 16.7 & 13.3 & 3.3 & 86.7 & 50.0 & 100.0 & 95.0 & 28.3 & 100.0 & 91.7 & 55.0 & 100.0 & 100.0 & 48.3 & 53.3 & 95.0 & 25.0 & 100.0 & 48.3 & 75.0 & 98.3 \\

GR00T N1.7
& 15.0 & 23.3 & 33.3 & 0.0 & 61.7 & 5.0 & 80.0 & 80.0 & 41.7 & 65.0 & 71.7 & 51.7 & 86.7 & 80.0 & 3.3 & 86.7 & 50.0 & 75.0 & 83.3 & 0.0 & 76.7 & 81.7 \\

MolmoAct2
& 41.7 & 28.3 & 83.3 & 86.7 & 60.0 & 18.3 & 81.7 & 78.3 & 41.7 & 91.7 & 95.0 & 56.7 & 76.7 & 76.7 & 23.3 & 80.0 & 60.0 & 73.3 & 86.7 & 0.0 & 88.3 & 83.3 \\

Anchor-Align
& 26.7 & 20.0 & 38.3 & 0.0 & 36.7 & 16.7 & 70.0 & 68.3 & 20.0 & 73.3 & 63.3 & 83.3 & 78.3 & 73.3 & 0.0 & 21.7 & 31.7 & 71.7 & 60.0 & 3.3 & 76.7 & 61.7 \\

RIPT-VLA & 23.3 & 16.7 & 23.3 & 0.0 & 45.0 & 16.7 & 90.0 & 96.7 & 38.3 & 78.3 & 93.3 & 45.0 & 85.0 & 75.0 & 0.0 & 21.7 & 26.7 & 85.0 & 88.3 & 0.0 & 88.3 & 83.3 \\

Cosmos Policy
& 76.7 & 70.0 & 33.3 & 3.3 & 20.0 & 33.3 & 90.0 & 90.0 & 66.7 & 90.0 & 90.0 & 100.0 & 100.0 & 90.0 & 46.7 & 100.0 & 83.3 & 93.3 & 96.7 & 0.0 & 96.7 & 90.0 \\

FastWAM
& 25.0 & 6.7 & 15.0 & 0.0 & 78.3 & 31.7 & 96.7 & 95.0 & 38.3 & 100.0 & 100.0 & 93.3 & 100.0 & 100.0 & 0.0 & 0.0 & 86.7 & 71.7 & 0.0 & 0.0 & 96.7 & 85.0 \\

LingBot-VA
& 46.7 & 51.7 & 88.3 & 0.0 & 83.3 & 38.3 & 0.0 & 98.3 & 0.0 & 76.7 & 80.0 & 90.0 & 91.7 & 85.0 & 6.7 & 26.7 & 98.3 & 96.7 & 60.0 & 91.7 & 98.3 & 93.3 \\

\bottomrule
\end{tabular}%
}

\vspace{2mm}

\textbf{(b) Runtime Perturbations}
\vspace{1.5mm}

\resizebox{\textwidth}{!}{%
\begin{tabular}{lcccccccccccccccccccc}
\toprule
\textbf{Model} & \textbf{D1} & \textbf{D2} & \textbf{D3} & \textbf{D4} & \textbf{D5} & \textbf{D6} & \textbf{D7} & \textbf{D8} & \textbf{D9} & \textbf{D10} & \textbf{D11} & \textbf{D12} & \textbf{D13} & \textbf{D14} & \textbf{D15} & \textbf{D16} & \textbf{D17} & \textbf{D18} & \textbf{D19} & \textbf{D20} \\
\midrule

OpenVLA
& 0.0 & 1.7 & 6.7 & 43.3 & 16.7 & 66.7 & 45.0 & 43.3 & 46.7 & 40.0 & 21.7 & 43.3 & 46.7 & 41.7 & 50.0 & 61.7 & 35.0 & 6.7 & 20.0 & 51.7 \\

OpenVLA-OFT
& 0.0 & 18.3 & 28.3 & 96.7 & 40.0 & 100.0 & 45.0 & 65.0 & 93.3 & 86.7 & 73.3 & 91.7 & 98.3 & 86.7 & 91.7 & 93.3 & 95.0 & 3.3 & 46.7 & 95.0 \\

OpenVLA-OFT$_{\mathrm{w}}$
& 0.0 & 11.7 & 16.7 & 81.7 & 46.7 & 100.0 & 81.7 & 60.0 & 70.0 & 61.7 & 71.7 & 68.3 & 70.0 & 73.3 & 73.3 & 73.3 & 70.0 & 0.0 & 51.7 & 83.3 \\

OpenVLA-OFT$_{\mathrm{m}}$
& 0.0 & 21.7 & 6.7 & 91.7 & 45.0 & 80.0 & 96.7 & 1.7 & 96.7 & 91.7 & 90.0 & 83.3 & 91.7 & 93.3 & 91.7 & 95.0 & 96.7 & 0.0 & 60.0 & 95.0 \\

OpenVLA-OFT+
& 0.0 & 23.3 & 11.7 & 88.3 & 41.7 & 80.0 & 78.3 & 1.7 & 93.3 & 71.7 & 71.7 & 78.3 & 78.3 & 71.7 & 85.0 & 81.7 & 80.0 & 0.0 & 56.7 & 90.0 \\

UniVLA
& 0.0 & 26.7 & 15.0 & 83.3 & 26.7 & 100.0 & 91.7 & 21.7 & 86.7 & 83.3 & 75.0 & 86.7 & 88.3 & 75.0 & 80.0 & 83.3 & 73.3 & 11.7 & 68.3 & 85.0 \\

$\pi_0$
& 0.0 & 10.0 & 10.0 & 41.7 & 18.3 & 70.0 & 41.7 & 36.7 & 48.3 & 38.3 & 50.0 & 43.3 & 30.0 & 11.7 & 36.7 & 41.7 & 5.0 & 0.0 & 15.0 & 48.3 \\

$\pi_0$-FAST
& 0.0 & 0.0 & 0.0 & 25.0 & 18.3 & 66.7 & 16.7 & 0.0 & 26.7 & 30.0 & 23.3 & 21.7 & 30.0 & 23.3 & 21.7 & 28.3 & 20.0 & 0.0 & 26.7 & 33.3 \\

$\pi_{0.5}$
& 5.0 & 36.7 & 11.7 & 75.0 & 43.3 & 100.0 & 43.3 & 20.0 & 73.3 & 81.7 & 66.7 & 76.7 & 71.7 & 68.3 & 76.7 & 80.0 & 68.3 & 28.3 & 70.0 & 83.3 \\

X-VLA
& 0.0 & 16.7 & 31.7 & 98.3 & 43.3 & 100.0 & 61.7 & 26.7 & 98.3 & 95.0 & 95.0 & 98.3 & 95.0 & 93.3 & 98.3 & 15.0 & 83.3 & 60.0 & 95.0 & 95.0 \\

GR00T N1.7
& 0.0 & 25.0 & 15.0 & 81.7 & 26.7 & 90.0 & 25.0 & 70.0 & 76.7 & 75.0 & 73.3 & 73.3 & 83.3 & 78.3 & 78.3 & 85.0 & 73.3 & 0.0 & 50.0 & 78.3 \\

MolmoAct2
& 1.7 & 21.7 & 18.3 & 85.0 & 45.0 & 100.0 & 66.7 & 61.7 & 78.3 & 71.7 & 73.3 & 75.0 & 78.3 & 63.3 & 81.7 & 83.3 & 83.3 & 25.0 & 73.3 & 83.3 \\

Anchor-Align
& 0.0 & 8.3 & 18.3 & 66.7 & 21.7 & 86.7 & 38.3 & 38.3 & 75.0 & 60.0 & 58.3 & 78.3 & 81.7 & 65.0 & 71.7 & 85.0 & 66.7 & 0.0 & 53.3 & 68.3 \\

RIPT-VLA & 0.0 & 13.3 & 26.7 & 90.0 & 38.3 & 100.0 & 45.0 & 53.3 & 86.7 & 81.7 & 66.7 & 78.3 & 90.0 & 80.0 & 86.7 & 86.7 & 75.0 & 0.0 & 50.0 & 91.7 \\

Cosmos Policy
& 40.0 & 43.3 & 23.3 & 96.7 & 10.0 & 100.0 & 96.7 & 70.0 & 100.0 & 93.3 & 86.7 & 90.0 & 90.0 & 66.7 & 83.3 & 90.0 & 83.3 & 33.3 & 83.3 & 96.7 \\

FastWAM
& 1.7 & 5.0 & 18.3 & 96.7 & 45.0 & 100.0 & 18.3 & 90.0 & 98.3 & 98.3 & 93.3 & 96.7 & 98.3 & 83.3 & 95.0 & 96.7 & 95.0 & 31.7 & 88.3 & 96.7 \\

LingBot-VA
& 21.7 & 36.7 & 93.3 & 95.0 & 95.0 & 95.0 & 71.7 & 88.3 & 96.7 & 85.0 & 91.7 & 91.7 & 96.7 & 95.0 & 96.7 & 96.7 & 46.7 & 1.7 & 100.0 & 96.7 \\

\bottomrule
\end{tabular}%
}

\end{table*}

\subsection{Complete Feedback Utilization Results}
\label{app:rq1a_full}

Table~\ref{tab:rq1a_feedback_all} reports all five policies under live and frozen
observations, with and without each dynamic perturbation.

\onecolumn

\begingroup
\footnotesize
\setlength{\tabcolsep}{2.2pt}
\renewcommand{\arraystretch}{1.10}
\setlength{\aboverulesep}{0pt}
\setlength{\belowrulesep}{0pt}
\setlength{\LTleft}{0pt}
\setlength{\LTright}{0pt}
\setlength{\LTcapwidth}{\textwidth}
\setlength{\LTpre}{4pt}
\setlength{\LTpost}{4pt}


\begin{longtable}{@{}p{0.195\textwidth}*{10}{>{\centering\arraybackslash}p{\dimexpr0.0805\textwidth-2\tabcolsep\relax}}@{}}

\caption{Complete feedback-utilization results. Success rates are percentages; $0/1$ denotes absence/presence of a perturbation, and live/freeze denotes the visual-observation condition. \textbf{Live} allows the policy to receive updated observations and infer actions normally, whereas \textbf{Freeze} holds the visual observation fixed at its pre-event value for a specified interval. For each model, the freeze duration is set to one-eighth of the number of inference steps required to complete unperturbed tasks, with freezing initiated immediately before the perturbation event. The feedback gain $G_{\mathrm{feedback}}=(S_{1,\mathrm{live}}-S_{1,\mathrm{freeze}})-(S_{0,\mathrm{live}}-S_{0,\mathrm{freeze}})$ is measured in percentage points. The two panels cover D01--D10 and D11--D20, respectively.}
\label{tab:rq1a_feedback_all}\\

\toprule
\multicolumn{11}{c}{\textit{(a): D01--D10}}\\
\midrule
\textbf{Metric} & \textbf{D01} & \textbf{D02} & \textbf{D03} & \textbf{D04} & \textbf{D05} & \textbf{D06} & \textbf{D07} & \textbf{D08} & \textbf{D09} & \textbf{D10}\\
\midrule
\endfirsthead

\multicolumn{11}{l}{\textit{Table \thetable\ (continued)}}\\
\toprule
\multicolumn{11}{c}{\textit{(a): D01--D10 (continued)}}\\
\midrule
\textbf{Metric} & \textbf{D01} & \textbf{D02} & \textbf{D03} & \textbf{D04} & \textbf{D05} & \textbf{D06} & \textbf{D07} & \textbf{D08} & \textbf{D09} & \textbf{D10}\\
\midrule
\endhead

\midrule
\multicolumn{11}{r}{\textit{Continued on the next page}}\\
\endfoot

\bottomrule
\endlastfoot

\multicolumn{11}{l}{\textbf{OpenVLA-OFT}}\\*
$S_{0,\mathrm{live}}$ & 97.1 & 97.1 & 97.1 & 97.1 & 97.1 & 97.1 & 97.1 & 97.1 & 97.1 & 97.1\\*
$S_{0,\mathrm{freeze}}$ & 45.6 & 43.8 & 33.8 & 50.0 & 42.5 & 55.0 & 48.1 & 48.1 & 48.1 & 48.1\\*
$S_{1,\mathrm{live}}$ & 5.4 & 24.3 & 17.6 & 95.8 & 16.8 & 95.3 & 38.3 & 56.7 & 94.2 & 87.9\\*
$S_{1,\mathrm{freeze}}$ & 0.0 & 3.8 & 0.0 & 41.9 & 13.8 & 43.6 & 21.3 & 41.3 & 48.1 & 47.5\\*
$\mathbf{G_{\mathrm{feedback}}}$ & \textbf{-46.1} & \textbf{-32.8} & \textbf{-45.8} & \textbf{6.9} & \textbf{-51.5} & \textbf{9.6} & \textbf{-31.9} & \textbf{-33.6} & \textbf{-2.9} & \textbf{-8.6}\\
\midrule

\multicolumn{11}{l}{\boldmath$\pi_{0.5}$\unboldmath}\\*
$S_{0,\mathrm{live}}$ & 93.3 & 93.3 & 93.3 & 93.3 & 93.3 & 93.3 & 93.3 & 93.3 & 93.3 & 93.3\\*
$S_{0,\mathrm{freeze}}$ & 86.9 & 91.3 & 80.0 & 83.8 & 85.6 & 92.9 & 85.6 & 85.6 & 85.6 & 85.6\\*
$S_{1,\mathrm{live}}$ & 50.4 & 76.9 & 19.8 & 92.9 & 21.1 & 97.4 & 16.3 & 65.8 & 92.1 & 88.8\\*
$S_{1,\mathrm{freeze}}$ & 41.3 & 64.4 & 15.6 & 85.6 & 41.3 & 92.1 & 38.8 & 83.8 & 86.3 & 86.3\\*
$\mathbf{G_{\mathrm{feedback}}}$ & \textbf{2.7} & \textbf{10.5} & \textbf{-9.1} & \textbf{-2.3} & \textbf{-27.8} & \textbf{4.8} & \textbf{-30.2} & \textbf{-25.6} & \textbf{-1.9} & \textbf{-5.2}\\
\midrule

\multicolumn{11}{l}{\textbf{Anchor-Align}}\\*
$S_{0,\mathrm{live}}$ & 91.3 & 91.3 & 91.3 & 91.3 & 91.3 & 91.3 & 91.3 & 91.3 & 91.3 & 91.3\\*
$S_{0,\mathrm{freeze}}$ & 78.1 & 83.8 & 56.9 & 72.5 & 72.5 & 78.6 & 74.4 & 74.4 & 74.4 & 74.4\\*
$S_{1,\mathrm{live}}$ & 16.3 & 38.2 & 15.3 & 88.8 & 13.9 & 79.8 & 12.1 & 39.2 & 90.8 & 75.4\\*
$S_{1,\mathrm{freeze}}$ & 16.3 & 30.0 & 11.3 & 74.4 & 25.6 & 80.0 & 19.4 & 57.5 & 73.1 & 75.6\\*
$\mathbf{G_{\mathrm{feedback}}}$ & \textbf{-13.1} & \textbf{0.7} & \textbf{-30.3} & \textbf{-4.4} & \textbf{-30.5} & \textbf{-12.9} & \textbf{-24.2} & \textbf{-35.2} & \textbf{0.8} & \textbf{-17.1}\\
\midrule

\multicolumn{11}{l}{\textbf{FastWAM}}\\*
$S_{0,\mathrm{live}}$ & 97.8 & 97.8 & 97.8 & 97.8 & 97.8 & 97.8 & 97.8 & 97.8 & 97.8 & 97.8\\*
$S_{0,\mathrm{freeze}}$ & 92.5 & 83.1 & 52.5 & 91.9 & 81.9 & 94.3 & 91.9 & 91.9 & 91.9 & 91.9\\*
$S_{1,\mathrm{live}}$ & 20.7 & 17.3 & 17.6 & 92.1 & 18.3 & 94.1 & 16.6 & 89.7 & 94.9 & 96.0\\*
$S_{1,\mathrm{freeze}}$ & 23.8 & 13.1 & 6.3 & 91.3 & 36.3 & 92.9 & 30.6 & 90.6 & 90.0 & 91.9\\*
$\mathbf{G_{\mathrm{feedback}}}$ & \textbf{-8.3} & \textbf{-10.4} & \textbf{-33.9} & \textbf{-5.0} & \textbf{-33.9} & \textbf{-2.2} & \textbf{-19.9} & \textbf{-6.8} & \textbf{-1.0} & \textbf{-1.8}\\
\midrule

\multicolumn{11}{l}{\textbf{Cosmos Policy}}\\*
$S_{0,\mathrm{live}}$ & 97.5 & 97.5 & 97.5 & 97.5 & 97.5 & 97.5 & 97.5 & 97.5 & 97.5 & 97.5\\*
$S_{0,\mathrm{freeze}}$ & 93.1 & 93.8 & 74.4 & 95.6 & 82.5 & 94.3 & 91.9 & 91.9 & 91.9 & 91.9\\*
$S_{1,\mathrm{live}}$ & 21.7 & 26.9 & 21.4 & 95.8 & 9.0 & 97.4 & 75.8 & 80.8 & 97.5 & 91.7\\*
$S_{1,\mathrm{freeze}}$ & 15.0 & 20.0 & 15.6 & 94.4 & 50.0 & 96.4 & 85.6 & 90.0 & 92.5 & 90.0\\*
$\mathbf{G_{\mathrm{feedback}}}$ & \textbf{2.3} & \textbf{3.2} & \textbf{-17.4} & \textbf{-0.4} & \textbf{-56.1} & \textbf{-2.2} & \textbf{-15.4} & \textbf{-14.8} & \textbf{-0.6} & \textbf{-4.0}\\

\end{longtable}


\addtocounter{table}{-1}

\begin{longtable}{@{}p{0.195\textwidth}*{10}{>{\centering\arraybackslash}p{\dimexpr0.0805\textwidth-2\tabcolsep\relax}}@{}}

\caption[]{Complete feedback-utilization results (continued). Panel (b): D11--D20.}\\

\toprule
\multicolumn{11}{c}{\textit{(b): D11--D20}}\\
\midrule
\textbf{Metric} & \textbf{D11} & \textbf{D12} & \textbf{D13} & \textbf{D14} & \textbf{D15} & \textbf{D16} & \textbf{D17} & \textbf{D18} & \textbf{D19} & \textbf{D20}\\
\midrule
\endfirsthead

\multicolumn{11}{l}{\textit{Table \thetable\ (continued)}}\\
\toprule
\multicolumn{11}{c}{\textit{(b): D11--D20 (continued)}}\\
\midrule
\textbf{Metric} & \textbf{D11} & \textbf{D12} & \textbf{D13} & \textbf{D14} & \textbf{D15} & \textbf{D16} & \textbf{D17} & \textbf{D18} & \textbf{D19} & \textbf{D20}\\
\midrule
\endhead

\midrule
\multicolumn{11}{r}{\textit{Continued on the next page}}\\
\endfoot

\bottomrule
\endlastfoot

\multicolumn{11}{l}{\textbf{OpenVLA-OFT}}\\*
$S_{0,\mathrm{live}}$ & 97.1 & 97.1 & 97.1 & 97.1 & 97.1 & 97.1 & 97.1 & 97.1 & 97.1 & 97.1\\*
$S_{0,\mathrm{freeze}}$ & 48.1 & 48.1 & 48.1 & 48.1 & 41.3 & 48.1 & 48.1 & 41.9 & 32.5 & 48.1\\*
$S_{1,\mathrm{live}}$ & 84.6 & 79.6 & 70.1 & 70.4 & 91.7 & 93.8 & 93.3 & 18.3 & 62.9 & 95.4\\*
$S_{1,\mathrm{freeze}}$ & 45.0 & 47.5 & 46.3 & 43.8 & 43.1 & 41.3 & 42.5 & 1.3 & 43.8 & 47.5\\*
$\mathbf{G_{\mathrm{feedback}}}$ & \textbf{-9.4} & \textbf{-16.9} & \textbf{-25.1} & \textbf{-22.3} & \textbf{-7.3} & \textbf{3.5} & \textbf{1.8} & \textbf{-38.1} & \textbf{-45.4} & \textbf{-1.1}\\
\midrule

\multicolumn{11}{l}{\boldmath$\pi_{0.5}$\unboldmath}\\*
$S_{0,\mathrm{live}}$ & 93.3 & 93.3 & 93.3 & 93.3 & 93.3 & 93.3 & 93.3 & 93.3 & 93.3 & 93.3\\*
$S_{0,\mathrm{freeze}}$ & 85.6 & 85.6 & 85.6 & 85.6 & 87.5 & 85.6 & 88.1 & 88.1 & 71.3 & 86.3\\*
$S_{1,\mathrm{live}}$ & 86.3 & 87.5 & 83.7 & 83.3 & 90.4 & 92.9 & 87.6 & 65.3 & 78.3 & 94.2\\*
$S_{1,\mathrm{freeze}}$ & 80.6 & 84.4 & 76.9 & 85.6 & 87.5 & 88.1 & 85.6 & 19.4 & 71.3 & 85.0\\*
$\mathbf{G_{\mathrm{feedback}}}$ & \textbf{-2.1} & \textbf{-4.6} & \textbf{-0.9} & \textbf{-10.0} & \textbf{-2.9} & \textbf{-2.9} & \textbf{-3.2} & \textbf{40.7} & \textbf{-15.0} & \textbf{2.1}\\
\midrule

\multicolumn{11}{l}{\textbf{Anchor-Align}}\\*
$S_{0,\mathrm{live}}$ & 91.3 & 91.3 & 91.3 & 91.3 & 91.3 & 91.3 & 91.3 & 91.3 & 91.3 & 91.3\\*
$S_{0,\mathrm{freeze}}$ & 74.4 & 74.4 & 74.4 & 74.4 & 72.5 & 74.4 & 88.1 & 71.3 & 58.1 & 74.4\\*
$S_{1,\mathrm{live}}$ & 73.3 & 81.3 & 79.6 & 79.6 & 90.8 & 92.1 & 85.9 & 21.9 & 68.8 & 88.3\\*
$S_{1,\mathrm{freeze}}$ & 71.3 & 75.6 & 71.3 & 76.9 & 83.1 & 82.5 & 73.8 & 12.5 & 60.6 & 79.4\\*
$\mathbf{G_{\mathrm{feedback}}}$ & \textbf{-14.8} & \textbf{-11.3} & \textbf{-8.5} & \textbf{-14.2} & \textbf{-11.0} & \textbf{-7.3} & \textbf{9.0} & \textbf{-10.7} & \textbf{-25.0} & \textbf{-7.9}\\
\midrule

\multicolumn{11}{l}{\textbf{FastWAM}}\\*
$S_{0,\mathrm{live}}$ & 97.8 & 97.8 & 97.8 & 97.8 & 97.8 & 97.8 & 97.8 & 97.8 & 97.8 & 97.8\\*
$S_{0,\mathrm{freeze}}$ & 91.9 & 91.9 & 91.9 & 91.9 & 80.6 & 91.9 & 91.3 & 83.8 & 58.8 & 91.9\\*
$S_{1,\mathrm{live}}$ & 94.2 & 88.9 & 83.8 & 86.5 & 93.4 & 89.7 & 94.8 & 38.2 & 81.2 & 93.2\\*
$S_{1,\mathrm{freeze}}$ & 86.3 & 89.4 & 77.5 & 84.4 & 88.8 & 90.0 & 94.4 & 19.4 & 61.3 & 90.6\\*
$\mathbf{G_{\mathrm{feedback}}}$ & \textbf{2.0} & \textbf{-6.3} & \textbf{0.4} & \textbf{-3.8} & \textbf{-12.5} & \textbf{-6.2} & \textbf{-6.1} & \textbf{4.9} & \textbf{-19.1} & \textbf{-3.3}\\
\midrule

\multicolumn{11}{l}{\textbf{Cosmos Policy}}\\*
$S_{0,\mathrm{live}}$ & 97.5 & 97.5 & 97.5 & 97.5 & 97.5 & 97.5 & 97.5 & 97.5 & 97.5 & 97.5\\*
$S_{0,\mathrm{freeze}}$ & 91.9 & 91.9 & 91.9 & 91.9 & 83.1 & 91.9 & 97.5 & 84.4 & 71.3 & 91.9\\*
$S_{1,\mathrm{live}}$ & 90.8 & 87.5 & 81.7 & 70.8 & 92.5 & 93.3 & 87.4 & 32.5 & 93.3 & 95.8\\*
$S_{1,\mathrm{freeze}}$ & 88.8 & 95.0 & 81.3 & 82.5 & 95.6 & 92.5 & 90.6 & 3.8 & 76.3 & 95.6\\*
$\mathbf{G_{\mathrm{feedback}}}$ & \textbf{-3.5} & \textbf{-13.1} & \textbf{-5.2} & \textbf{-17.3} & \textbf{-17.5} & \textbf{-4.8} & \textbf{-3.3} & \textbf{15.7} & \textbf{-9.2} & \textbf{-5.4}\\

\end{longtable}

\endgroup

\FloatBarrier

\subsection{Complete Recovery Intervention Results}
\label{app:rq1c_full}

Figure~\ref{fig:rq1c_heatmap} and Table~\ref{tab:rq1c_interventions}
provide the recovery gains and success rates for Hold, Retry, and Restore (Reset).

\begin{figure}[!htbp]
    \centering
    \includegraphics[width=0.99\textwidth]{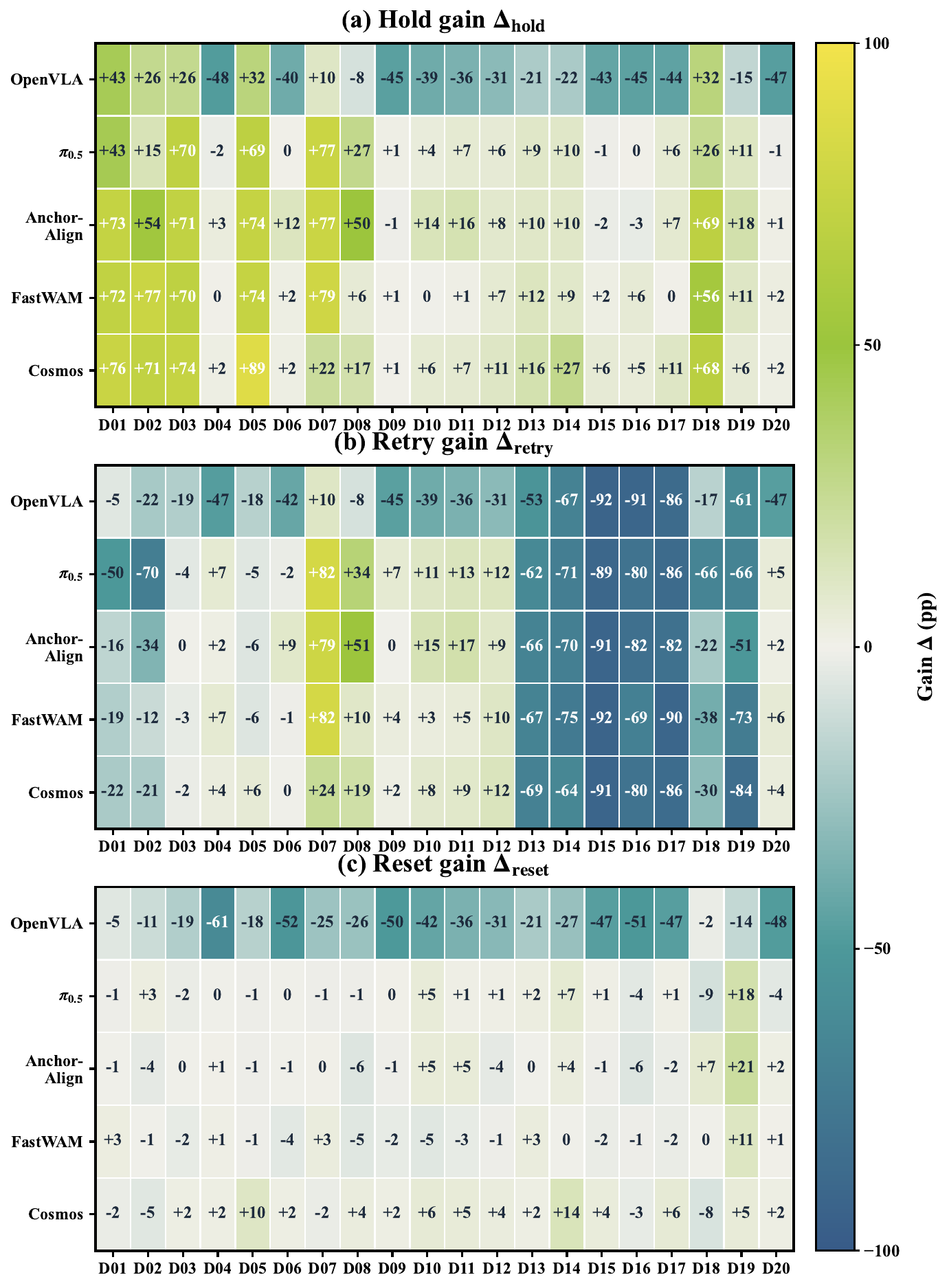}
    \caption{Recovery-intervention gains, $\Delta_I=S_I-S_{\mathrm{continue}}$
    (percentage points), for Hold, Retry, and Restore (labeled Reset in the figure). Positive values indicate improvement over Continue, while negative values indicate degradation.}
    \label{fig:rq1c_heatmap}
\end{figure}
\FloatBarrier
\begin{table*}[t]
\centering
\caption{
Success rates (\%) under Hold, Retry, and Restore (Reset) interventions across all dynamic perturbations.
Both panels retain the same intervention and model order.
}
\label{tab:rq1c_interventions}

\footnotesize
\setlength{\tabcolsep}{3.5pt}
\renewcommand{\arraystretch}{1.15}

\begin{tabular}{@{}l*{10}{S[table-format=3.1]}@{}}
\toprule
\multicolumn{11}{c}{\textit{(a) D01--D10}} \\
\midrule
\textbf{Model}
& \multicolumn{1}{c}{\textbf{D01}}
& \multicolumn{1}{c}{\textbf{D02}}
& \multicolumn{1}{c}{\textbf{D03}}
& \multicolumn{1}{c}{\textbf{D04}}
& \multicolumn{1}{c}{\textbf{D05}}
& \multicolumn{1}{c}{\textbf{D06}}
& \multicolumn{1}{c}{\textbf{D07}}
& \multicolumn{1}{c}{\textbf{D08}}
& \multicolumn{1}{c}{\textbf{D09}}
& \multicolumn{1}{c}{\textbf{D10}} \\
\midrule

\multicolumn{11}{l}{\textbf{Hold}} \\
OpenVLA-OFT   & 48.1 & 48.8 & 46.3 & 47.5 & 49.4 & 55.7 & 48.8 & 48.8 & 48.8 & 48.8 \\
$\pi_{0.5}$   & 93.8 & 92.5 & 92.5 & 90.6 & 91.3 & 97.9 & 93.1 & 93.1 & 93.1 & 93.1 \\
Anchor-Align  & 89.4 & 90.0 & 89.4 & 91.9 & 88.8 & 92.1 & 89.4 & 89.4 & 89.4 & 89.4 \\
FastWAM       & 92.5 & 93.1 & 90.0 & 92.5 & 92.5 & 96.4 & 95.6 & 95.6 & 95.6 & 95.6 \\
Cosmos Policy & 98.1 & 98.1 & 96.3 & 98.1 & 98.1 & 99.3 & 98.1 & 98.1 & 98.1 & 98.1 \\

\addlinespace[3pt]
\multicolumn{11}{l}{\textbf{Retry}} \\
OpenVLA-OFT   & 0.0 & 0.0 & 0.6 & 48.8 & 0.0 & 53.6 & 48.8 & 48.8 & 48.8 & 48.8 \\
$\pi_{0.5}$   & 0.0 & 7.5 & 18.1 & 98.8 & 16.9 & 95.7 & 98.8 & 99.4 & 98.8 & 99.4 \\
Anchor-Align  & 0.6 & 1.9 & 17.5 & 90.6 & 9.4 & 89.3 & 90.6 & 90.6 & 90.6 & 90.6 \\
FastWAM       & 1.9 & 4.4 & 16.9 & 98.8 & 12.5 & 92.9 & 98.8 & 99.4 & 98.8 & 99.4 \\
Cosmos Policy & 0.0 & 5.6 & 20.6 & 99.4 & 15.0 & 97.1 & 99.4 & 100.0 & 99.4 & 100.0 \\

\addlinespace[3pt]
\multicolumn{11}{l}{\textbf{Reset}} \\
OpenVLA-OFT   & 0.6 & 11.3 & 0.6 & 34.4 & 0.0 & 43.6 & 13.1 & 30.6 & 44.4 & 45.6 \\
$\pi_{0.5}$   & 49.4 & 80.6 & 20.0 & 91.9 & 20.6 & 97.1 & 15.0 & 65.0 & 92.5 & 93.8 \\
Anchor-Align  & 15.6 & 32.5 & 17.5 & 89.4 & 13.8 & 79.3 & 12.5 & 33.1 & 90.0 & 80.6 \\
FastWAM       & 23.8 & 15.6 & 17.5 & 93.1 & 17.5 & 90.0 & 20.0 & 84.4 & 92.5 & 91.3 \\
Cosmos Policy & 20.0 & 21.9 & 24.4 & 97.5 & 19.4 & 99.3 & 73.8 & 84.4 & 99.4 & 97.5 \\

\addlinespace[7pt]
\toprule
\multicolumn{11}{c}{\textit{(b) D11--D20}} \\
\midrule
\textbf{Model}
& \multicolumn{1}{c}{\textbf{D11}}
& \multicolumn{1}{c}{\textbf{D12}}
& \multicolumn{1}{c}{\textbf{D13}}
& \multicolumn{1}{c}{\textbf{D14}}
& \multicolumn{1}{c}{\textbf{D15}}
& \multicolumn{1}{c}{\textbf{D16}}
& \multicolumn{1}{c}{\textbf{D17}}
& \multicolumn{1}{c}{\textbf{D18}}
& \multicolumn{1}{c}{\textbf{D19}}
& \multicolumn{1}{c}{\textbf{D20}} \\
\midrule

\multicolumn{11}{l}{\textbf{Hold}} \\
OpenVLA-OFT   & 48.8 & 48.8 & 48.8 & 48.8 & 48.8 & 48.8 & 48.8 & 48.8 & 48.1 & 48.8 \\
$\pi_{0.5}$   & 93.1 & 93.1 & 93.1 & 93.1 & 89.4 & 93.1 & 94.4 & 91.9 & 89.4 & 93.1 \\
Anchor-Align  & 89.4 & 89.4 & 89.4 & 89.4 & 88.8 & 89.4 & 92.5 & 91.3 & 86.3 & 89.4 \\
FastWAM       & 95.6 & 95.6 & 95.6 & 95.6 & 95.6 & 95.6 & 95.0 & 93.8 & 91.9 & 95.6 \\
Cosmos Policy & 98.1 & 98.1 & 98.1 & 98.1 & 98.1 & 98.1 & 98.1 & 98.8 & 99.4 & 98.1 \\

\addlinespace[3pt]
\multicolumn{11}{l}{\textbf{Retry}} \\
OpenVLA-OFT   & 48.8 & 48.8 & 16.3 & 3.1 & 0.0 & 3.1 & 6.9 & 0.0 & 1.9 & 48.8 \\
$\pi_{0.5}$   & 99.4 & 99.4 & 21.9 & 12.5 & 1.3 & 12.5 & 1.9 & 0.0 & 11.9 & 99.4 \\
Anchor-Align  & 90.6 & 90.6 & 13.1 & 9.4 & 0.0 & 10.0 & 3.8 & 0.0 & 17.5 & 90.6 \\
FastWAM       & 99.4 & 99.4 & 16.9 & 11.9 & 1.3 & 20.6 & 4.4 & 0.0 & 8.1 & 99.4 \\
Cosmos Policy & 100.0 & 100.0 & 13.1 & 6.9 & 1.3 & 13.1 & 1.9 & 0.6 & 9.4 & 100.0 \\

\addlinespace[3pt]
\multicolumn{11}{l}{\textbf{Reset}} \\
OpenVLA-OFT   & 48.8 & 48.1 & 48.1 & 43.8 & 45.0 & 43.1 & 46.3 & 15.0 & 49.4 & 47.5 \\
$\pi_{0.5}$   & 87.5 & 88.8 & 86.3 & 90.0 & 91.3 & 88.8 & 88.8 & 56.9 & 96.3 & 90.0 \\
Anchor-Align  & 78.8 & 77.5 & 79.4 & 83.1 & 90.0 & 86.3 & 83.1 & 29.4 & 90.0 & 90.0 \\
FastWAM       & 91.3 & 87.5 & 86.9 & 86.3 & 91.3 & 88.8 & 92.5 & 38.1 & 92.5 & 93.8 \\
Cosmos Policy & 95.6 & 91.3 & 83.8 & 84.4 & 96.9 & 90.6 & 93.1 & 23.1 & 98.8 & 97.5 \\
\bottomrule
\end{tabular}%
\end{table*}

\FloatBarrier

\Needspace{30\baselineskip}
\subsection{Complete Robustness Composition Results}
\label{app:rq2_full}

Table~\ref{tab:rq2_composed_sr} lists all evaluated pairwise and three-way
compositions, grouped by static and dynamic intervention types.
\begin{table}[H]
\centering
\caption{
Success rates (\%) under pairwise and three-way perturbation compositions.
Compositions are grouped by perturbation type.
}
\label{tab:rq2_composed_sr}
\small
\setlength{\tabcolsep}{4.5pt}
\renewcommand{\arraystretch}{1.08}

\resizebox{\linewidth}{!}{
\begin{tabular}{llccccc}
\toprule
\textbf{Case} &
\textbf{Composition} &
\textbf{OpenVLA-OFT} &
\textbf{$\pi_{0.5}$} &
\textbf{Anchor-Align} &
\textbf{FastWAM} &
\textbf{Cosmos Policy} \\
\midrule

\rowcolor{black!15}
\multicolumn{7}{c}{\textit{Static + Static}}\\
P01 & S11 + S15
& 30.4 & 0.0 & 0.0 & 5.0 & 50.8 \\
P02 & S09 + S13
& 12.5 & 12.5 & 10.0 & 9.2 & 12.9 \\
P03 & S19 + S21
& 82.9 & 86.3 & 82.5 & 5.8 & 85.4 \\
P04 & S08 + S22
& 82.5 & 80.8 & 73.8 & 76.3 & 80.4 \\

\addlinespace[2pt]
\rowcolor{black!15}
\multicolumn{7}{c}{\textit{Static + Dynamic}} \\
P05 & S11 + D01
& 7.9 & 56.3 & 14.6 & 15.8 & 21.7 \\
P06 & S09 + D13
& 12.5 & 12.1 & 11.3 & 11.7 & 11.7 \\
P07 & S15 + D18
& 0.0 & 0.0 & 0.0 & 0.0 & 0.4 \\

\addlinespace[2pt]
\rowcolor{black!15}
\multicolumn{7}{c}{\textit{Dynamic + Dynamic}} \\
P08 & D02 + D18
& 0.0 & 22.9 & 0.8 & 0.0 & 1.3 \\
P09 & D03 + D19
& 60.0 & 73.8 & 67.5 & 73.8 & 80.8 \\
P10 & D18 + D20
& 0.0 & 17.5 & 0.0 & 0.0 & 2.9 \\

\midrule
\rowcolor{black!15}
\multicolumn{7}{c}{\textit{Three-way Compositions}} \\
T01 & S11 + S21 + D01
& 7.1 & 56.7 & 14.6 & 14.6 & 24.6 \\
T02 & S13 + S15 + D18
& 0.0 & 0.0 & 0.0 & 0.0 & 0.0 \\
T03 & S09 + S22 + D13
& 12.5 & 12.9 & 11.7 & 10.0 & 12.1 \\
T04 & S13 + D02 + D18
& 0.0 & 8.3 & 0.0 & 0.0 & 0.0 \\
T05 & S13 + S22 + D09
& 80.8 & 86.7 & 75.0 & 75.0 & 86.7 \\
T06 & S10 + S18 + D06
& 83.8 & 87.1 & 73.3 & 80.4 & 87.1 \\
T07 & S11 + S22 + D07
& 81.3 & 31.7 & 12.9 & 23.8 & 85.4 \\
T08 & S19 + S21 + D14
& 84.6 & 85.0 & 81.3 & 6.3 & 85.4 \\
T09 & S08 + S13 + D17
& 69.6 & 80.0 & 52.5 & 76.3 & 63.8 \\
T10 & S14 + S21 + D18
& 0.0 & 21.7 & 0.0 & 0.0 & 1.7 \\

\bottomrule
\end{tabular}
}
\end{table}
\FloatBarrier

\clearpage
\section{Perturbation Design}
\label{app:perturbation_visualizations}

The following panels illustrate all 22 static and 20 dynamic perturbations.
Each row shows one perturbation using aligned views or consecutive frames;
the complete definitions are given in Appendix~\ref{app:perturbation_catalogue}.



\makeatletter
\setlength{\@fptop}{0pt}
\setlength{\@dblfptop}{0pt}
\makeatother

\begin{figure}[!htp]
  \centering
  \includegraphics[width=0.88\textwidth]{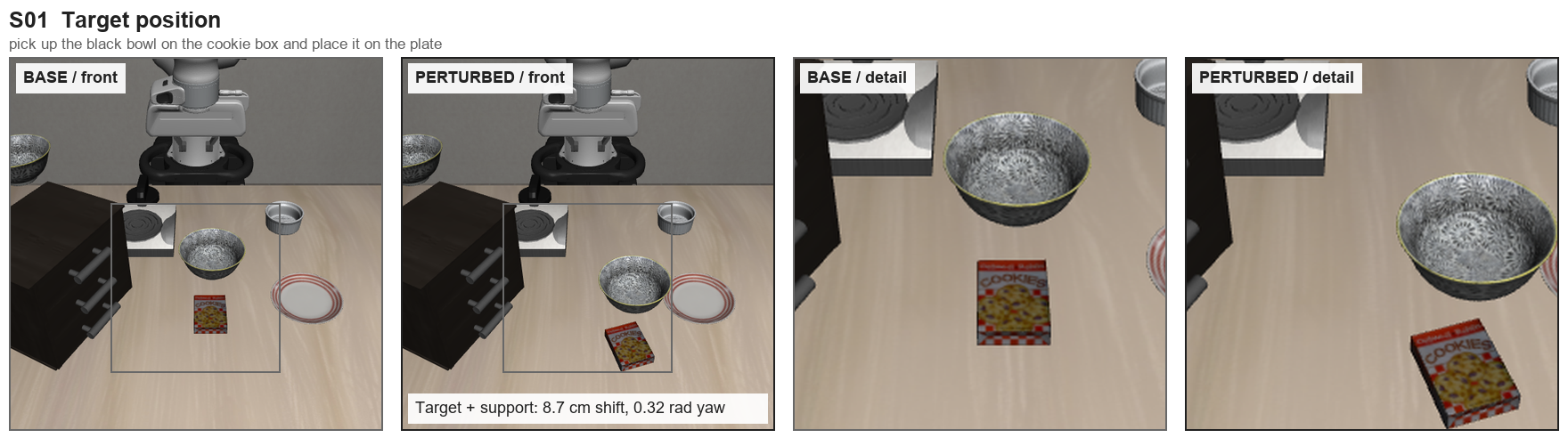}\par\vspace{-2pt}
  \includegraphics[width=0.88\textwidth]{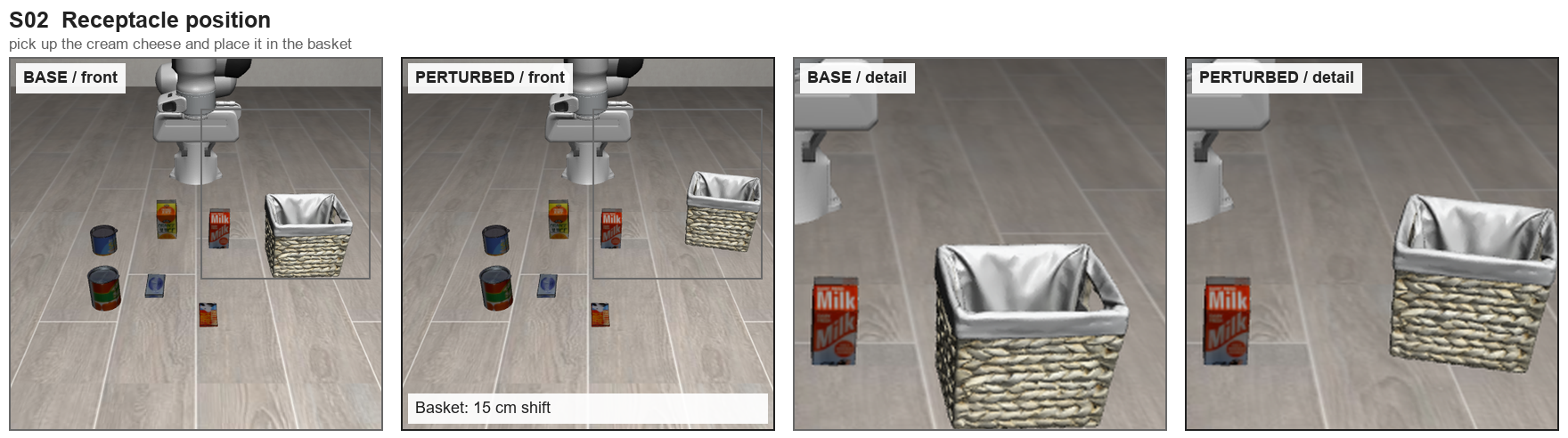}\par\vspace{-2pt}
  \includegraphics[width=0.88\textwidth]{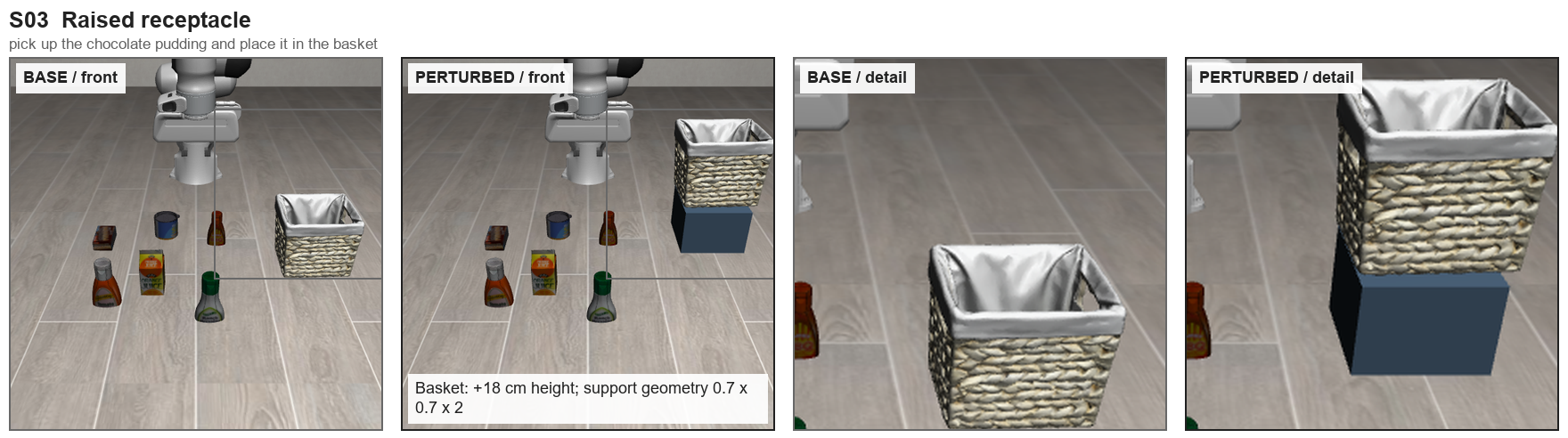}\par\vspace{-2pt}
  \includegraphics[width=0.88\textwidth]{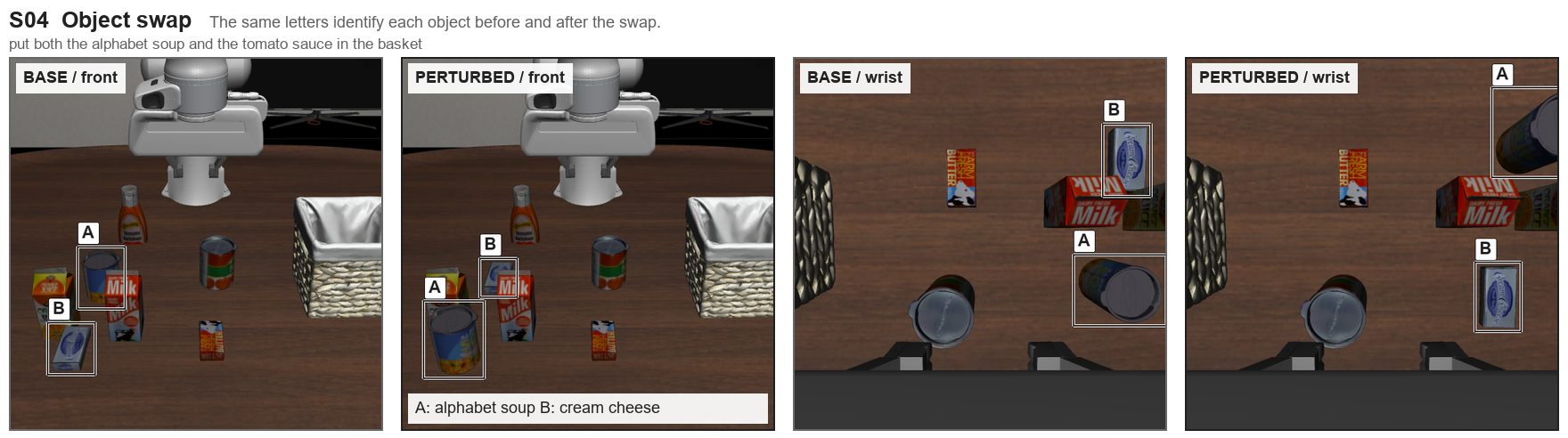}\par\vspace{-2pt}
  \includegraphics[width=0.88\textwidth]{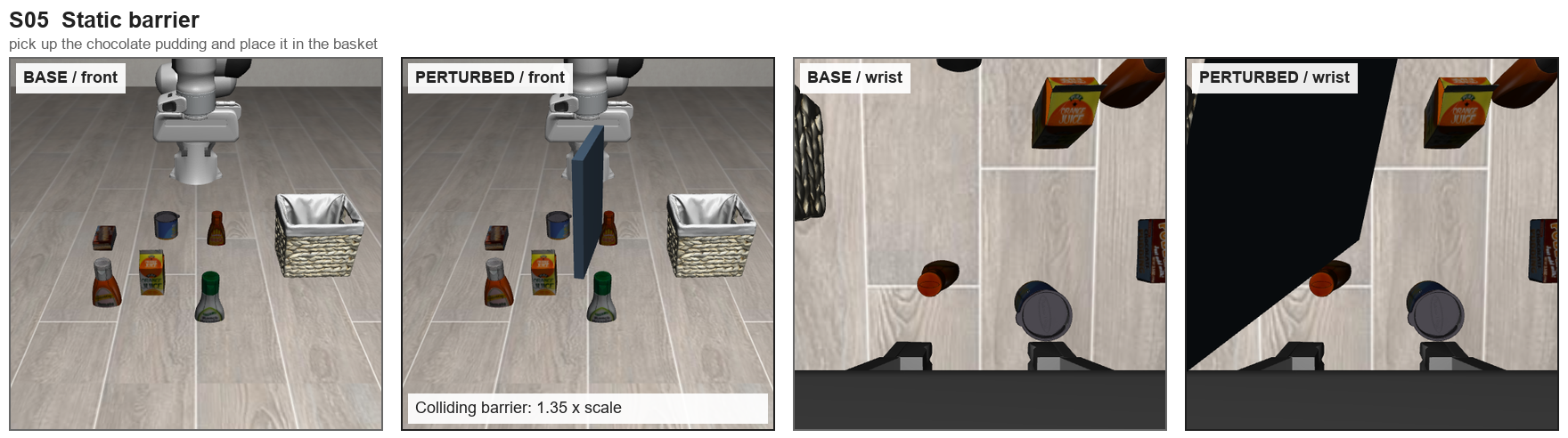}\par\vspace{-2pt}
  \caption{Static perturbations S01--S05. Each row contrasts BASE and the perturbed scene using the external and wrist views.}
\end{figure}
\clearpage

\begin{figure}[!htp]
  \centering
  \includegraphics[width=0.88\textwidth]{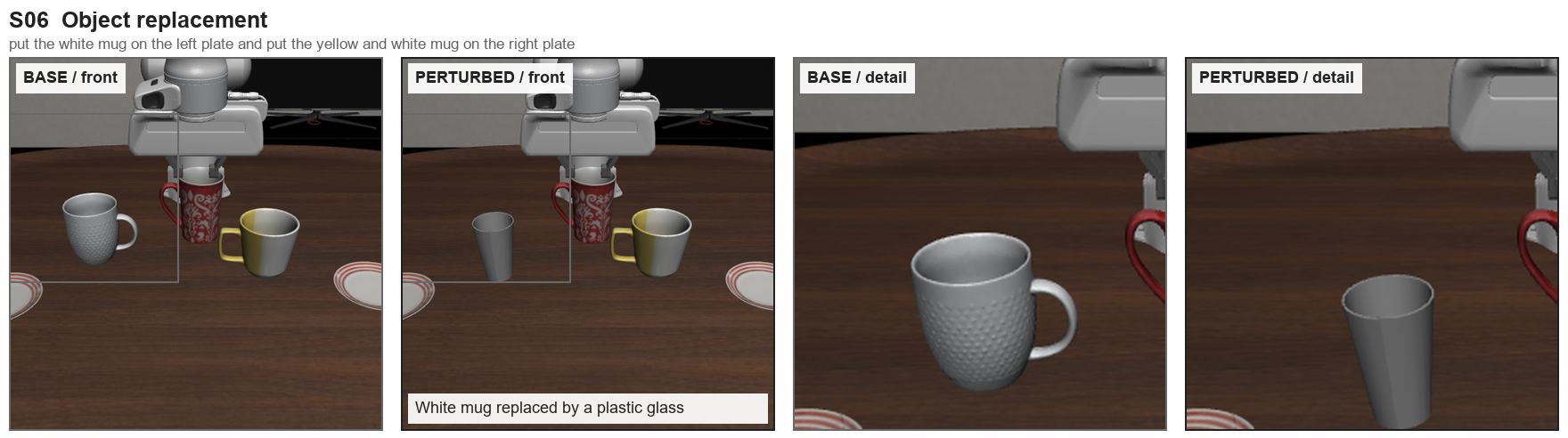}\par\vspace{-2pt}
  \includegraphics[width=0.88\textwidth]{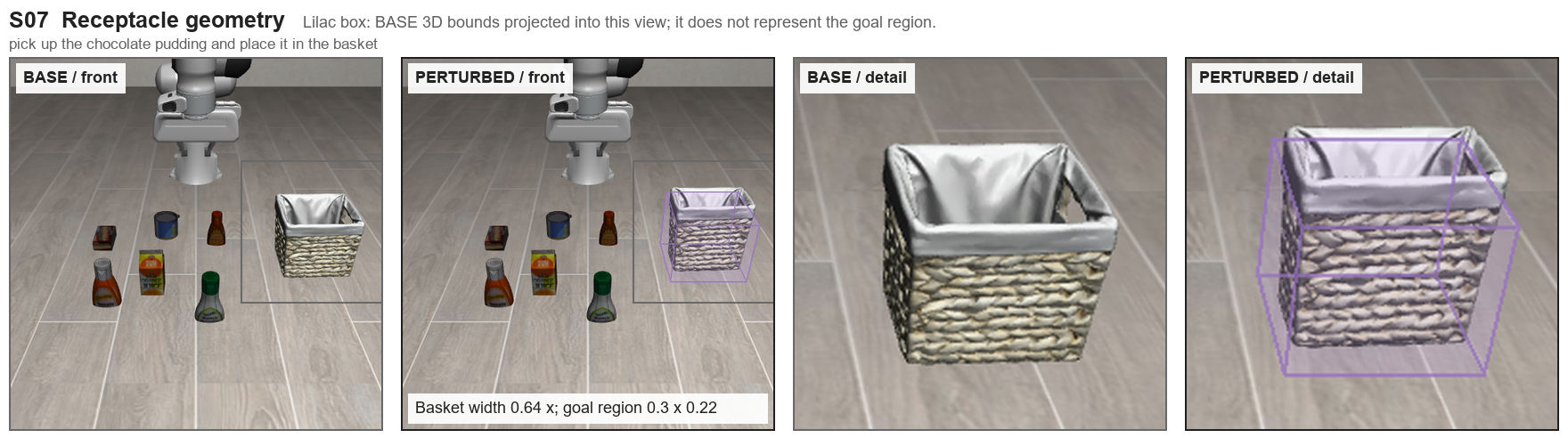}\par\vspace{-2pt}
  \includegraphics[width=0.88\textwidth]{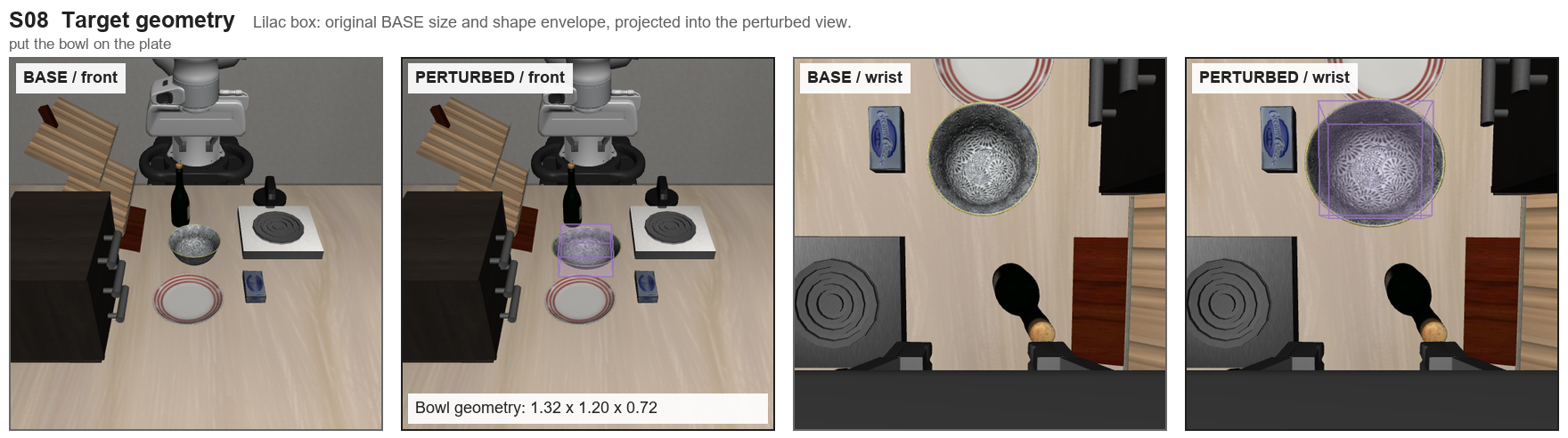}\par\vspace{-2pt}
  \includegraphics[width=0.88\textwidth]{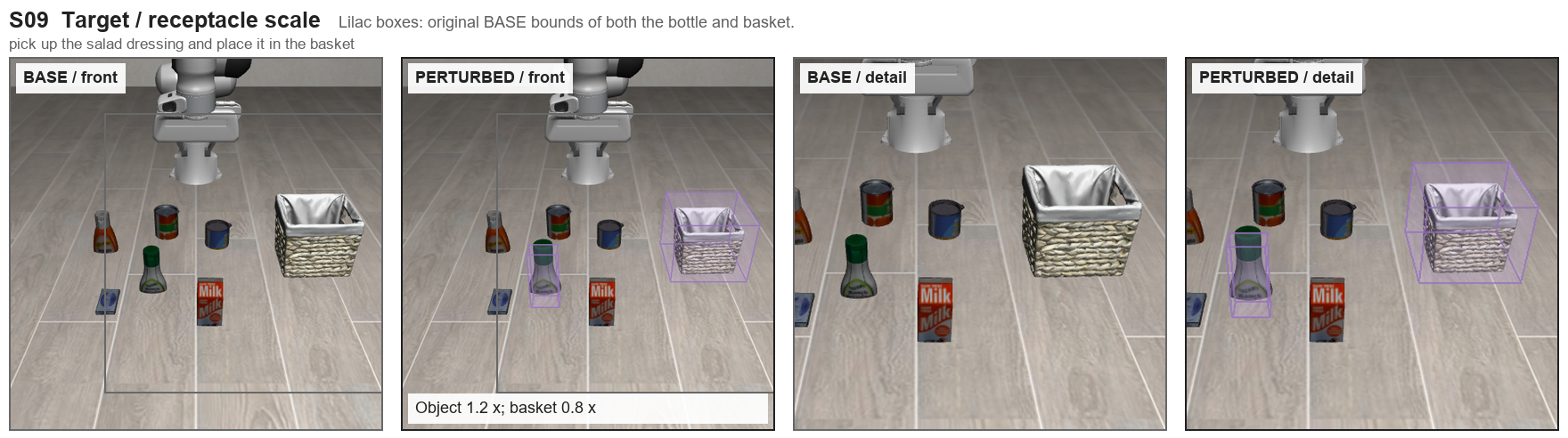}\par\vspace{-2pt}
  \includegraphics[width=0.88\textwidth]{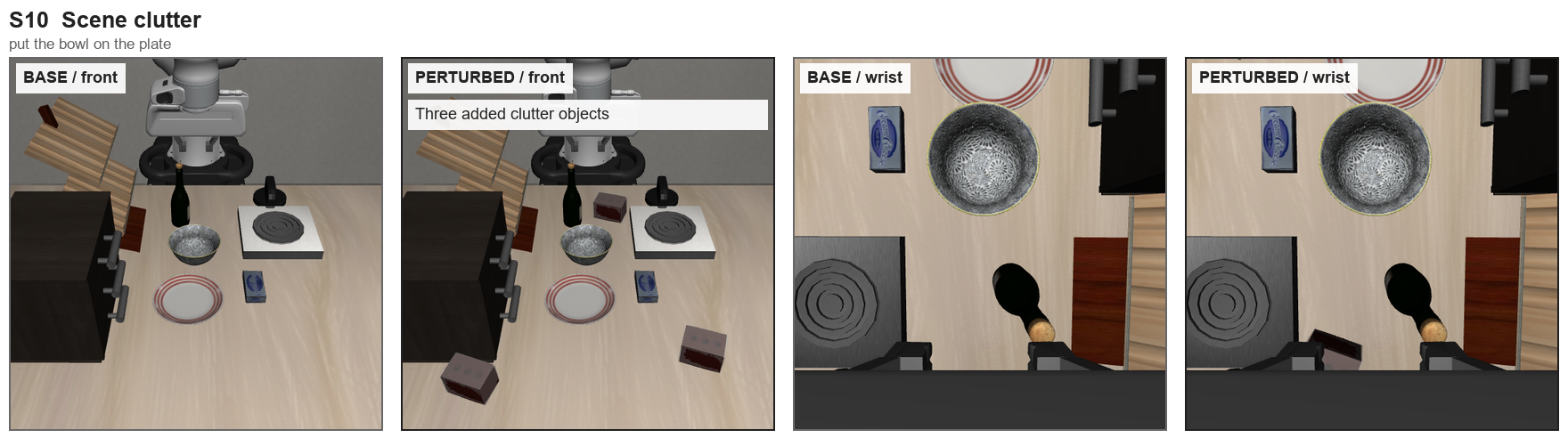}\par\vspace{-2pt}
  \caption{Static perturbations S06--S10.}
\end{figure}
\clearpage

\begin{figure}[!htp]
  \centering
  \includegraphics[width=0.88\textwidth]{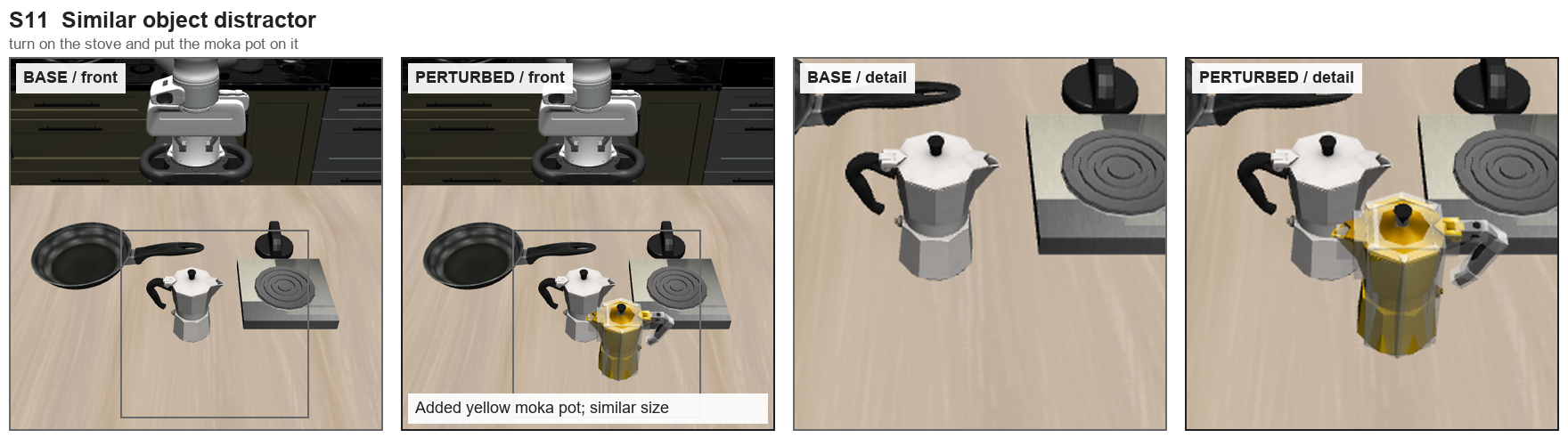}\par\vspace{-2pt}
  \includegraphics[width=0.88\textwidth]{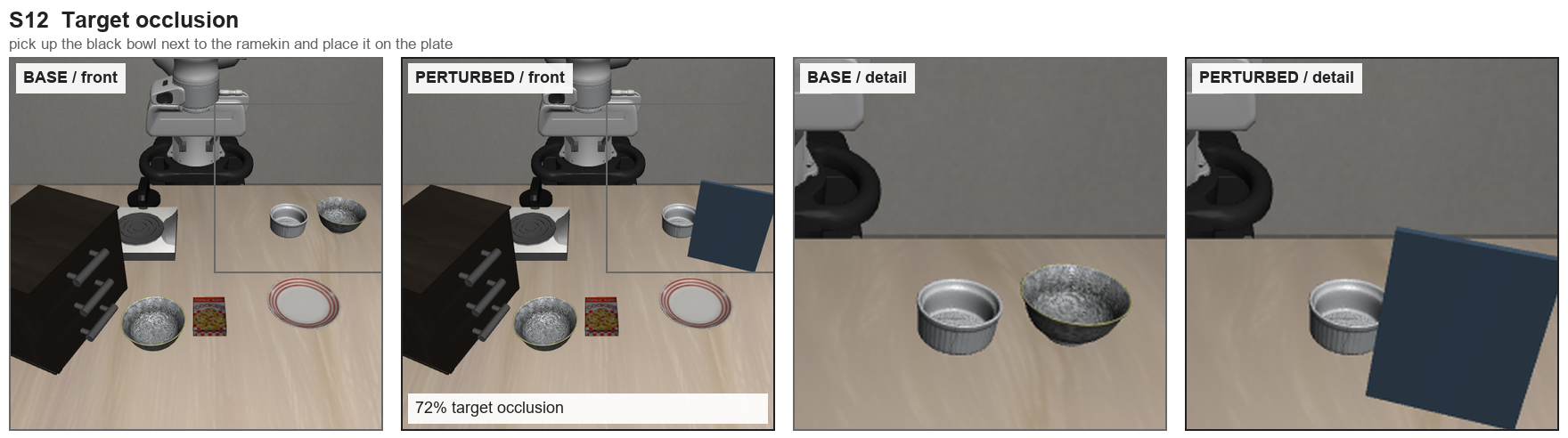}\par\vspace{-2pt}
  \includegraphics[width=0.88\textwidth]{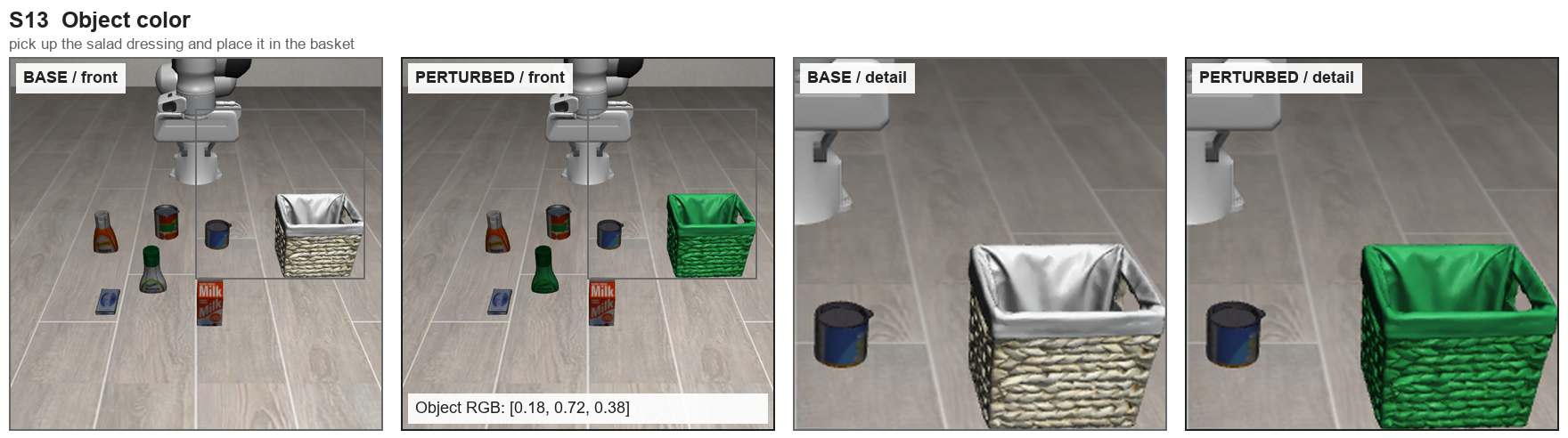}\par\vspace{-2pt}
  \includegraphics[width=0.88\textwidth]{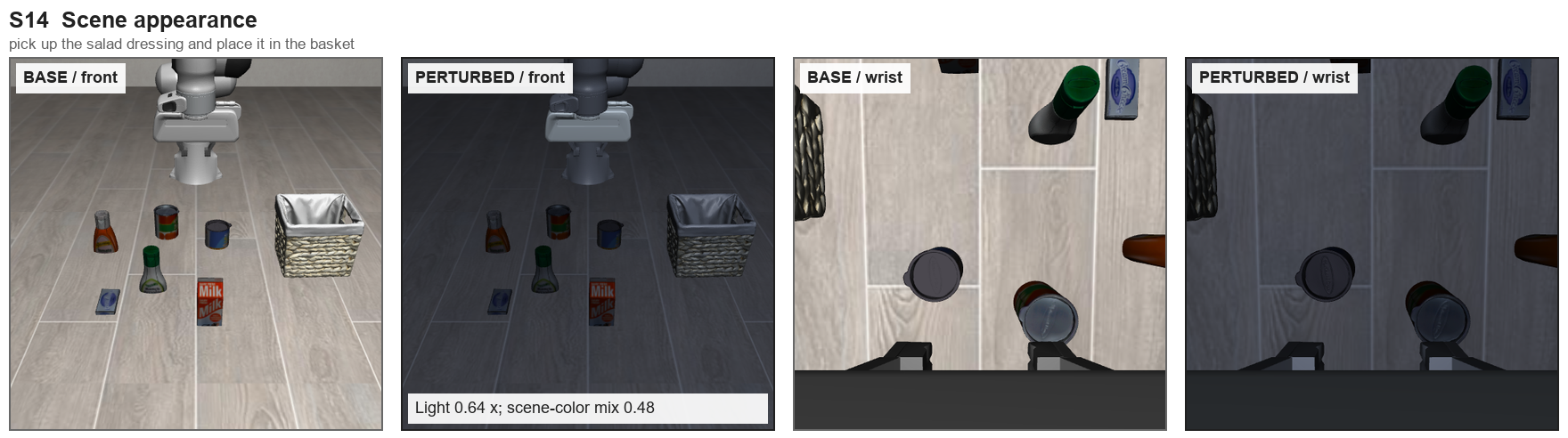}\par\vspace{-2pt}
  \caption{Static perturbations S11--S14.}
\end{figure}
\clearpage

\begin{figure}[!htp]
  \centering
  \includegraphics[width=0.88\textwidth]{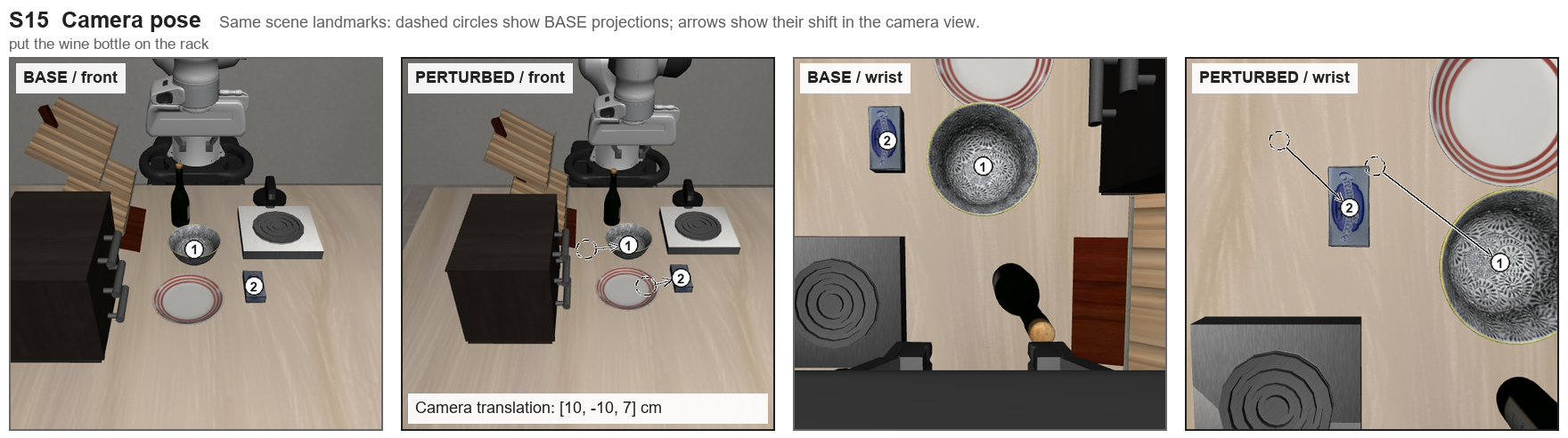}\par\vspace{-2pt}
  \includegraphics[width=0.88\textwidth]{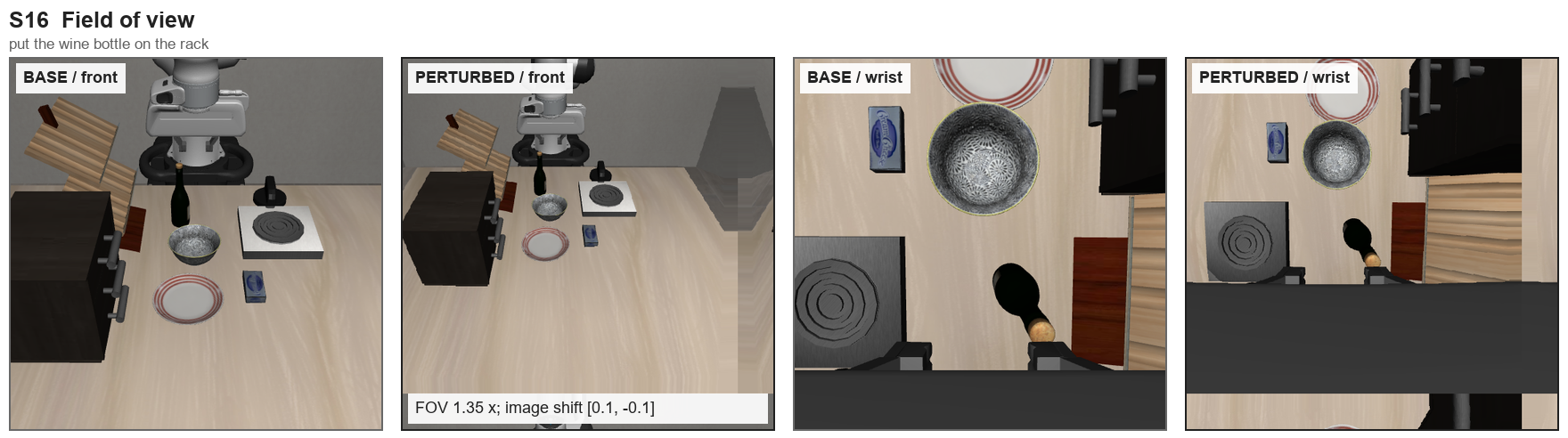}\par\vspace{-2pt}
  \includegraphics[width=0.88\textwidth]{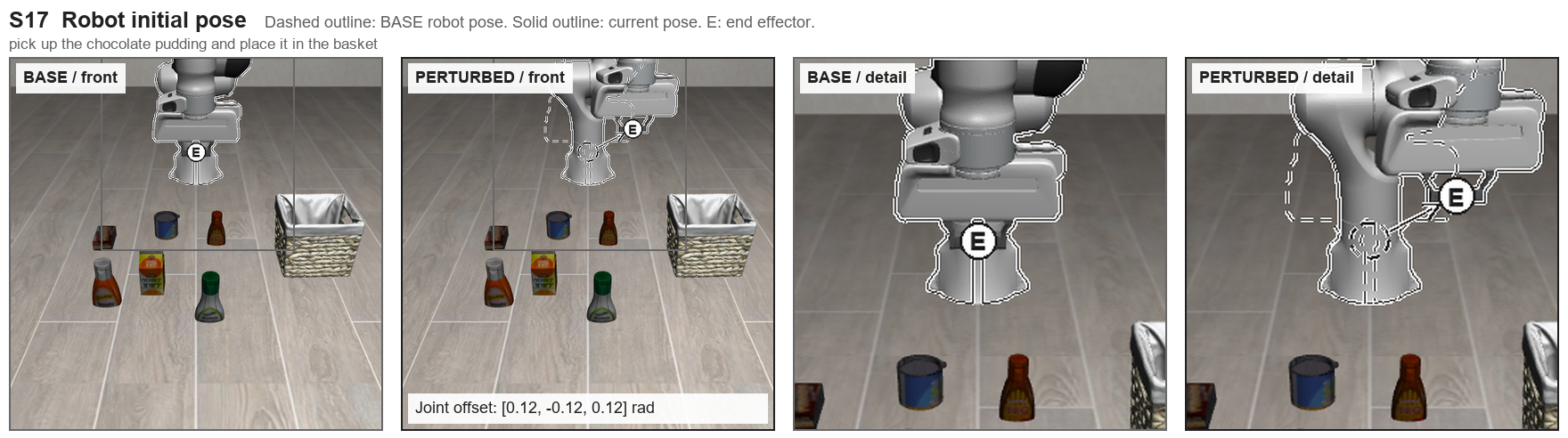}\par\vspace{-2pt}
  \includegraphics[width=0.88\textwidth]{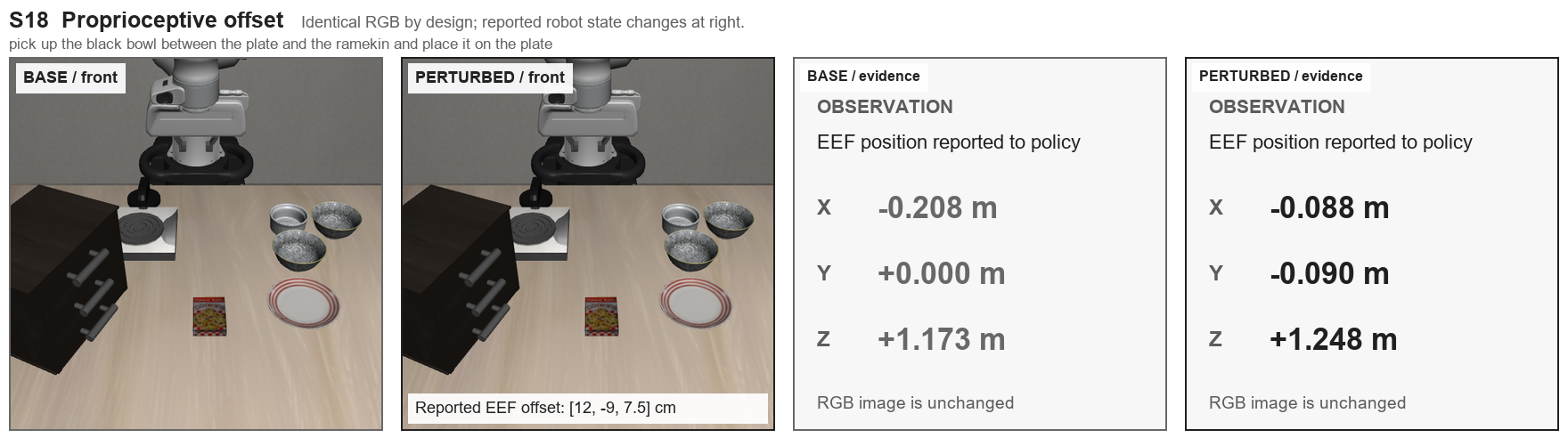}\par\vspace{-2pt}
  \caption{Static perturbations S15--S18.}
\end{figure}
\clearpage

\begin{figure}[!htp]
  \centering
  \includegraphics[width=0.88\textwidth]{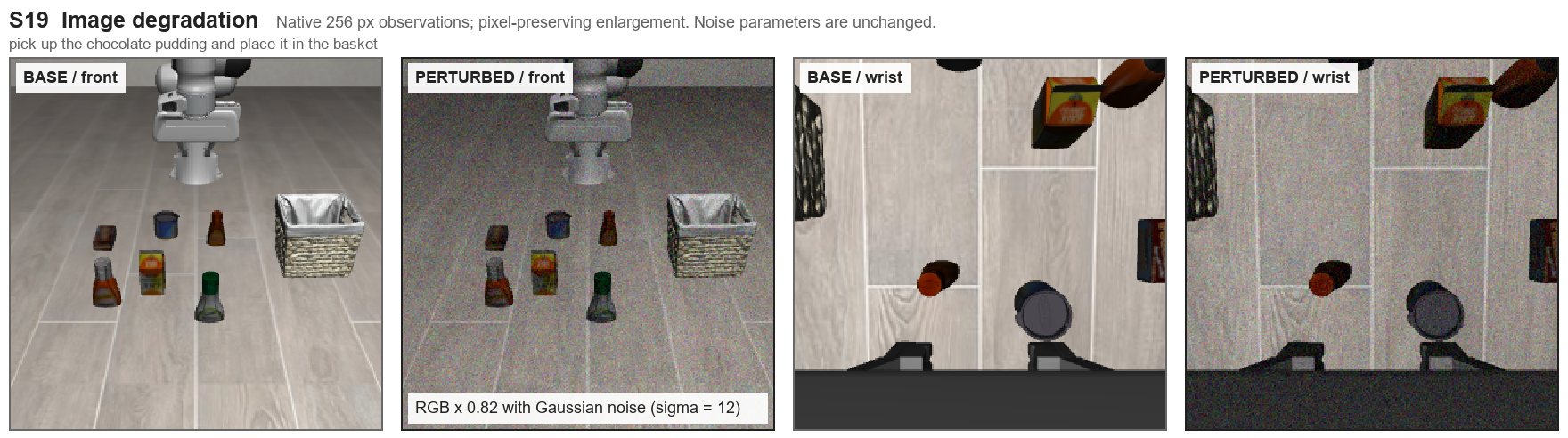}\par\vspace{-2pt}
  \includegraphics[width=0.88\textwidth]{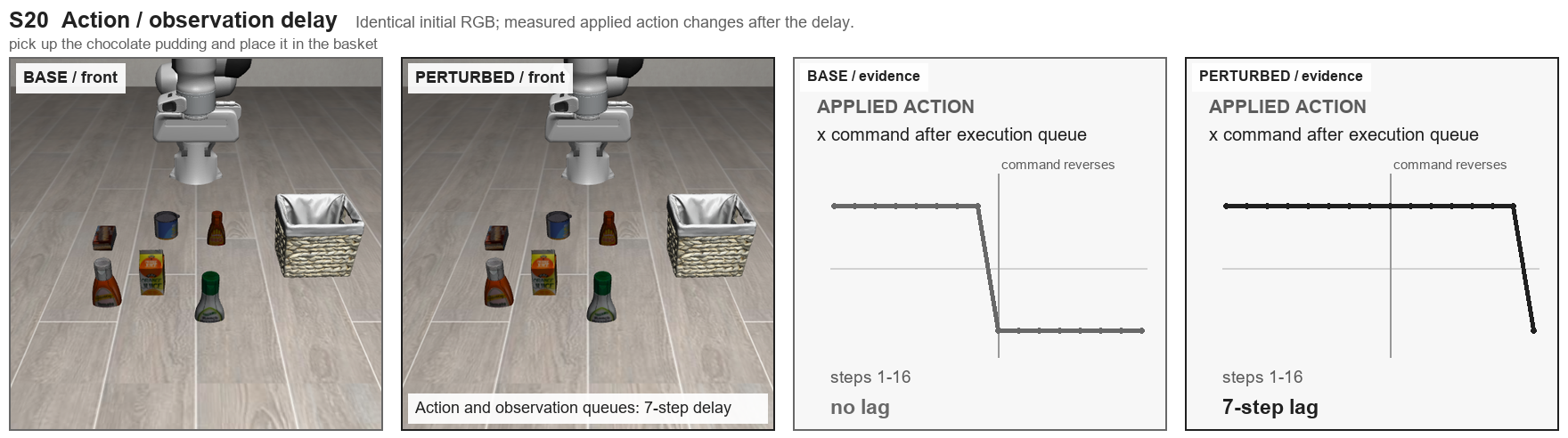}\par\vspace{-2pt}
  \includegraphics[width=0.88\textwidth]{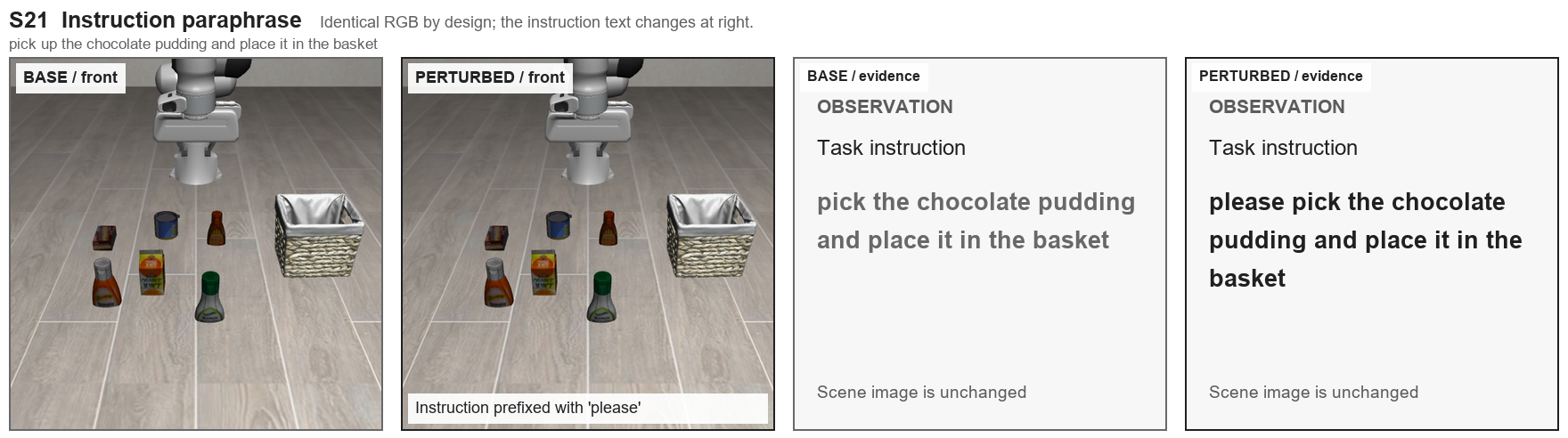}\par\vspace{-2pt}
  \includegraphics[width=0.88\textwidth]{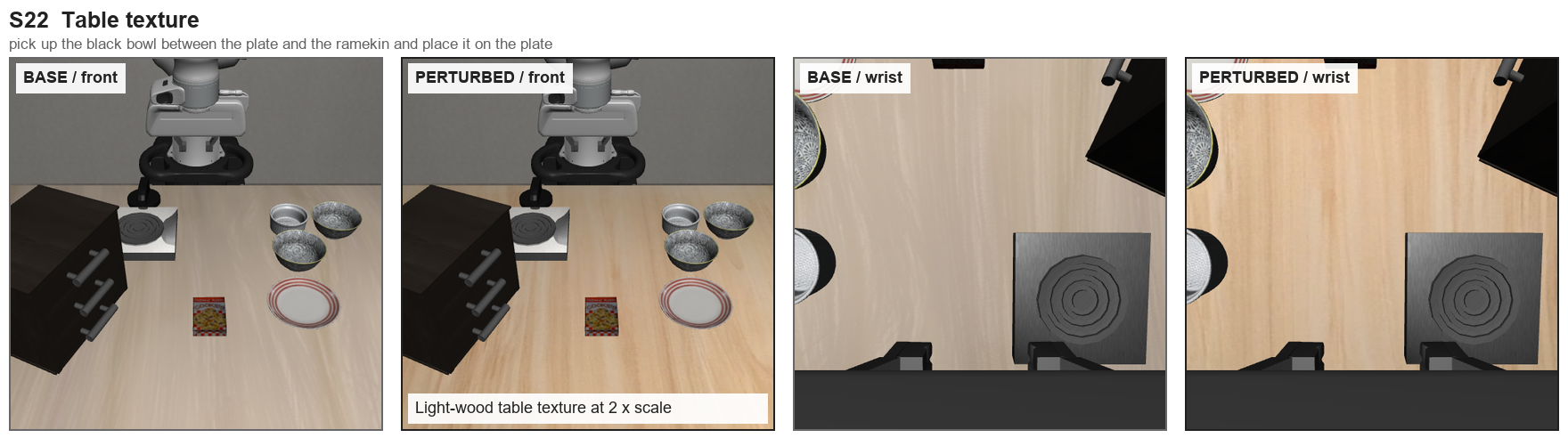}\par\vspace{-2pt}
  \caption{Static perturbations S19--S22.}
\end{figure}
\clearpage

\begin{figure}[!htp]
  \centering
  \includegraphics[width=0.88\textwidth]{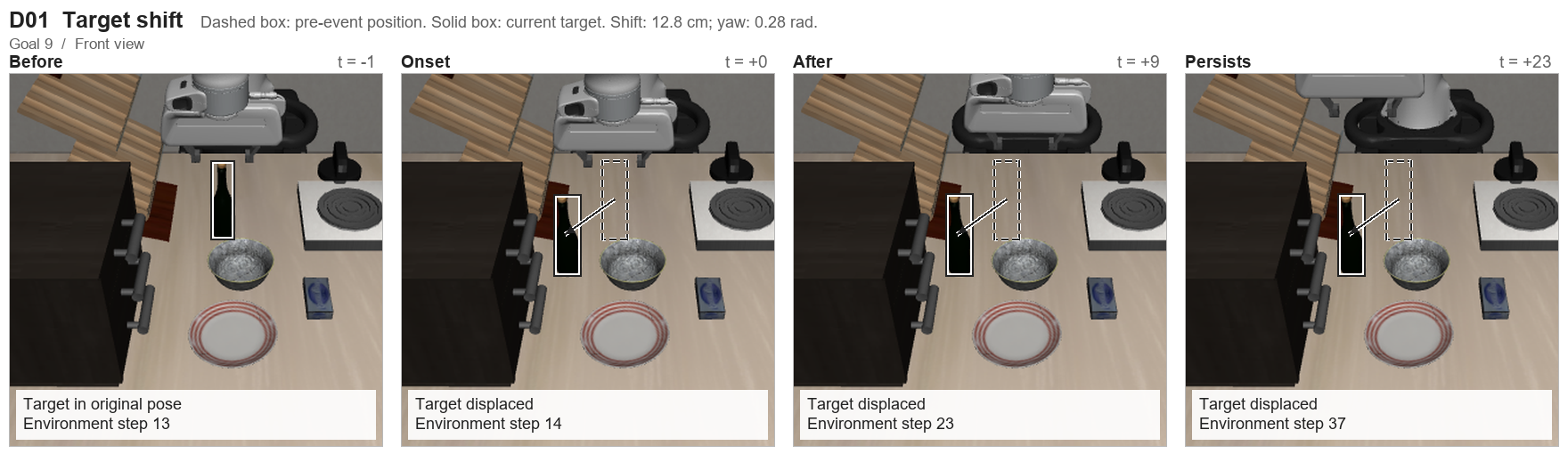}\par\vspace{-2pt}
  \includegraphics[width=0.88\textwidth]{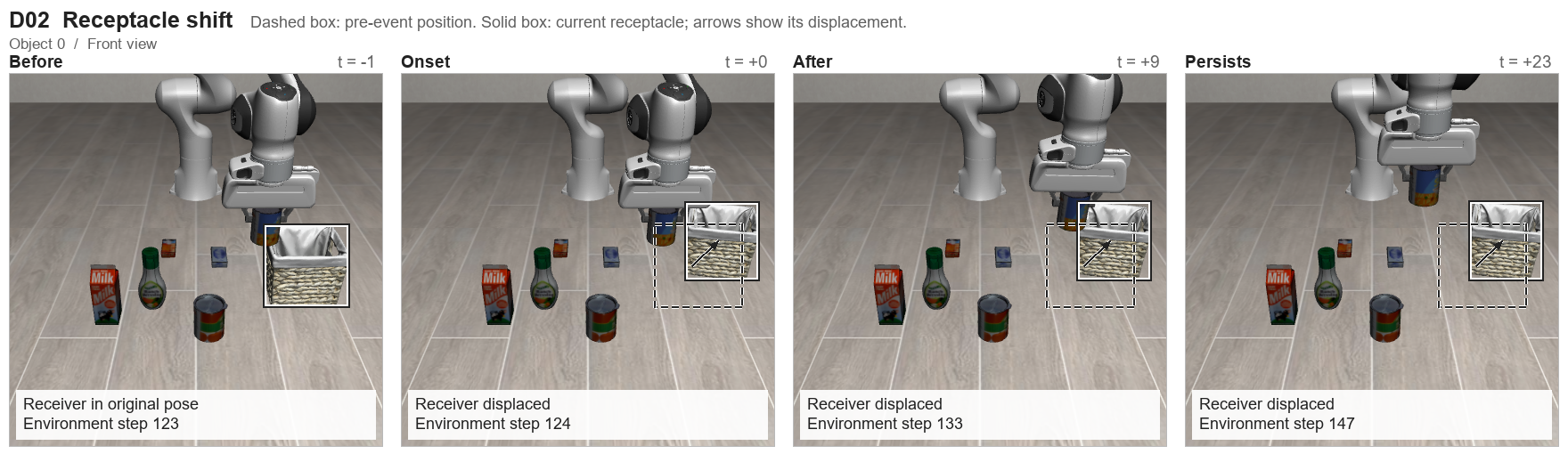}\par\vspace{-2pt}
  \includegraphics[width=0.88\textwidth]{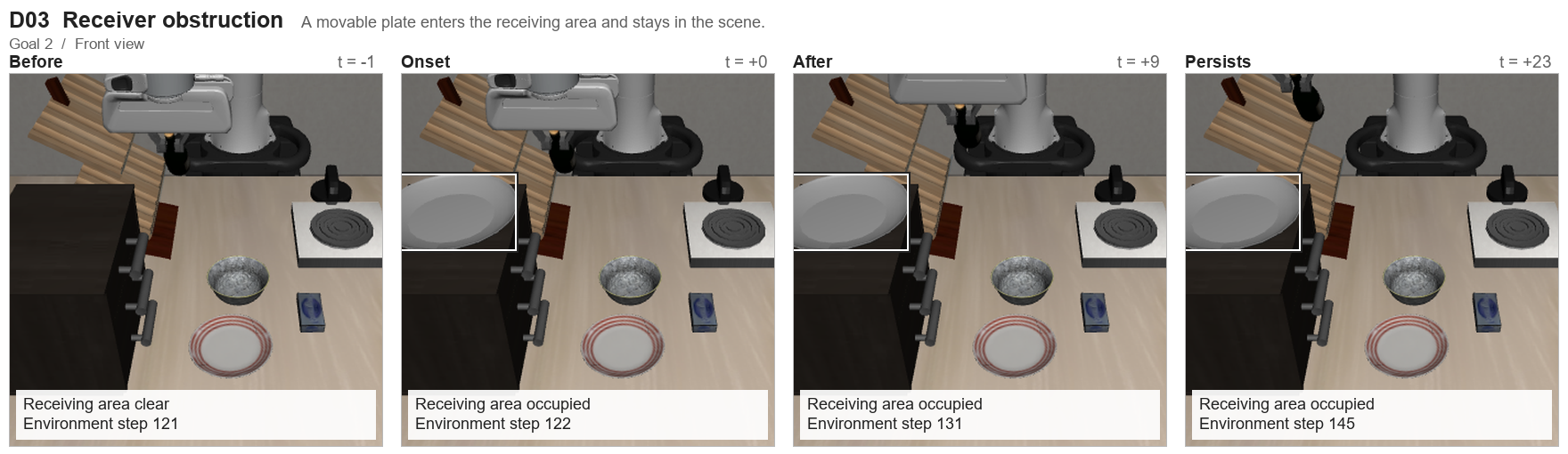}\par\vspace{-2pt}
  \includegraphics[width=0.88\textwidth]{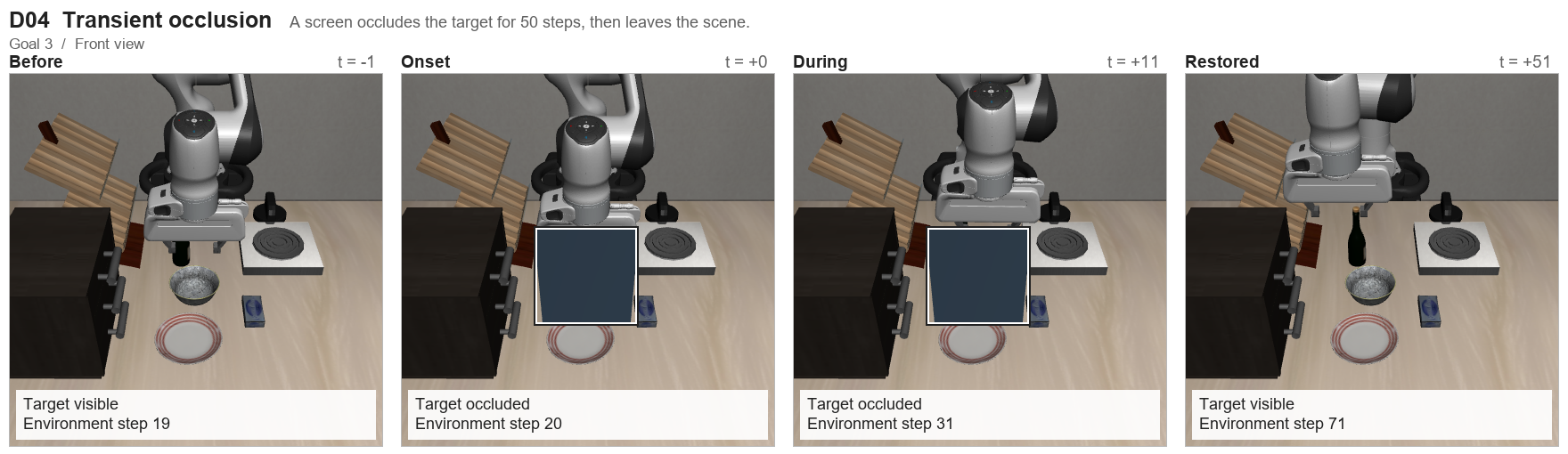}\par\vspace{-2pt}
  \includegraphics[width=0.88\textwidth]{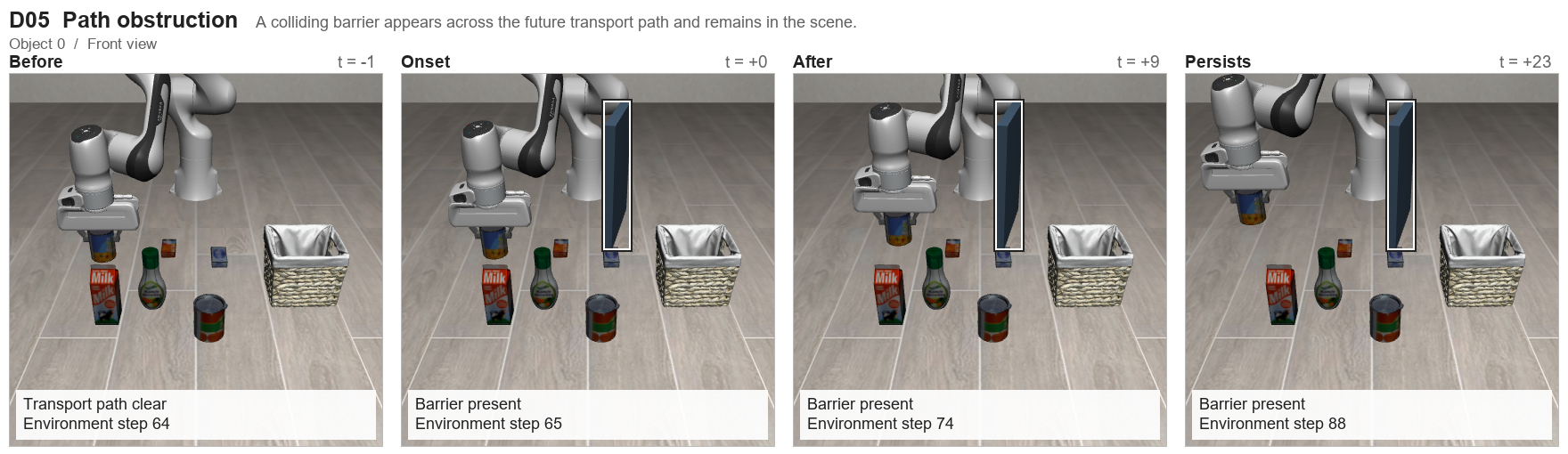}\par\vspace{-2pt}
  \caption{Dynamic perturbations D01--D05. Each row shows the temporal sequence around an intervention, using the views indicated in the panels.}
\end{figure}
\clearpage

\begin{figure}[!htp]
  \centering
  \includegraphics[width=0.88\textwidth]{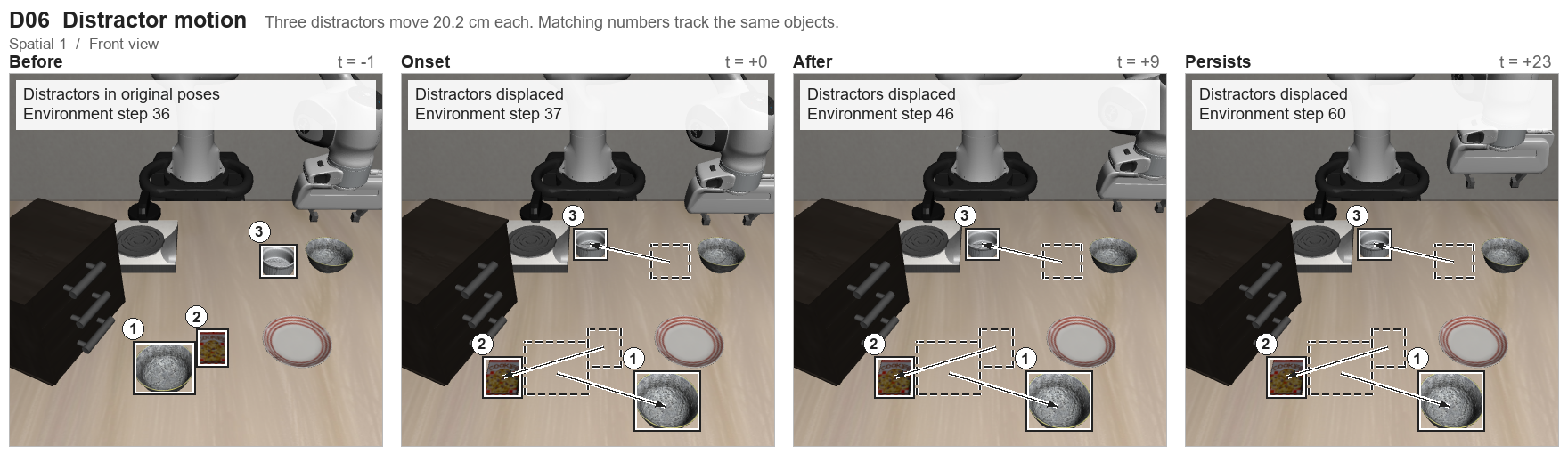}\par\vspace{-2pt}
  \includegraphics[width=0.88\textwidth]{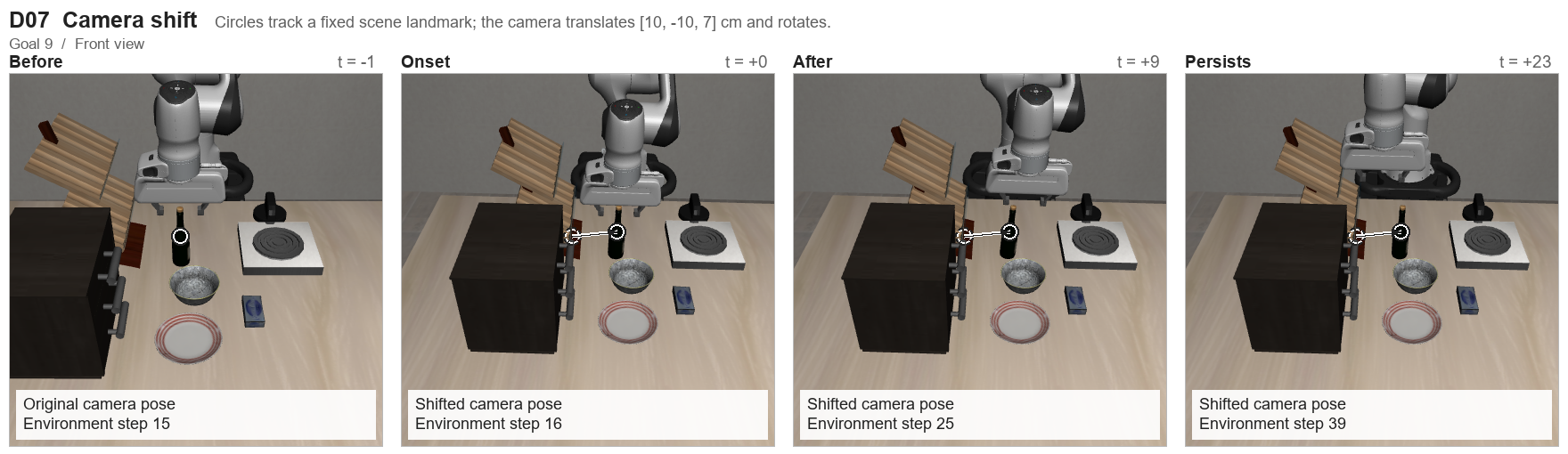}\par\vspace{-2pt}
  \includegraphics[width=0.88\textwidth]{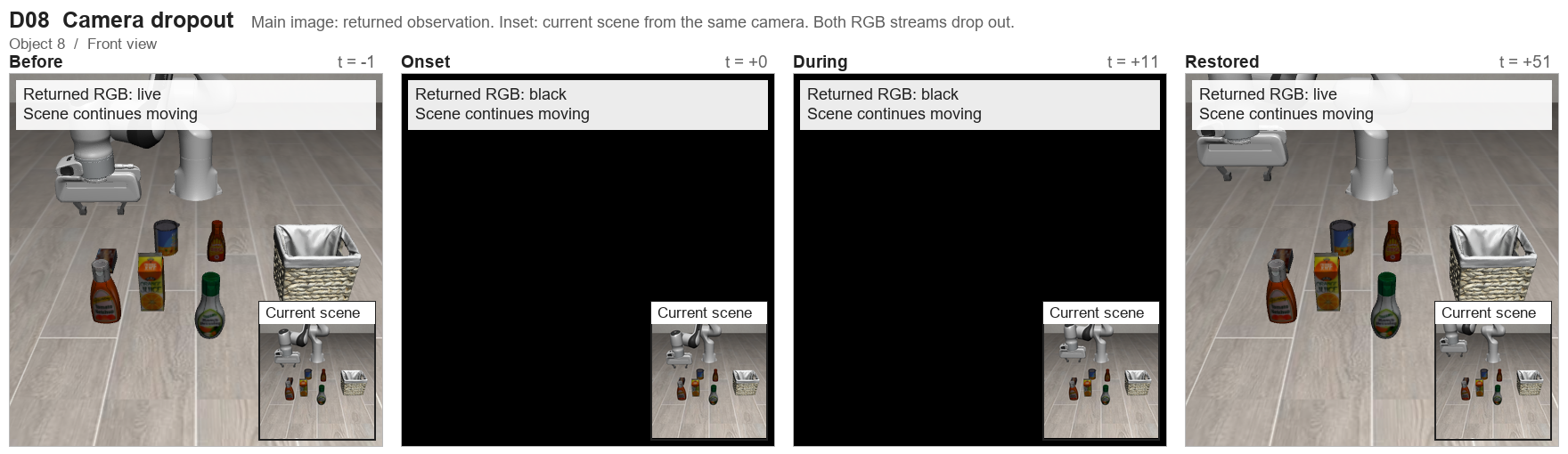}\par\vspace{-2pt}
  \includegraphics[width=0.88\textwidth]{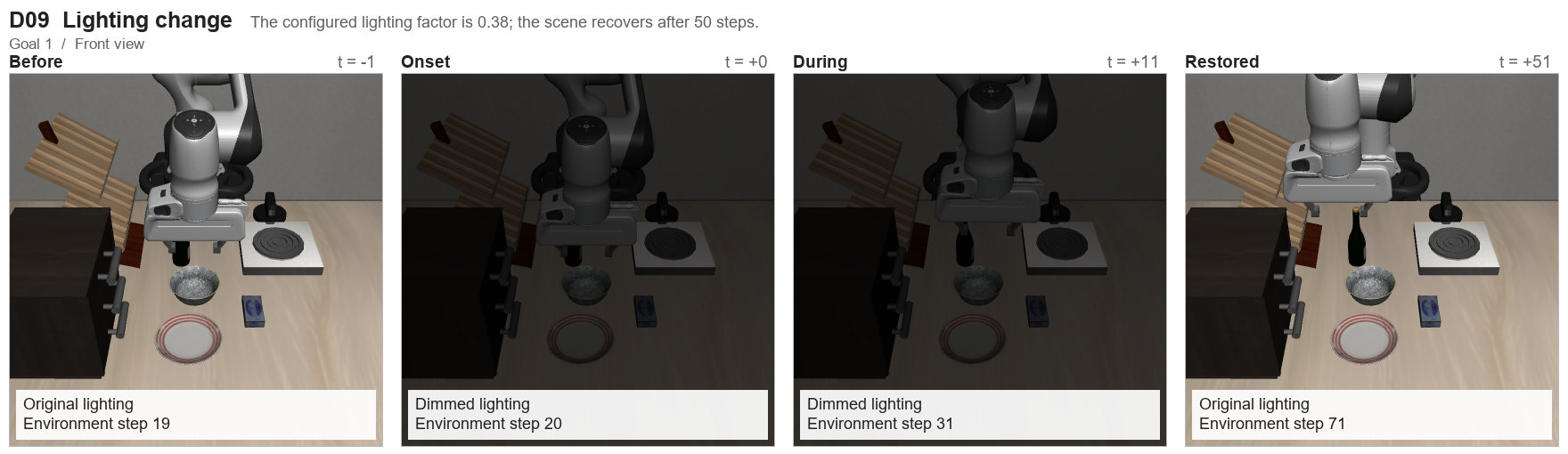}\par\vspace{-2pt}
  \includegraphics[width=0.88\textwidth]{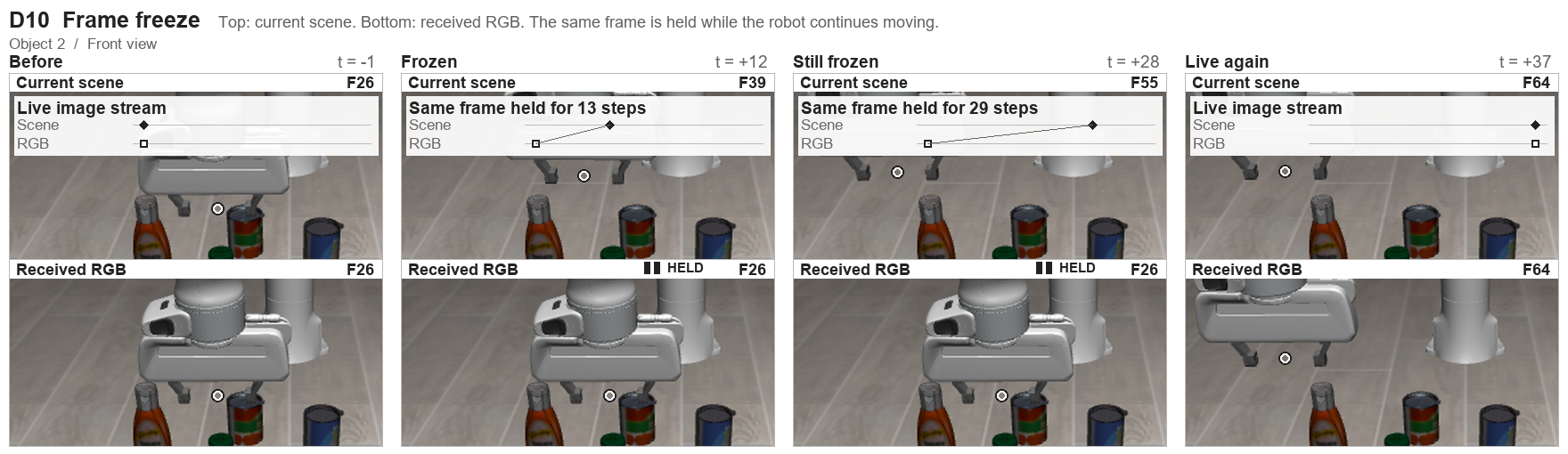}\par\vspace{-2pt}
  \caption{Dynamic perturbations D06--D10.}
\end{figure}
\clearpage

\begin{figure}[!htp]
  \centering
  \includegraphics[width=0.88\textwidth]{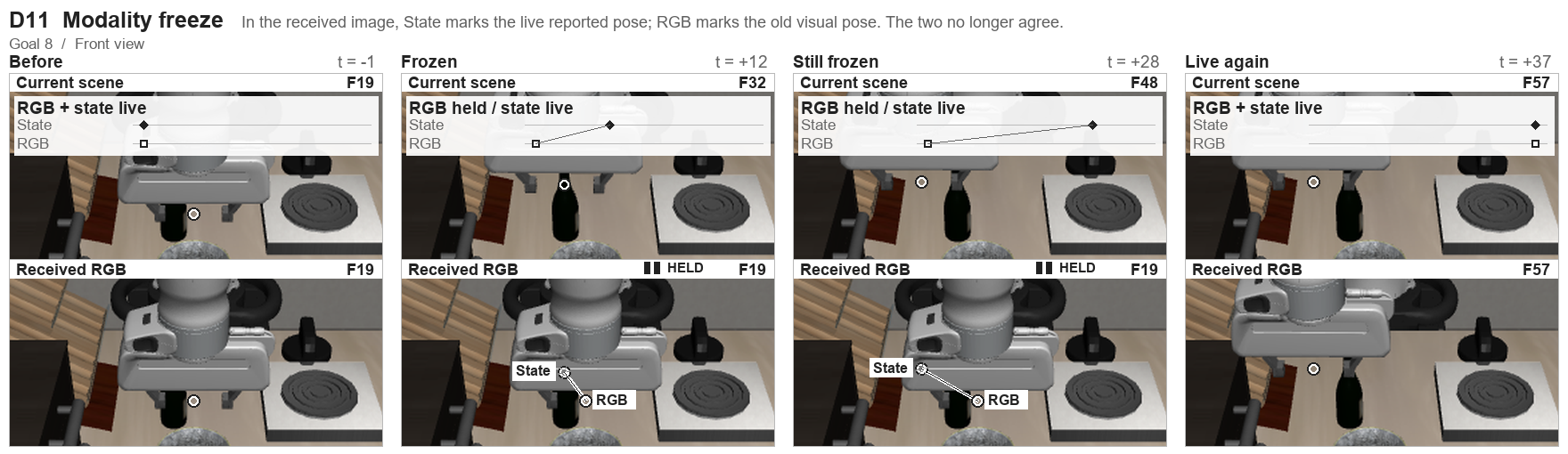}\par\vspace{-2pt}
  \includegraphics[width=0.88\textwidth]{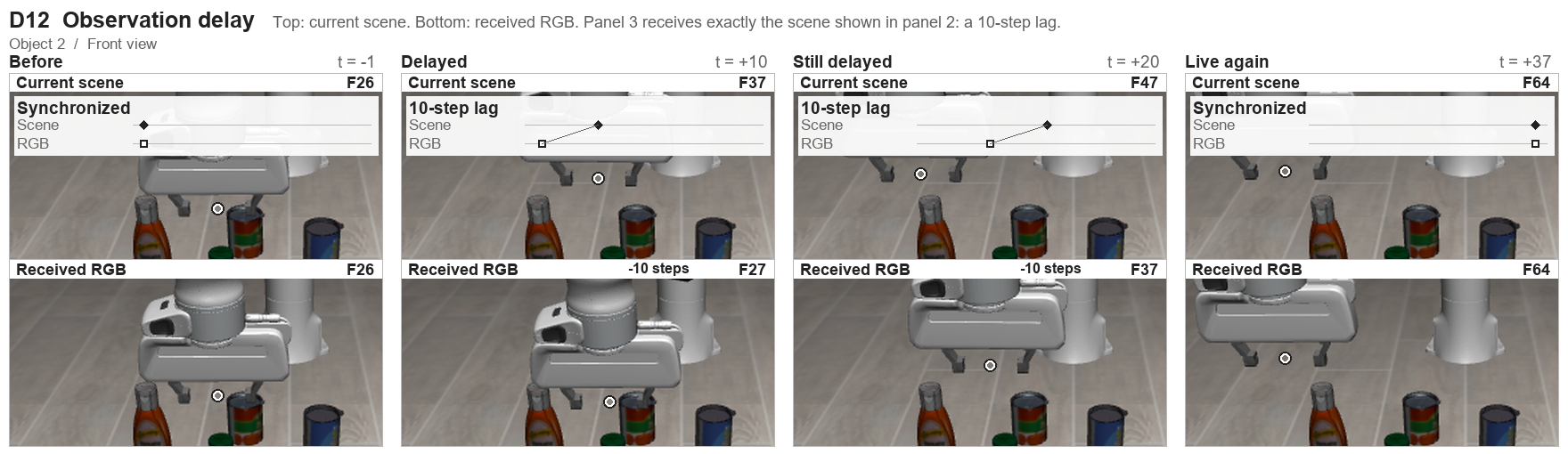}\par\vspace{-2pt}
  \includegraphics[width=0.88\textwidth]{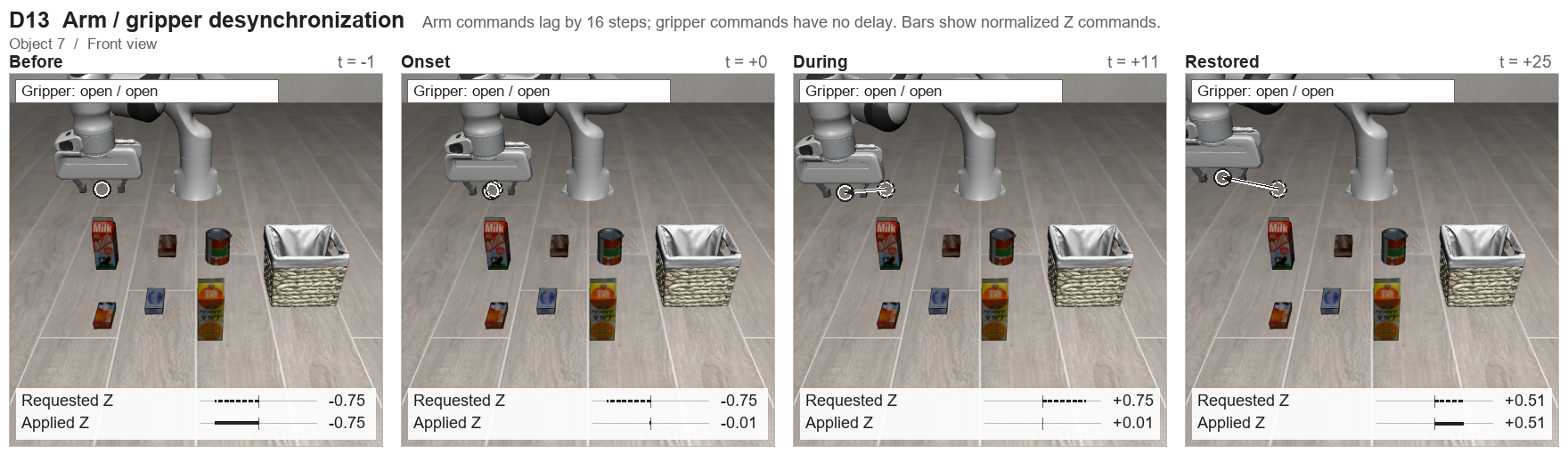}\par\vspace{-2pt}
  \includegraphics[width=0.88\textwidth]{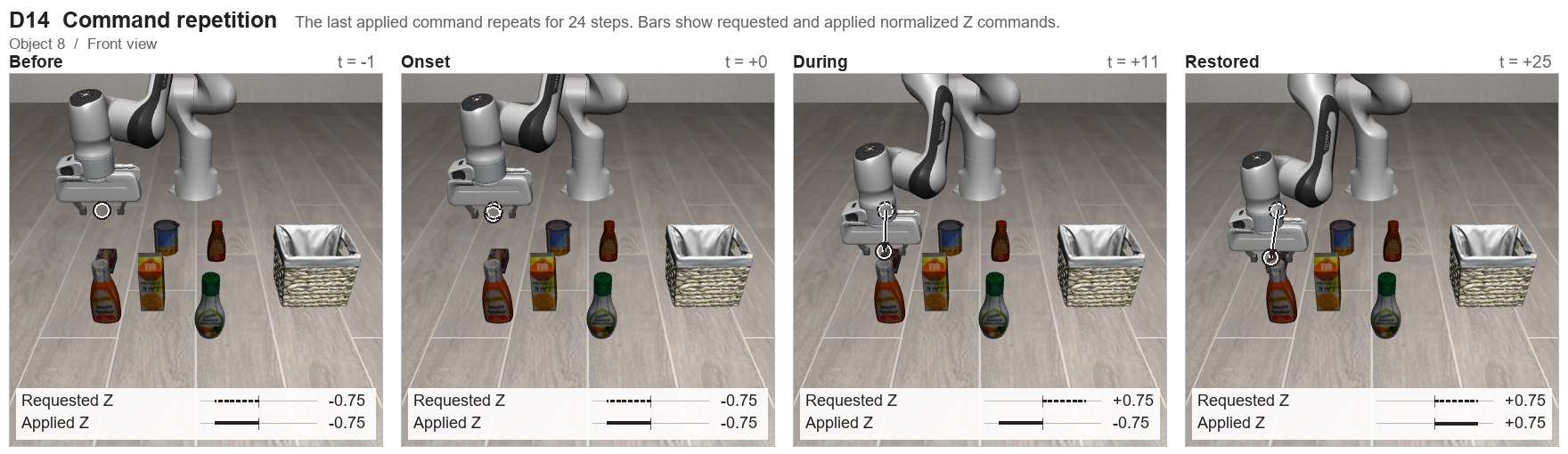}\par\vspace{-2pt}
  \includegraphics[width=0.88\textwidth]{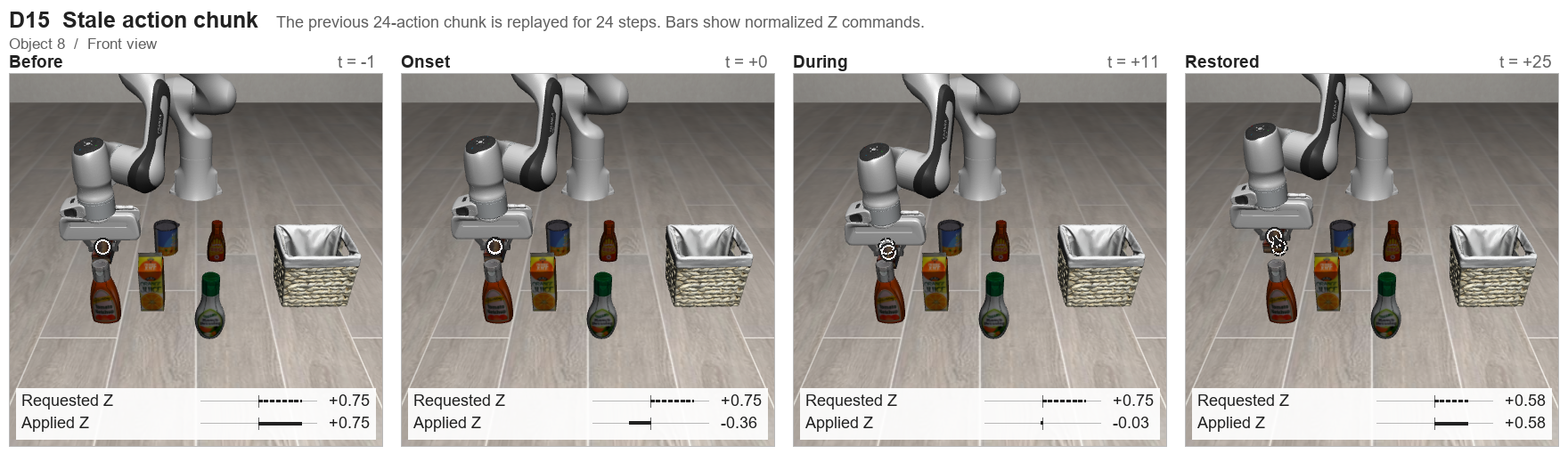}\par\vspace{-2pt}
  \caption{Dynamic perturbations D11--D15.}
\end{figure}
\clearpage

\begin{figure}[!htp]
  \centering
  \includegraphics[width=0.88\textwidth]{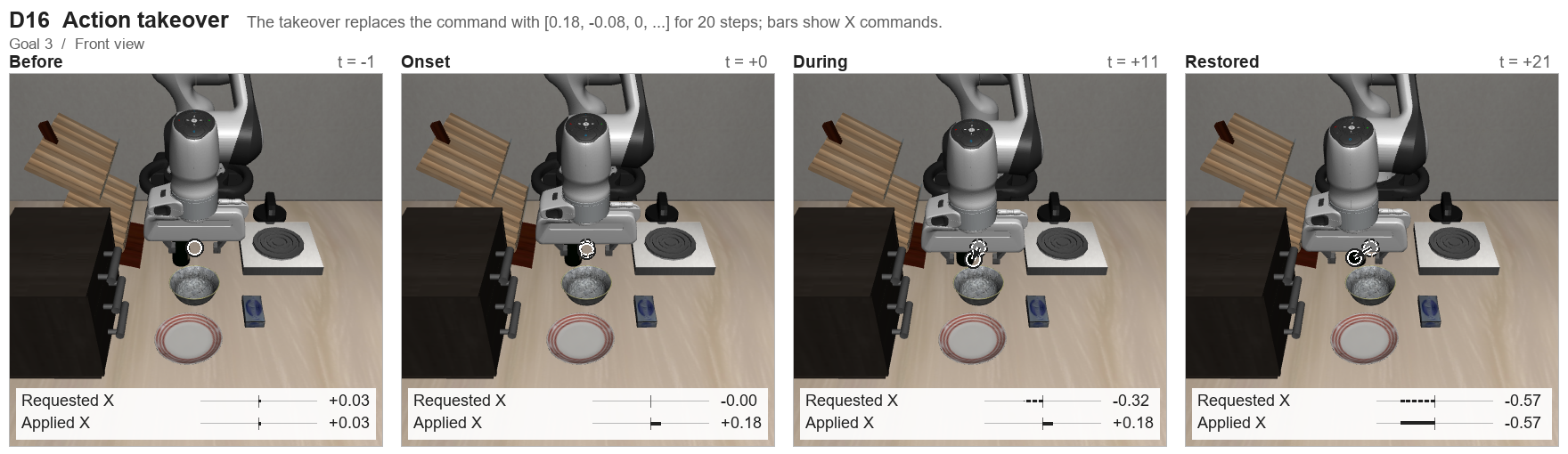}\par\vspace{-2pt}
  \includegraphics[width=0.88\textwidth]{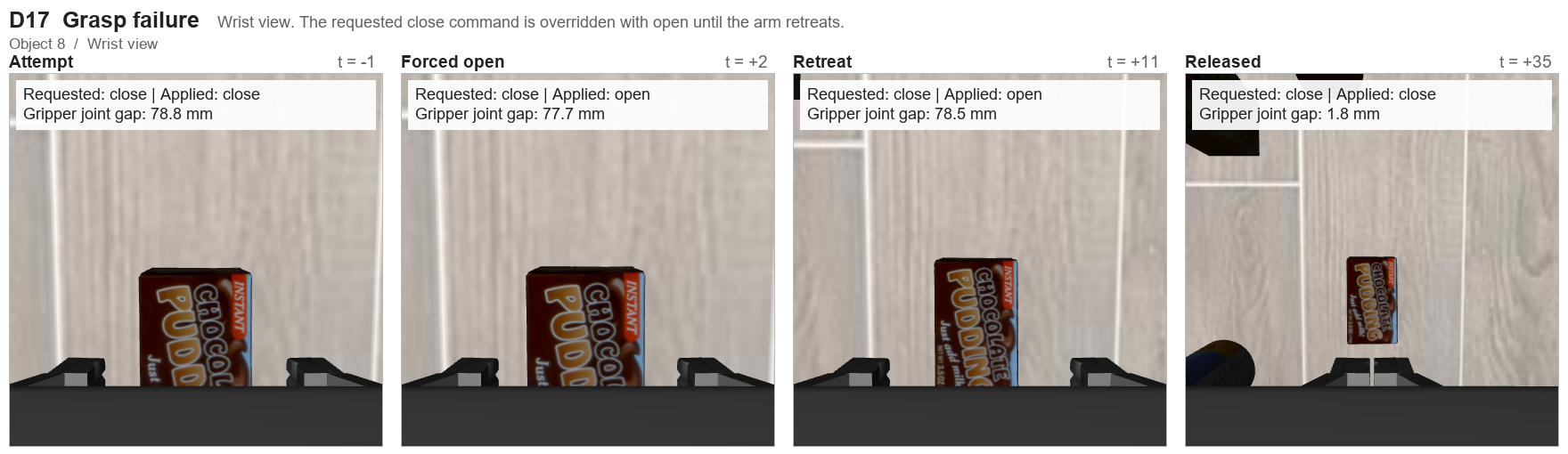}\par\vspace{-2pt}
  \includegraphics[width=0.88\textwidth]{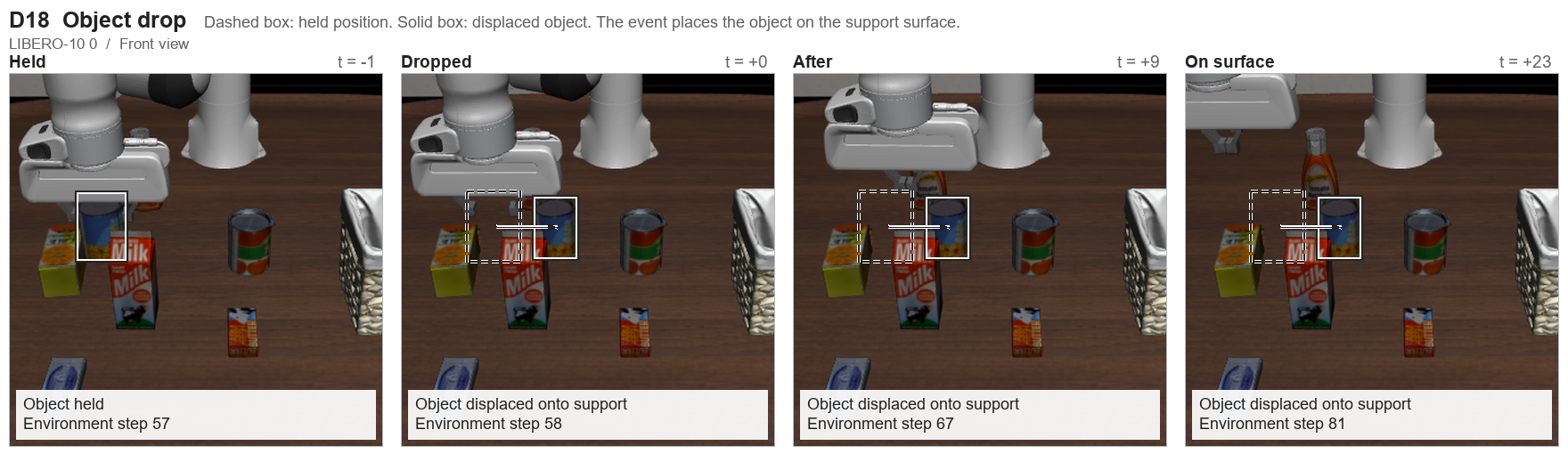}\par\vspace{-2pt}
  \includegraphics[width=0.88\textwidth]{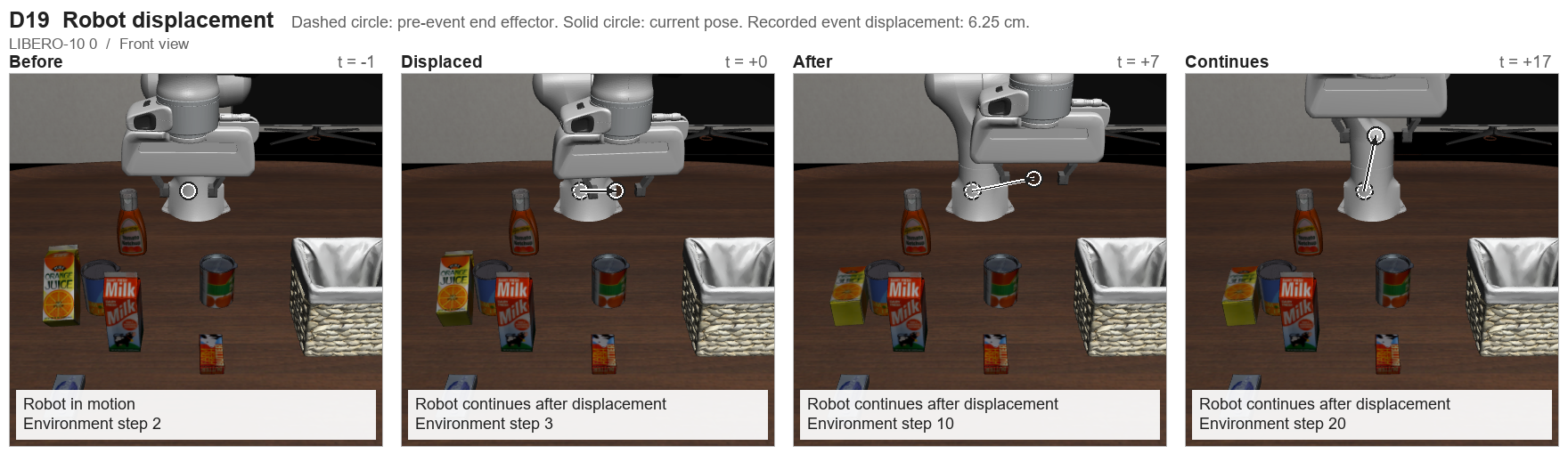}\par\vspace{-2pt}
  \includegraphics[width=0.88\textwidth]{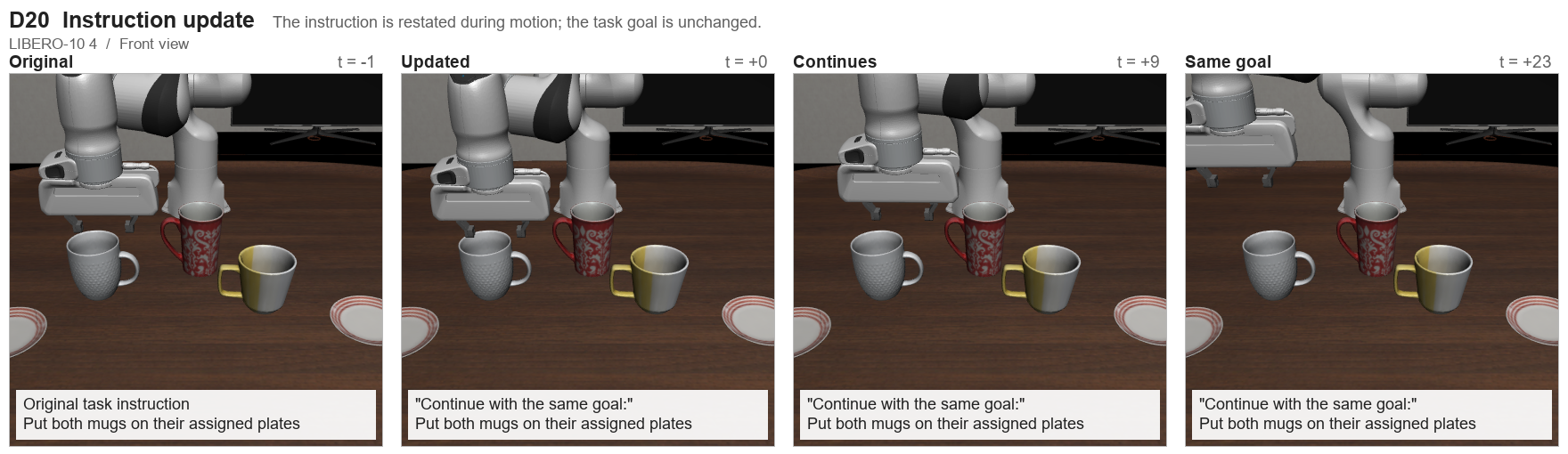}\par\vspace{-2pt}
  \caption{Dynamic perturbations D16--D20.}
\end{figure}
\clearpage

\begin{figure}[!htbp]
    \centering
    \includegraphics[width=\textwidth]{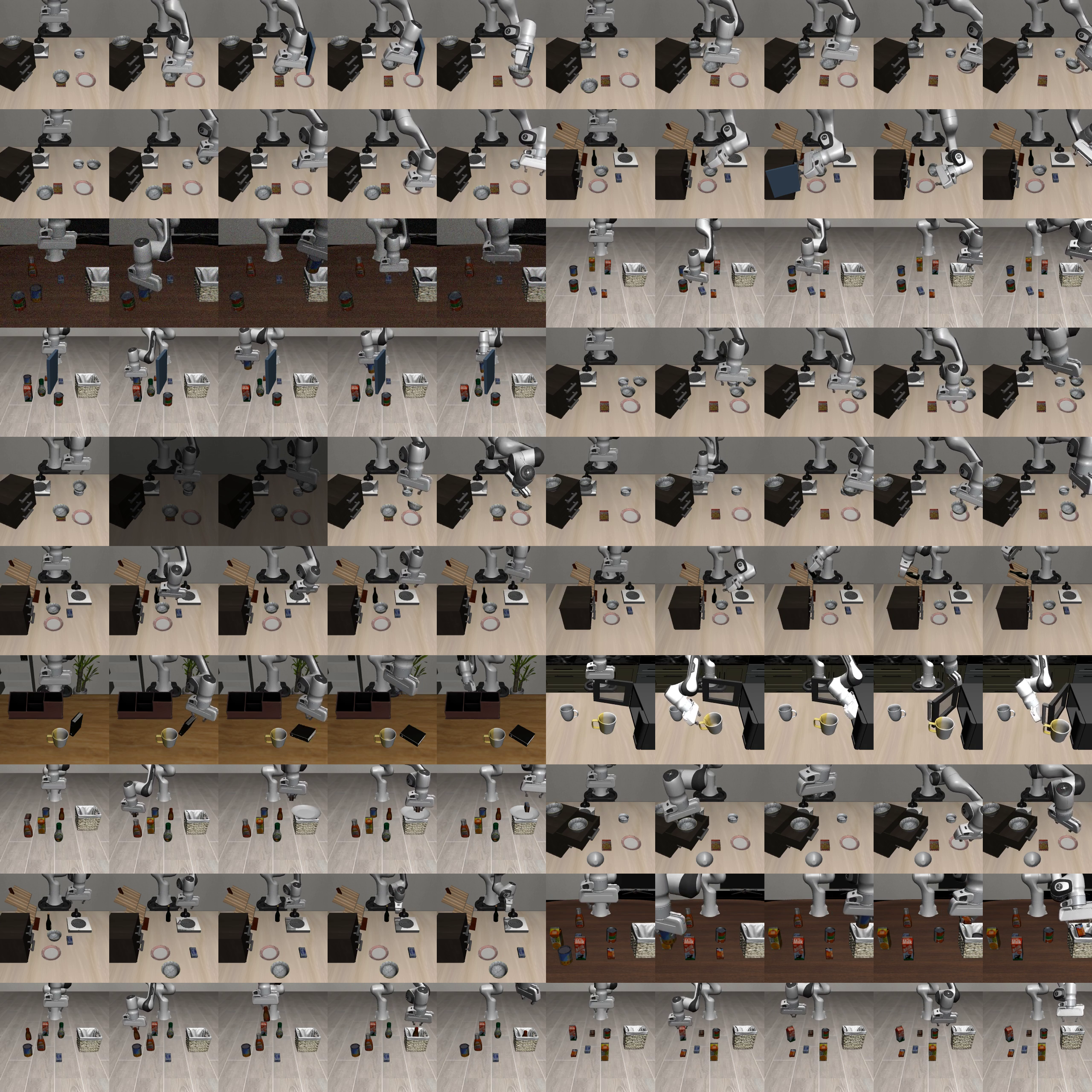}
    \caption{Representative failure rollouts under perturbations; each row contains two five-frame sequences, read from left to right.}
    \label{fig:appendix_failure_rollouts}
\end{figure}
\clearpage

\begingroup
\normalfont
\normalsize
\newlength{\LPTableWidth}

\section{Perturbation Catalogue}
\label{app:perturbation_catalogue}

This section supplements the LIBERO-Pro2.0 results with definitions of all
22 static and 20 dynamic perturbations and the task partition used in the
controlled eight-task studies. The examples in Appendix~\ref{app:perturbation_visualizations}
illustrate these definitions; the nominal, unperturbed task is the BASE reference.

\paragraph{Design scope.}
Perturbations cover object and scene geometry, appearance, sensing, temporal
alignment, action execution, robot state, and language. Each type shares a
semantic definition across suites; affected objects, magnitudes, directions,
and event settings are specified per task. These parameter choices do not
define additional perturbation types.

\paragraph{Task validity.}
The original goal and success predicate are preserved. Geometry and state
changes retain reachable, collision-free configurations and feasible task
completion. Planar pose changes combine horizontal translation and yaw;
pitch and roll are excluded from this category. Physical parameters such as
mass, inertia, friction, stiffness, and damping are not varied. Grasp loss and
robot displacement are represented by injected outcome states or bounded
control interventions, rather than simulated impact or falling forces.

\paragraph{Initialization and runtime events.}
Static perturbations are introduced at reset; altered scene or fault settings
persist while normal robot and object motion remains possible. Dynamic
perturbations are triggered at task-relevant phases, such as approach,
grasp, transport, or placement. Trigger conditions, duration, and release rules
are specified by each case. Restoring a sensor or control channel does not
imply rolling back the complete scene. Experiment-specific freeze or waiting
durations are distinct from these general mechanism definitions.

\Needspace{10\baselineskip}

\subsection{Category Definitions and Assignment Principle}
\label{app:category_mapping}

We assign each perturbation to one category according to its primary
intervention target. The categories organize perturbations by their
dominant operational effect and are not mutually exclusive descriptions
of every downstream consequence.
\textit{Condition} perturbations modify task-relevant object or receiver
configurations, relative layouts, or geometric constraints on approach
and placement.
\textit{Environment} perturbations modify object instances or scene
context, including clutter, distractors, occlusion, visual-domain
appearance, support-surface appearance, and external visual conditions
such as camera motion or lighting changes.
\textit{Observation} perturbations modify the formation, quality,
availability, or synchronization of the inputs delivered to the policy.
\textit{Execution} perturbations modify control timing or action
execution.
\textit{Robot} perturbations modify robot initialization, calibration,
or robot-related state and manipulation outcomes.
\textit{Language} perturbations modify instruction wording while
preserving task semantics.
Some perturbations may affect multiple aspects of task execution; each
is assigned according to its primary intervention target. This mapping
is used consistently to compute the category averages in Tables~1--3.

\begin{table}[htb]
    \centering
    \small
    \caption{Mapping of static and runtime perturbations to evaluation
    categories.}
    \label{tab:category_mapping}
    \begin{tabular}{lll}
        \toprule
        Category & Static perturbations & Runtime perturbations \\
        \midrule
        Condition
            & S01--S05, S07--S09
            & D01--D06 \\
        Environment
            & S06, S10--S14, S22
            & D07--D09 \\
        Observation
            & S15--S16, S19
            & D10--D13 \\
        Execution
            & S20
            & D14--D16 \\
        Robot
            & S17--S18
            & D17--D19 \\
        Language
            & S21
            & D20 \\
        \bottomrule
    \end{tabular}
\end{table}

\subsection{Static Perturbations}
\label{app:static_perturbation_catalogue}

Table~\ref{tab:static_perturbation_design} summarizes each initialization-time
intervention and the constraints that preserve the manipulation task.
\begingroup
\setlength{\tabcolsep}{4pt}
\renewcommand{\arraystretch}{1.08}
\setlength{\LTleft}{0pt}
\setlength{\LTright}{0pt plus 1fill}
\setlength{\LTcapwidth}{\linewidth}
\setlength{\LPTableWidth}{\dimexpr\linewidth-6\tabcolsep\relax}
\begin{longtable}{@{}>{\raggedright\arraybackslash}p{0.065\LPTableWidth}>{\raggedright\arraybackslash}p{0.225\LPTableWidth}>{\raggedright\arraybackslash}p{0.39\LPTableWidth}>{\raggedright\arraybackslash}p{0.32\LPTableWidth}@{}}
\caption{Static perturbation definitions. Interventions are introduced at reset. Modified geometry or fault settings persist; initial-state changes do not prevent normal subsequent motion. Alternatives within a row are selected by the task configuration.}\label{tab:static_perturbation_design}\\
\toprule
\textbf{ID} & \textbf{Perturbation} & \textbf{Intervention} & \textbf{Constraint} \\
\midrule
\endfirsthead
\multicolumn{4}{l}{\tablename\ \thetable\ (continued)}\\
\toprule
\textbf{ID} & \textbf{Perturbation} & \textbf{Intervention} & \textbf{Constraint} \\
\midrule
\endhead
\midrule
\multicolumn{4}{r}{Continued on the next page}\\
\endfoot
\bottomrule
\endlastfoot
S01 & Target planar pose & Translate the target in the horizontal plane and rotate it in yaw. & Preserve the target identity and intended goal. \\
\addlinespace[2pt]
S02 & Receiver planar pose & Translate or rotate the receiver to change the placement destination and approach. & Preserve receiver identity and task goal. \\
\addlinespace[2pt]
S03 & Receiver or support height & Change receiver or support elevation using a support platform or a coupled layout adjustment. & Retain a collision-free initial layout and feasible placement. \\
\addlinespace[2pt]
S04 & Object position swap & Swap the planar positions of two objects; use a safe layout offset when a direct swap is unsuitable. & Keep object identities, task roles, and goal unchanged. \\
\addlinespace[2pt]
S05 & Restricted approach or path & Add geometry that constrains approach or transport. & Retain valid inverse kinematics and a complete collision-free route. \\
\addlinespace[2pt]
S06 & Instance replacement & Replace the target with a same-class or unseen asset specified by the task configuration. & Preserve semantic class and graspability. \\
\addlinespace[2pt]
S07 & Functional-part geometry & Change an opening, rim, or receiving region used for grasping or placement. & Retain the intended functional role. \\
\addlinespace[2pt]
S08 & Size or aspect ratio & Apply uniform or anisotropic target scaling. & Change geometric extent while preserving target identity. \\
\addlinespace[2pt]
S09 & Opening clearance and pose & Jointly change object/container geometry and the relative insertion configuration. & Use collision geometry with full containment, positive clearance, and sufficient depth. \\
\addlinespace[2pt]
S10 & Irrelevant clutter & Add irrelevant objects or change their spatial density. & Preserve target and receiver; support clutter without interpenetration. \\
\addlinespace[2pt]
S11 & Similar distractors & Add objects of the same class or similar color or shape. & The intended target remains the same. \\
\addlinespace[2pt]
S12 & Persistent occlusion & Place and retain a foreground occluder between the camera and the target region. & Keep the occlusion intervention active during the rollout. \\
\addlinespace[2pt]
S13 & Object appearance & Change target or receiver color, texture, reflectance, or visual surface state. & Preserve task semantics; do not vary physical material parameters. \\
\addlinespace[2pt]
S14 & Contrast or scene domain & Change object-background contrast or the visual scene domain. & Preserve the intended manipulation task. \\
\addlinespace[2pt]
S15 & Camera extrinsics & Apply fixed camera position/orientation offsets or cross-camera calibration offsets. & Offsets are fixed for the episode; affected cameras are case-specific. \\
\addlinespace[2pt]
S16 & Camera image formation & Change field of view, resolution, cropping, or lens distortion. & The configured imaging transformation remains active. \\
\addlinespace[2pt]
S17 & Initial robot state & Change initial joints, end-effector pose/orientation, or gripper opening. & Modify initialization; the robot then executes normally. \\
\addlinespace[2pt]
S18 & Robot calibration & Offset the base, tool center point, joint zero, or grasp center. & Keep the configured calibration error fixed. \\
\addlinespace[2pt]
S19 & Observation quality & Apply compression artifacts, noise, blur, exposure contamination, or missing depth. & Affect only the configured observation channels. \\
\addlinespace[2pt]
S20 & Fixed timing or action window & Apply a fixed delay or change control rate, replanning rate, or the action execution window. & Use the configured timing intervention; distinguish it from model defaults. \\
\addlinespace[2pt]
S21 & Instruction paraphrase & Replace the instruction with a semantic paraphrase. & Preserve goal, referenced objects, and action order. \\
\addlinespace[2pt]
S22 & Support texture & Replace the visual texture of the work surface. & Preserve collision geometry, friction, and initial object poses. \\*
\end{longtable}
\endgroup

\Needspace{10\baselineskip}
\subsection{Dynamic Perturbations}
\label{app:dynamic_perturbation_catalogue}

Table~\ref{tab:dynamic_perturbation_design} summarizes runtime interventions,
including their trigger phases and event lifetimes.
\begingroup
\setlength{\tabcolsep}{4pt}
\renewcommand{\arraystretch}{1.08}
\setlength{\LTleft}{0pt}
\setlength{\LTright}{0pt plus 1fill}
\setlength{\LTcapwidth}{\linewidth}
\setlength{\LPTableWidth}{\dimexpr\linewidth-6\tabcolsep\relax}
\begin{longtable}{@{}>{\raggedright\arraybackslash}p{0.065\LPTableWidth}>{\raggedright\arraybackslash}p{0.215\LPTableWidth}>{\raggedright\arraybackslash}p{0.43\LPTableWidth}>{\raggedright\arraybackslash}p{0.29\LPTableWidth}@{}}
\caption{Dynamic perturbation definitions. Trigger phase and event lifetime are part of each case configuration. A restored sensor or control channel does not imply rollback of the complete scene. IDs match the paper visualizations.}\label{tab:dynamic_perturbation_design}\\
\toprule
\textbf{ID} & \textbf{Perturbation} & \textbf{Intervention} & \textbf{Trigger and lifetime} \\
\midrule
\endfirsthead
\multicolumn{4}{l}{\tablename\ \thetable\ (continued)}\\
\toprule
\textbf{ID} & \textbf{Perturbation} & \textbf{Intervention} & \textbf{Trigger and lifetime} \\
\midrule
\endhead
\midrule
\multicolumn{4}{r}{Continued on the next page}\\
\endfoot
\bottomrule
\endlastfoot
D01 & Target pose update & Change target horizontal position or yaw, requiring updated localization and approach. & Approach/pre-grasp; retain the update as configured. \\
\addlinespace[2pt]
D02 & Receiver pose update & Move or rotate the receiver without changing the task goal. & Post-grasp/pre-place; retain the update as configured. \\
\addlinespace[2pt]
D03 & Receiver occupation & Introduce an occupying or blocking object near the receiving region. & Pre-place; occupation/release follows the case configuration. \\
\addlinespace[2pt]
D04 & Temporary occlusion & Insert an occluder between camera and target, then remove it. & Task phase; enter, hold for a window, then exit. \\
\addlinespace[2pt]
D05 & Path obstacle & Place an obstacle on the upcoming transport route while preserving an alternative collision-free path. & Transport; persistence follows the case configuration. \\
\addlinespace[2pt]
D06 & Distractor rearrangement & Move irrelevant objects or rearrange local clutter; exclude the target and receiver. & Task phase; apply the configured scene update. \\
\addlinespace[2pt]
D07 & Camera motion & Change camera position to represent shake, drift, or a persistent offset. & Task phase; transient or persistent, as configured. \\
\addlinespace[2pt]
D08 & Camera dropout & Invalidate a selected camera observation stream. & Task phase; restore the stream after the event. \\
\addlinespace[2pt]
D09 & Lighting or exposure & Change RGB illumination, shadow, or exposure. & Task phase; restore after the event window. \\
\addlinespace[2pt]
D10 & Frame-stream fault & Drop, repeat, or freeze configured RGB/depth frames. & Task phase; apply the fault during the configured window. \\
\addlinespace[2pt]
D11 & Modality misalignment & Offset timestamps or freeze a modality while other modalities may continue to update. & Task phase; configured modality and event window. \\
\addlinespace[2pt]
D12 & Latency spike & Temporarily delay observation delivery or inference results. & Task phase; restore normal timing after the spike. \\
\addlinespace[2pt]
D13 & Arm--gripper desynchronization & Apply different delays or timing variations to arm and gripper commands. & Task phase; configured timing window. \\
\addlinespace[2pt]
D14 & Command loss or repetition & Disrupt the action stream; the listed implementation repeats the previous action. & Task phase; short command-fault window. \\
\addlinespace[2pt]
D15 & Stale action chunk & Replay an old action chunk or interrupt replanning, temporarily executing outdated actions. & State change; configurable chunk length and stale duration. \\
\addlinespace[2pt]
D16 & Bounded takeover & Inject contact-free bounded actions or a reachable state offset, respecting joint/workspace limits. & Task phase; temporary intervention, then return control. \\
\addlinespace[2pt]
D17 & Grasp failure result & Hold the gripper open on the first grasp attempt while retaining the possibility of retrying. & Grasp attempt; release after about 0.07 m of retreat or window expiry. \\
\addlinespace[2pt]
D18 & Reachable drop state & Place the transported target at a reachable, collision-free recovery location; do not simulate falling forces. & Transport; inject a recoverable outcome state. \\
\addlinespace[2pt]
D19 & Robot joint displacement & Inject one reachable joint displacement without self-collision and within joint/workspace limits. & Robot in motion; one state displacement, not a force simulation. \\
\addlinespace[2pt]
D20 & Instruction restatement & Restate or clarify the instruction while preserving goal, objects, and required action order. & Task phase; apply the equivalent instruction update. \\*
\end{longtable}
\endgroup

\Needspace{10\baselineskip}
\subsection{Training and Evaluation Task Partition}
\label{app:task_partition}

For the controlled eight-task studies, two BASE tasks per suite are reserved
for evaluation; the remaining eight per suite form the training pool
(Table~\ref{tab:perturbation_task_partition}). The exact evaluation instructions
are listed in Table~\ref{tab:perturbation_evaluation_tasks}. These roles do not
imply that the evaluated pretrained checkpoints have never encountered the
evaluation tasks. Demonstration selection and trajectory counts are specified
separately for each training experiment. Long task~8 with D06 is not applicable
and is excluded from the corresponding dynamic-evaluation denominator.

\begingroup
\setlength{\tabcolsep}{4pt}
\renewcommand{\arraystretch}{1.08}
\setlength{\LTleft}{0pt}
\setlength{\LTright}{0pt plus 1fill}
\setlength{\LTcapwidth}{\linewidth}
\setlength{\LPTableWidth}{\dimexpr\linewidth-4\tabcolsep\relax}
\begin{longtable}{@{}>{\raggedright\arraybackslash}p{0.2\LPTableWidth}>{\raggedright\arraybackslash}p{0.24\LPTableWidth}>{\raggedright\arraybackslash}p{0.56\LPTableWidth}@{}}
\caption{Task partition for the controlled eight-task studies. IDs are zero-based indices in the pinned release manifest. The training pool is the complement of the evaluation tasks within each suite; it does not specify which demonstrations were actually used in a training run.}\label{tab:perturbation_task_partition}\\
\toprule
\textbf{Suite} & \textbf{Evaluation IDs} & \textbf{Training-pool IDs} \\
\midrule
\endfirsthead
\multicolumn{3}{l}{\tablename\ \thetable\ (continued)}\\
\toprule
\textbf{Suite} & \textbf{Evaluation IDs} & \textbf{Training-pool IDs} \\
\midrule
\endhead
\midrule
\multicolumn{3}{r}{Continued on the next page}\\
\endfoot
\bottomrule
\endlastfoot
Spatial & 0, 8 & 1, 2, 3, 4, 5, 6, 7, 9 \\*
\addlinespace[2pt]
Object & 1, 8 & 0, 2, 3, 4, 5, 6, 7, 9 \\*
\addlinespace[2pt]
Goal & 3, 6 & 0, 1, 2, 4, 5, 7, 8, 9 \\*
\addlinespace[2pt]
Long & 5, 8 & 0, 1, 2, 3, 4, 6, 7, 9 \\*
\end{longtable}
\endgroup

\begingroup
\setlength{\tabcolsep}{4pt}
\renewcommand{\arraystretch}{1.08}
\setlength{\LTleft}{0pt}
\setlength{\LTright}{0pt plus 1fill}
\setlength{\LTcapwidth}{\linewidth}
\setlength{\LPTableWidth}{\dimexpr\linewidth-4\tabcolsep\relax}
\begin{longtable}{@{}>{\raggedright\arraybackslash}p{0.14\LPTableWidth}>{\raggedright\arraybackslash}p{0.07\LPTableWidth}>{\raggedright\arraybackslash}p{0.79\LPTableWidth}@{}}
\caption{The eight evaluation tasks. Instructions are taken directly from the BASE scene descriptions, with whitespace normalized.}\label{tab:perturbation_evaluation_tasks}\\
\toprule
\textbf{Suite} & \textbf{ID} & \textbf{BASE task instruction} \\
\midrule
\endfirsthead
\multicolumn{3}{l}{\tablename\ \thetable\ (continued)}\\
\toprule
\textbf{Suite} & \textbf{ID} & \textbf{BASE task instruction} \\
\midrule
\endhead
\midrule
\multicolumn{3}{r}{Continued on the next page}\\
\endfoot
\bottomrule
\endlastfoot
Spatial & 0 & Pick the akita black bowl between the plate and the ramekin and place it on the plate \\*
\addlinespace[2pt]
Spatial & 8 & Pick the akita black bowl next to the plate and place it on the plate \\*
\addlinespace[2pt]
Object & 1 & Pick the cream cheese and place it in the basket \\*
\addlinespace[2pt]
Object & 8 & Pick the chocolate pudding and place it in the basket \\*
\addlinespace[2pt]
Goal & 3 & Open the top layer of the drawer and put the bowl inside \\*
\addlinespace[2pt]
Goal & 6 & Put the cream cheese on the bowl \\*
\addlinespace[2pt]
Long & 5 & pick up the book and place it in the back compartment of the caddy \\*
\addlinespace[2pt]
Long & 8 & put both moka pots on the stove \\*
\end{longtable}
\endgroup

\begingroup
\setlength{\tabcolsep}{4pt}
\renewcommand{\arraystretch}{1.08}
\setlength{\LTleft}{0pt}
\setlength{\LTright}{0pt plus 1fill}
\setlength{\LTcapwidth}{\linewidth}
\setlength{\LPTableWidth}{\dimexpr\linewidth-4\tabcolsep\relax}
\begin{longtable}{@{}>{\raggedright\arraybackslash}p{0.14\LPTableWidth}>{\raggedright\arraybackslash}p{0.07\LPTableWidth}>{\raggedright\arraybackslash}p{0.79\LPTableWidth}@{}}
\caption{The 32 training tasks. Instructions are taken directly from the BASE scene descriptions, with whitespace normalized.}\label{tab:perturbation_training_tasks}\\
\toprule
\textbf{Suite} & \textbf{ID} & \textbf{BASE task instruction} \\
\midrule
\endfirsthead
\multicolumn{3}{l}{\tablename\ \thetable\ (continued)}\\
\toprule
\textbf{Suite} & \textbf{ID} & \textbf{BASE task instruction} \\
\midrule
\endhead
\midrule
\multicolumn{3}{r}{Continued on the next page}\\
\endfoot
\bottomrule
\endlastfoot
Spatial & 1 & Pick the akita black bowl next to the ramekin and place it on the plate \\
\addlinespace[2pt]
Spatial & 2 & Pick the akita black bowl from table center and place it on the plate \\
\addlinespace[2pt]
Spatial & 3 & Pick the akita black bowl on the cookies box and place it on the plate \\
\addlinespace[2pt]
Spatial & 4 & Pick the akita black bowl in the top layer of the wooden cabinet and place it on the plate \\
\addlinespace[2pt]
Spatial & 5 & Pick the akita black bowl on the ramekin and place it on the plate \\
\addlinespace[2pt]
Spatial & 6 & Pick the akita black bowl next to the cookies box and place it on the plate \\
\addlinespace[2pt]
Spatial & 7 & Pick the akita black bowl on the stove and place it on the plate \\
\addlinespace[2pt]
Spatial & 9 & Pick the akita black bowl on the wooden cabinet and place it on the plate \\
\addlinespace[2pt]
Object & 0 & Pick the alphabet soup and place it in the basket \\
\addlinespace[2pt]
Object & 2 & Pick the salad dressing and place it in the basket \\
\addlinespace[2pt]
Object & 3 & Pick the bbq sauce and place it in the basket \\
\addlinespace[2pt]
Object & 4 & Pick the ketchup and place it in the basket \\
\addlinespace[2pt]
Object & 5 & Pick the tomato sauce and place it in the basket \\
\addlinespace[2pt]
Object & 6 & Pick the butter and place it in the basket \\
\addlinespace[2pt]
Object & 7 & Pick the milk and place it in the basket \\
\addlinespace[2pt]
Object & 9 & Pick the orange juice and place it in the basket \\
\addlinespace[2pt]
Goal & 0 & Open the middle layer of the drawer \\
\addlinespace[2pt]
Goal & 1 & Put the bowl on the stove \\
\addlinespace[2pt]
Goal & 2 & Put the wine bottle on the top of the drawer \\
\addlinespace[2pt]
Goal & 4 & Put the bowl on the top of the drawer \\
\addlinespace[2pt]
Goal & 5 & Push the plate to the front of the stove \\
\addlinespace[2pt]
Goal & 7 & Turn on the stove \\
\addlinespace[2pt]
Goal & 8 & Put the bowl on the plate \\
\addlinespace[2pt]
Goal & 9 & Put the wine bottle on the rack \\
\addlinespace[2pt]
Long & 0 & put both the alphabet soup and the tomato sauce in the basket \\
\addlinespace[2pt]
Long & 1 & put both the cream cheese box and the butter in the basket \\
\addlinespace[2pt]
Long & 2 & turn on the stove and put the moka pot on it \\
\addlinespace[2pt]
Long & 3 & put the black bowl in the bottom drawer of the cabinet and close it \\
\addlinespace[2pt]
Long & 4 & put the white mug on the left plate and put the yellow and white mug on the right plate \\
\addlinespace[2pt]
Long & 6 & put the white mug on the plate and put the chocolate pudding to the right of the plate \\
\addlinespace[2pt]
Long & 7 & put both the alphabet soup and the cream cheese box in the basket \\
\addlinespace[2pt]
Long & 9 & put the yellow and white mug in the microwave and close it \\
\end{longtable}
\endgroup

\subsection{Evaluation Protocol}
\label{app:evaluation_protocol}

We evaluate all policies from fixed checkpoints under identical benchmark settings. Each policy receives observation modalities matching its architecture, while environment settings, perturbation parameters, and task success criteria are held constant across all models. We retain each policy's original action representation and prediction horizon unless explicitly altered in diagnostic ablation experiments. No privileged information about the perturbation is available to the policy in standard evaluation runs.

An episode is marked successful only if the agent fulfills the original task completion requirements within the evaluation time horizon. Episodes that end due to timeout or failed manipulation are counted as failures.

Our main evaluation metric is task success rate (SR). For model $m$ and perturbation case $c$, we define
\begin{equation}
\mathrm{SR}_{m,c}
=
\frac{1}{|\mathcal{T}_c|}
\sum_{t \in \mathcal{T}_c}
\frac{1}{N_{t,c}}
\sum_{r=1}^{N_{t,c}}
\mathbb{I}
\left[
\mathrm{Success}(m,t,c,r)
\right],
\end{equation}
where $\mathcal{T}_c$ is the set of tasks compatible with perturbation $c$, $N_{t,c}$ denotes the total number of rollouts for task $t$ under perturbation $c$, and $\mathbb{I}[\cdot]$ represents the binary success indicator function.

\paragraph{Execution Horizon Configuration.}
For RQ1-B, we investigate the effect of policy update frequency by varying the number of executed actions $K$ before obtaining a new observation and generating the next action chunk, while keeping each policy's prediction horizon $H$ fixed. We consider four nominal execution ratios, $K/H \in \{1/8, 1/4, 1/2, 1\}$. Since different policies use different prediction horizons, the corresponding values of $K$ vary across models, as detailed in Table~\ref{tab:execution_horizon}. For $\pi_{0.5}$, non-integer values are rounded to valid integer action steps.

\begin{table}[t]
\centering
\caption{Executed action lengths $K$ for different policies and nominal execution horizon ratios in RQ1-B. The prediction horizon $H$ remains fixed for each policy.}
\label{tab:execution_horizon}
\small
\setlength{\tabcolsep}{6pt}
\renewcommand{\arraystretch}{1.1}
\begin{tabular}{lccccc}
\toprule
\textbf{Model} & $\boldsymbol{H}$ & $\boldsymbol{1/8H}$ & $\boldsymbol{1/4H}$ & $\boldsymbol{1/2H}$ & $\boldsymbol{1H}$ \\
\midrule
OpenVLA-OFT   & 8  & 1 & 2 & 4  & 8  \\
$\pi_{0.5}$   & 10 & 1 & 3 & 5  & 10 \\
Anchor-Align  & 8  & 1 & 2 & 4  & 8  \\
FastWAM       & 32 & 4 & 8 & 16 & 32 \\
Cosmos Policy & 16 & 2 & 4 & 8  & 16 \\
\bottomrule
\end{tabular}
\end{table}

\subsection{Training Setup}
\label{app:train_setup}

\paragraph{Training Data.}
We fine-tune $\pi_{0.5}$ using a mixture of perturbation-augmented demonstrations and original LIBERO demonstrations.
For each of the ten perturbation cases across five static--dynamic families, we select two tasks from each of the four LIBERO suites, yielding eight training tasks per case.
We collect ten demonstrations per task, resulting in $10 \times 8 \times 10 = 800$ perturbation-augmented trajectories.
To preserve the policy's original manipulation capabilities, we additionally include standard LIBERO demonstrations.
Specifically, we sample 20 demonstrations from each of the 40 original tasks across the four suites, yielding another $40 \times 20 = 800$ trajectories.
The two datasets are mixed at a 1:1 ratio, resulting in a total of 1,600 training trajectories.

\paragraph{Fine-tuning Configuration.}
We adopt Low-Rank Adaptation (LoRA) for parameter-efficient fine-tuning of $\pi_{0.5}$.
The LoRA rank is set to 32 for the Gemma vision-language model (VLM) and 16 for the Gemma action head.
We use an effective batch size of 144 and train for one epoch over the mixed dataset.

\endgroup

\endgroup
\end{document}